%% file: main.tex
\documentclass[]{fairmeta}

\usepackage{amsmath}
\usepackage{amssymb}
\usepackage{mathtools}
\usepackage{amsthm}
\usepackage{bbm}

\theoremstyle{plain}

\theoremstyle{definition}

\theoremstyle{remark}

\usepackage[textsize=tiny]{todonotes}

\usepackage{fancyvrb}
\usepackage{listings}
\usepackage{tcolorbox}
\usepackage{booktabs}
\usepackage{xcolor}

\newtcolorbox{promptbox}{
  colback=gray!5,
  colframe=gray!40,
  boxrule=0.1pt,
  arc=2pt,
  left=6pt,right=6pt,top=6pt,bottom=6pt
}

\lstdefinestyle{prompt}{
  basicstyle=\ttfamily\scriptsize,
  columns=fullflexible,
  breaklines=true,
  frame=none
}
\usepackage{xurl}
\usepackage{array}
\usepackage{pifont}
\definecolor{darkorange}{RGB}{255,140,0}
\definecolor{crimson}{RGB}{220,20,60}

\newcounter{fnrust}

\title{Detecting \textit{Functional} Memorization in Code Language Models}

\author[1,2]{Matthieu Meeus}
\author[1]{Anil Ramakrishna}
\author[1]{Shengyuan Hu}
\author[1]{Matthew Grange}
\author[1]{Zheng Xu}
\author[1]{Luca Melis}

\affiliation[1]{Meta}
\affiliation[2]{Imperial College London}

\abstract{Large language models (LLMs) are increasingly used to generate code at scale. Meanwhile, prior work has investigated whether training data may be recoverable from model outputs, by auditing the textual overlap between training examples and model generations. Code, however, can preserve the same logic while differing substantially in syntax and structure. We here study \emph{functional memorization}: the leakage of training data logic from LLM generations in ways that textual audits fail to detect. We leverage AI coding agents to generate diverse test inputs for training data functionality and evaluate whether model-generated continuations produce the same outputs. We formalize this through a \emph{counterfactual} framework, comparing target models (exposed to specific code) against reference models (not exposed) and requiring functional equivalence only for the target. We instantiate this framework across 4 open-source models and explicitly filter for functions with meaningful logic in 5 programming languages. We find that 0.3--3.4\% of filtered functions are counterfactually functionally memorized, i.e., reproduced with equivalent behavior in restructured code that textual metrics fail to detect. We further show that LLM-based judges offer a scalable proxy for execution-based testing, achieving a true positive rate of 68\% at 1\% false positive rate, and find that functional memorization is associated with semantic duplication in the training corpus.}

\correspondence{Luca Melis at \email{lucamelis@meta.com}}

\begin{document}

\maketitle

\section{Introduction}
\input{sections/1_introduction}

\section{Background}
\label{sec:background}
\input{sections/2_background}

\section{Functional memorization}
\label{sec:method}
\input{sections/3_method}

\section{Experimental setup}
\label{sec:exp_setup}
\input{sections/4_exp_setup}

\section{Results}
\input{sections/5_results}

\section{Related work}
\label{sec:related_work}
\input{sections/6_related_work}

\section{Limitations and future work}

\input{sections/7_discussion}

\bibliographystyle{plainnat}
\bibliography{bibliography}

%%%%%%%%%%%%%%%%%%%%%%%%%%%%%%%%%%%%%%%%%%%%%%%%%%%%%%%%%%%%%%%%%%%%%%%%%%%%%%%
%%%%%%%%%%%%%%%%%%%%%%%%%%%%%%%%%%%%%%%%%%%%%%%%%%%%%%%%%%%%%%%%%%%%%%%%%%%%%%%
% APPENDIX
%%%%%%%%%%%%%%%%%%%%%%%%%%%%%%%%%%%%%%%%%%%%%%%%%%%%%%%%%%%%%%%%%%%%%%%%%%%%%%%
%%%%%%%%%%%%%%%%%%%%%%%%%%%%%%%%%%%%%%%%%%%%%%%%%%%%%%%%%%%%%%%%%%%%%%%%%%%%%%%
\newpage
\appendix
\onecolumn

\section{Filtering for meaningful functional logic}
\label{app:filtering}
\input{appendix/filtering}

\section{Additional results for (near-)verbatim memorization rates}
\label{app:add_verbatim_results}
\input{appendix/add_verbatim_results}

\section{Details for execution-based similarity testing}
\label{app:details_agent}
\input{appendix/details_execution}

\clearpage
\newpage

\section{Details for code similarity metrics}
\label{app:details_metrics}
\input{appendix/details_metrics}

\clearpage
\newpage

\section{Qualitative examples of memorized functions}
\label{app:examples}

\input{appendix/examples}

\clearpage
\newpage

\section{Additional results across similarity metrics}
\label{app:add_metrics}
\input{appendix/add_metrics_plots}

\clearpage
\newpage

\section{Duplication analysis across programming languages}
\label{app:add_duplication}
\input{appendix/add_duplication_plots}

\section{Post-training mitigations}
\label{app:add_posttraining}
\input{appendix/add_posttraining}

\end{document}

%% file: sections/1_introduction.tex
Large language models (LLMs) are increasingly used to generate code, both as standalone models~\citep{roziere2023code,feng2020codebert,olmo2025olmo,li2022competition} and as core components of coding agents~\citep{wired2024claudecode,nytimes2026aicoding}. These capabilities are acquired by training on large corpora of source code, often scraped from public repositories with permissive licenses~\citep{allal2025smollm2smolgoesbig,li2022competition,roziere2023code}. However, filtering out non-permissively licensed code is difficult~\citep{katzy2024exploratory}, while training data is also increasingly sourced from user coding sessions~\citep{github2024copilotpolicy,techcrunch2025anthropic}.

At the same time, prior work has studied the extent to which training data might be recoverable from model outputs, as studied extensively for natural language~\citep{carlini2021extracting,nasr2025scalable,ippolito2023preventing}. Similar techniques have also been applied to code, measuring near-verbatim leakage of source code~\citep{al2024traces,salerno2025much} and secrets such as API keys~\citep{nie2025decoding,huang2024your}.

However, existing work on code leakage focuses almost exclusively on \emph{textual} similarity: does the model's output match the training data token-for-token, or nearly so? This framing misses a critical dimension when models are trained on code containing proprietary logic, e.g., algorithms for recommendation systems, trading, or content moderation, that owners would reasonably want to protect. Indeed, code can be \emph{functionally equivalent} while being textually dissimilar: variables can be renamed, comments added or removed, control flow restructured, or entirely different algorithms used to achieve the same functionality---as has been widely studied in code clone detection~\citep{roy2007survey,ren2020codebleu,song2024revisiting}. Hence, if a model internalizes logic from its training data and reproduces it in a different form, memorization auditing based on textual similarity will fail to detect it.% while the underlying proprietary logic would still be leaked.

\begin{figure*}[t]
\centering
\includegraphics[width=0.7\linewidth]{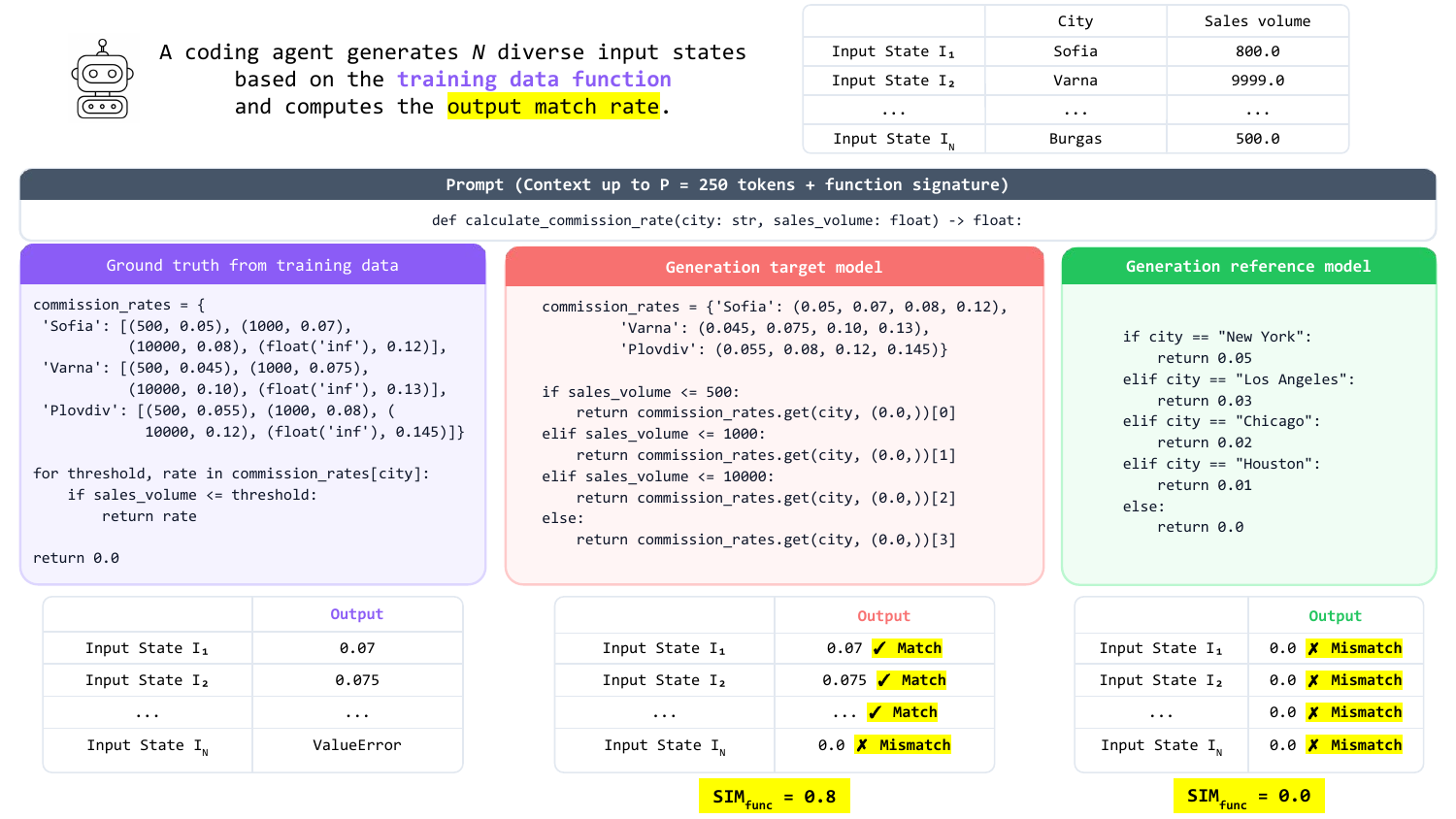}
\caption{\textbf{Execution-based testing reveals counterfactual \emph{functional} memorization}. We prompt OLMo-3-32B with a function signature from its training data and compare the continuations generated by the target and reference models against the ground-truth. A coding agent generates $N$ diverse inputs, executes the original function and the generations and compares their outputs. Textual auditing fails to flag the target model's generation (BLEU~$=0.26$), yet execution-based testing reveals that it has memorized and reproduces the same hard-coded logic from the training data ($\textsc{SIM}_\text{func}=0.8$), while the reference model hallucinates ($\textsc{SIM}_\text{func}=0.0$). Full example in Appendix~\ref{app:examples_functional}.}
\label{fig:commission_example}
\end{figure*}

\textbf{Contributions.} We here formalize \emph{functional memorization}: given a training data function $x$ and a generation $x^\star_T$ from a target model prompted with the function signature, we consider $x$ functionally memorized when the generation is functionally equivalent ($\textsc{SIM}_\text{func}(x, x_T^\star) \ge \tau_\text{func}$) despite low textual overlap ($\textsc{SIM}_\text{text}(x, x_T^\star) < \tau_\text{text}$). To distinguish memorization from generalization~\citep{liulanguage}, we instantiate a \emph{counterfactual} setup~\citep{feldman2020does,zhang2023counterfactual,hayes2024measuring} and require a reference model---as similar as possible to the target but without exposure to the data of interest---to \emph{not} produce a functionally equivalent completion. 

To study the leakage of functional logic in its most abstract form, we evaluate whether a model generation $x^\star_T$ exhibits the same \emph{behavior} as the training data function $x$ through execution-based testing. Coding agents~\citep{anthropic2025claudecode,openai2025codex} generate diverse input states, execute both $x$ and $x^\star_T$, and compute functional similarity $\textsc{SIM}_\text{func}$ as the match rate in output states. 

We instantiate this framework to 4 open-source models and 5 programming languages. For each setup, we extract functions from late-stage pretraining data and use an LLM-judge to select those containing meaningful logic, e.g., hard-coded business rules, that the original author would reasonably want to protect. This is a deliberate design choice: if \emph{any} of these functions leak, it constitutes a real-world IP violation and an audit that relies solely on textual similarity and misses such cases provides a false sense of security.

We find that 0.3--3.4\% of functions are counterfactually functionally memorized: reproduced by the target model with equivalent behavior that textual metrics fail to detect (Figure~\ref{fig:memorization_combined}). Figure~\ref{fig:commission_example} provides a compelling example: the target model reproduces training data logic in a completely restructured implementation. This functional leakage is largely orthogonal to near-verbatim memorization---of functions flagged by execution-based testing, over 90\% are missed by textual similarity. As a baseline, 0.3--1.0\% of functions are also counterfactually memorized near-verbatim, meaning that the two auditing strategies capture complementary phenomena at comparable scale. We recover functional memorization across programming languages, model scales, and training setups (Figure~\ref{fig:memorization_combined}, examples in Appendix~\ref{app:examples_function_across_setups}). Different datasets correspond to distinct levels of ``difficulty'', as reference models reproduce some subsets at higher rates even without explicit exposure to the target data, emphasizing the need for counterfactual calibration. Larger models memorize more, with OLMo-3-32B exhibiting 20--70\% higher functional memorization rates than its 7B counterpart.

Since agentic execution is costly (on average 2 minutes per sample), we then evaluate scalable alternative metrics: structural metrics proposed in the literature on code clone detection~\citep{ren2020codebleu}, embedding-based similarity and LLM-judges~\citep{song2024revisiting,nikiema2025small}. We find that LLM-judges offer the best proxy, achieving a true positive rate (TPR) of 68\% at 1\% false positive rate (FPR). 

Finally, we find that functional memorization is associated with the accumulation of semantically similar variants in the training data, suggesting semantic deduplication~\citep{abbas2023semdedup} as a mitigation technique. 

Together, our results provide the first evidence that code LLMs may memorize and reproduce training logic in forms that evade textual auditing, underscoring the need for functional auditing and mitigation.

%% file: sections/2_background.tex
\textbf{Training data extraction.} We consider an LLM $\mathcal{M}$ trained on dataset $\mathcal{D}$ that contains source code. Following prior work on memorization~\citep{carlini2022quantifying,nasr2025scalable,ippolito2023preventing}, we study training data extraction attacks that prompt the model with a prefix $p$ (of token length $P$) taken from a training sequence, and then decode a continuation $x^\star \gets \mathcal{M}(p)$. We focus on \emph{greedy decoding}, i.e., consecutively sampling the token with the greatest predicted probability, and quantify leakage by comparing the ground-truth continuation $x$ to $x^\star$ using similarity metric $\textsc{SIM}$. Prior work has considered different textual similarity metrics, ranging from exact~\citep{carlini2021extracting,carlini2022quantifying} to near-verbatim matches~\citep{ippolito2023preventing}. 

\textbf{Counterfactual memorization.} A key challenge when studying memorization through extraction is distinguishing it from generalization~\citep{liulanguage}: when a model generates a sequence similar to its training data, is it reproducing memorized content or independently arriving at a natural solution? To address this,~\citet{zhang2023counterfactual,feldman2020does} propose using a definition of \emph{counterfactual} memorization, characterizing how a model's predictions change when a particular sample is omitted during training. Building on this,~\citet{hayes2024measuring} instantiate counterfactual memorization in the context of training data extraction from LLMs. Let a target model $\mathcal{M}_T$ be trained on dataset $\mathcal{D}$ containing the training sample $z = p||x$, and let a \emph{reference} model $\mathcal{M}_R$ be trained on $\mathcal{D} \setminus \{z\}$. Then $z$ is counterfactually memorized if $\mathcal{M}_T$ produces $x$ when prompted with $p$ using greedy decoding, while $\mathcal{M}_R$ does not. Further, $z$ is said to be $k$-approximately counterfactually memorized if the edit distance between $x$ and the $\mathcal{M}_T$ completion is $\leq k$, while the edit distance between $x$ and the $\mathcal{M}_R$ completion is $> k$. In practice, counterfactual reference models are expensive to obtain, and have been approximated by earlier model checkpoints before training on target data~\citep{hayes2024measuring}. 

\textbf{Code similarity.} In parallel to memorization studies, prior work has studied the similarity between pieces of code in the context of clone detection and evaluation of synthesized code.~\citet{roy2007survey} categorize clones into four types: Type~I (identical up to whitespace/comments), Type~II (renaming of identifiers), Type~III (modified statements), and Type~IV (functional equivalence with different syntax).
Motivated by this taxonomy, many metrics have been proposed to detect clones beyond string matching.~\citet{ren2020codebleu} propose CodeBLEU, combining $n$-gram overlap with syntactic similarity (AST matching) and semantic signals (data-flow). Follow-up work propose refined structural similarity measures (e.g., AST edit distance)~\citep{song2024revisiting,yu2025multiple} and metrics based on embedding similarity using (code-specific) representation models~\citep{feng2020codebert,suresh2025cornstack}. Recently, LLM-judges were explored for clone detection, showing improved sensitivity for Type~III-IV similarity~\citep{almatrafi2025code,maveli2025can,nikiema2025small}. Finally,~\citet{liang2025hyclone} incorporates function \emph{execution} with LLM-generated inputs and find it to outperform other methods, including LLM-judges, for a set of curated Python functions.

%% file: sections/3_method.tex
\textbf{Formalization.} Our goal is to detect and quantify \emph{functional memorization}: cases where a code model reproduces the functional logic of a training function without reproducing its text (near) verbatim. We distinguish textual overlap, measured by $\textsc{SIM}_\text{text}$ (e.g., BLEU), from functional overlap, measured by $\textsc{SIM}_\text{func}$ between the ground-truth $x$ and the generated continuation $x^\star$. We say a sample $z = p || x$ is \emph{counterfactually functionally memorized} by target model $\mathcal{M}_T$ if its generation $x_T^\star$ satisfies: 
\begin{enumerate}
    \item functional equivalence, i.e., $\textsc{SIM}_\text{func}(x, x_T^\star) \ge \tau_\text{func}$;
    \item low textual overlap with $x$, i.e., $\textsc{SIM}_\text{text}(x, x_T^\star) < \tau_\text{text}$;
    \item a counterfactual divergence, i.e., a reference model's generation for the same prompt is \emph{not} functionally equivalent, or $\textsc{SIM}_\text{func}(x, x_R^\star) < \tau_\text{func}$.
\end{enumerate}

Together, this isolates a phenomenon distinct from verbatim leakage: the target model has internalized the \emph{logic} of the training data and reproduces it in a different surface form. Condition~(3), following~\citet{feldman2020does, zhang2023counterfactual,hayes2024measuring}, filters out high similarity attributable to the target model's generalization capabilities.

\textbf{Execution-based testing.} To examine functional leakage in its most abstract form~\citep{roy2007survey,liang2025hyclone}, we compute functional similarity between two pieces of code based on execution-verified input-output equivalence. Given a ground-truth function $x$ from the training data, we generate $N$ diverse input states, execute both $x$ and its model-generated counterpart $x^\star$ on each, and compare their output states. 

Executing functions from real-world training data is far from trivial. First, most functions cannot run in isolation: they depend on imported packages, call external APIs, or are methods of a class instance---all of which must be resolved or mocked before execution is even possible. Second, real-world functions exhibit behavior beyond pure input--output mappings: they read ambient state, mutate objects, write files, and produce logs, requiring richer notions of both input and output.

We formalize this as follows. An \emph{input state} $I_k$ specifies everything needed to execute $x$ once: argument values together with any ambient state the function reads (e.g., global attributes, mocked dependency behaviors, file contents). An \emph{output state} $\mathrm{Exec}(x, I_k)$ is the observation tuple produced by running $x$ under $I_k$: return values, any exception, the ordered log of mocked-dependency calls, mutations to \texttt{self} and mutable arguments, captured stdout/stderr, and side effects (file writes, requests). Two executions match only if their output-state tuples are identical. We define functional similarity between function $x$ and $x^\star$ as:

\begin{equation}
    \textsc{SIM}_\text{func}(x, x^\star) = \frac{1}{N} \sum_{k=1}^{N} \mathbbm{1}\bigl[\mathrm{Exec}(x, I_k) = \mathrm{Exec}(x^\star, I_k)\bigr].
\label{eq:exec_sim}
\end{equation}

To compute this at scale, we instantiate state-of-the-art AI coding agents~\citep{anthropic2025claudecode,openai2025codex}. Agents are particularly well-suited to this task: they construct an executable environment for arbitrary functions---by resolving dependencies, mocking external services, and initializing required global state---and generate diverse input states and capture resulting output states. To ensure validity, we require (i) at least $N_\text{in}$ distinct input states that successfully run on $x$, (ii) with minimum line coverage $C$ (i.e., fraction of the executable lines exercised across inputs) and (iii) at least $N_\text{out}$ distinct output states. We provide further details on execution-based testing in Section~\ref{sec:exp_setup} and Appendix~\ref{app:details_agent}, illustrate an example in Figure~\ref{fig:commission_example} and evaluate how other classes of metrics approximate this execution-based metric in Section~\ref{sec:results_func}.

%% file: sections/4_exp_setup.tex
\textbf{Target models and data.} We audit for the leakage of functional logic from code in \emph{real-world} LLM pretraining. At this scale, constructing an ideal counterfactual---a model trained on identical data minus the point of interest---quickly becomes computationally prohibitive. Instead, we leverage open-source artifacts to construct approximate counterfactual setups where the reference and target models share architecture and pretraining but differ in exposure to the target data. \textbf{(i) OLMo-3}~\citep{olmo2025olmo}. We study memorization during midtraining on the Dolmino corpus for both 32B and 7B models. We use pretrained checkpoints as references (`step656000' for 32B; `step1413814' for 7B) and midtrained checkpoints as targets (`step23842' for 32B; `step47684' for 7B). We consider code from two Dolmino subsets: CraneCode%(a set of permissively-licensed Python data preprocessed by the SwallowCode pipeline
~\citep{fujii2025rewritingpretrainingdataboosts}) and code in Python, Java, C, Rust, and Go from Stack-Edu~\citep{allal2025smollm2smolgoesbig} (top 20\% by quality). Notably, Stack-Edu also appears in pretraining, so our setup captures additional memorization upon re-exposure while CraneCode is heavily rewritten relative to any pretraining overlap. \textbf{(ii) StarCoder}~\citep{li2023starcoder}. A 15B-parameter model pretrained on the 250B-token corpus StarCoderData, then further trained on its 35B Python tokens for ${\sim}$2 epochs. We use StarCoderBase as the reference and StarCoder as the target model, with the Python subset as target data. \textbf{(iii)  CrystalCoder}~\citep{liu2023llm360}. A 7B-parameter model also pretrained on code from StarCoderData~\citep{li2023starcoder}. We use checkpoint `214387' as the reference model, the final model serves as the target and consider the Python subset from Phase 3 as target data.

\textbf{Data filtering.} We isolate \emph{functions} from entire coding files (relying on parsing for Python and indentation for other languages) with body lengths between 10 and 50 lines, and dismiss any processed functions with fill-in-the-middle~\citep{bavarian2022efficient}. To ensure our evaluation focuses on functions whose memorization would be genuinely concerning, we instantiate LLaMA-3.1-8B-Instruct~\citep{grattafiori2024llama} as judge to identify functions that contain meaningful logic. For each dataset, we randomly sample functions until we obtain 5,000 that pass this filter. The exceptions are Rust and Go, where the respective subsets contain only 2,612 and 3,735 qualifying functions. Further details are provided in Appendix~\ref{app:filtering}.

\textbf{Completion generation.} We query target and reference models to complete each function based on a prefix consisting of its signature (excluding the docstring) and its preceding context up to $P=250$ tokens. We generate continuations using greedy decoding with a maximum length of 500 tokens and then parse a valid function from the generated output. 

\textbf{Similarity metrics.} We here describe all metrics and their implementation details used throughout this work to examine the similarity between the training data function $x$ and its generated counterpart $x^\star$. 

\textit{Textual similarity.} We adopt the following text-based similarity metrics: \textit{Exact Match} (binary), the approximate measures \textit{BLEU} (1–4 gram overlap) and \textit{Edit Similarity} (Levenshtein distance normalized by $|x|$) and the normalized length of the longest common substring (\textit{Substring}). Following~\citet{ippolito2023preventing}, we set threshold $\tau_\text{text}=0.75$ for BLEU to measure near-verbatim memorization. 

\textit{Agentic execution-based testing.} We use Claude Code~\citep{anthropic2025claudecode} with Opus-4.7 as coding agent to compute functional similarity. We instantiate one agent per sample, consisting of the training data function and the target and reference model generations. The agent is instructed to generate $N=10$ input states per function and iteratively refine its scaffolding until sufficient evidence is obtained. We require at least $N_\text{in}=8$ distinct valid input states (i.e., $x$ executes without error), with a minimum line coverage of $C=80\%$ and at least $N_\text{out}=2$ distinct output states (to confirm the inputs exercise different code paths while still accommodating functions with binary outputs). If any of these conditions are not met after the agent's iterative attempts, the sample is marked inconclusive. Across setups, we recover a conclusive similarity in approximately 90\% of the functions. We provide more implementation details in Appendix~\ref{app:details_agent}, including the full prompt, system-level details, non-agentic ablations (yielding a coverage of only 12.8\%) and agentic ablations with other models and harnesses (e.g., Codex with GPT-5.5~\citep{openai2025codex}) (Figure~\ref{fig:corr_across_agents}). Table~\ref{tab:timing_exec} further reports execution time across languages, yielding on average 2 (Python) to 5 (Rust) minutes per sample. To consider two pieces of code functionally similar, we set the threshold $\tau_\text{func} = 0.8$ for execution-based similarity, requiring 80\% of the input-output behavior to be equivalent. This value allows for divergence on minimal execution paths (e.g., raising a \texttt{ValueError} versus returning \texttt{0.0} in Figure~\ref{fig:commission_example}) without considering fundamentally different functions. We report results for other threshold values in Figure~\ref{fig:threshold_sensitivity}.

\textit{Other functional similarity metrics.} In Section~\ref{sec:results_func}, we study whether the expensive execution-based testing could be reasonably approximated by other metrics proposed in the literature on code clone detection. For these, we consider 3 classes of functional similarity metrics: (i) \emph{Structural code metrics}: from CodeBLEU~\citep{ren2020codebleu}, a weighted sum of $n$-gram overlap, AST subtree matching, and data-flow matching, we use the default metric with equal weights (CodeBLEU (equal)), its syntax-only AST subtree match (CodeBLEU (syntax)) and data-flow-only variant (CodeBLEU (DFG)). We also consider TSED, the AST-edit-distance score of~\citet{song2024revisiting}.
(ii) \emph{Embedding-based metrics}: we compute the cosine similarity of embeddings produced by dedicated pretrained models, considering Qwen3-Embedding (0.6B, 4B and 8B)~\citep{yang2025qwen3}, and Nomic-Embed-Code-7B~\citep{suresh2025cornstack}. (iii) \emph{LLM judge}: we prompt an LLM to assess functional equivalence (with a score in $[0,1]$), using the prompts from~\citet{song2024revisiting} and~\citet{nikiema2025small} and our own functional-equivalence judge (Ours) (full prompts in Appendix~\ref{app:prompts_judge}). We instantiate the LLM judges with LLaMA-3.3-70B-Instruct~\citep{grattafiori2024llama} by default, and use Claude Sonnet-4.6~\citep{anthropic2026sonnet46} where explicitly specified.

%% file: sections/5_results.tex
\subsection{Baseline: quantifying near-verbatim memorization}
\label{sec:results_near_verbatim}

\begin{figure}[t] 
\centering 
\begin{subfigure}[t]{\linewidth} 
\centering 
\includegraphics[width=0.8\linewidth]{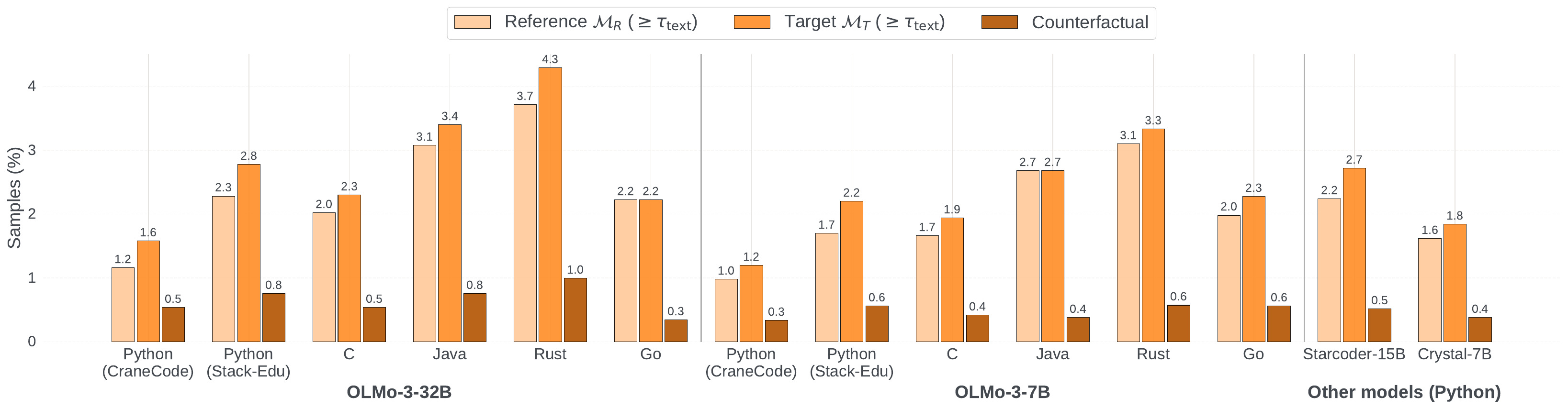} 
\label{fig:verbatim_mem_bleu} 
\end{subfigure} 
\vspace{0.5em} 
\begin{subfigure}[t]{\linewidth}
\centering 
\includegraphics[width=0.8\linewidth]{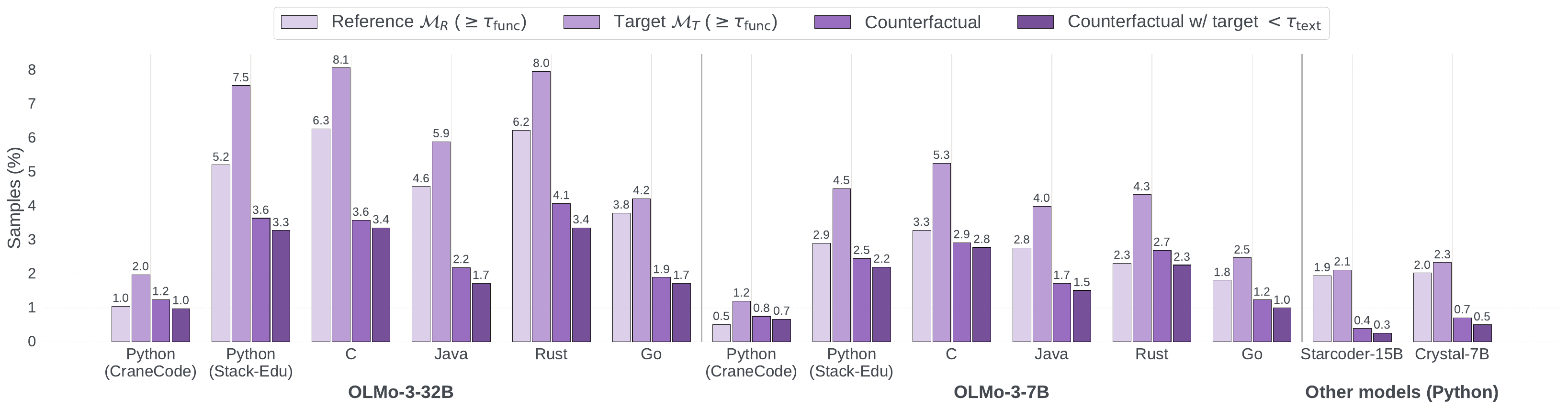} 
\label{fig:func_mem_execution_models}
\end{subfigure} 
\caption{Fraction of functions generated with $\textsc{SIM} \geq \tau$ for the reference and target model and the fraction that is counterfactually memorized. Results for OLMo-3-32B/7B across datasets, and for Python across StarCoder-15B and CrystalCoder-7B. (Top)~Near verbatim memorization measured via BLEU ($\tau_\text{text} =0.75$; exact match rates in Figure~\ref{fig:verbatim_mem_exact}). (Bottom)~Functional memorization measured via execution-based testing ($\tau_\text{func} = 0.8$), alongside the subset missed when relying solely on textual similarity. Examples in Figure~\ref{fig:commission_example} and Appendix~\ref{app:examples}.} 
\label{fig:memorization_combined} 
\end{figure}

As a baseline, we first quantify (near) verbatim memorization rates following prior work on natural language~\citep{carlini2022quantifying,ippolito2023preventing}. Figure~\ref{fig:memorization_combined} shows the fraction of samples whose generation matches the training data function near-verbatim for both the target and reference models, as well as the number memorized counterfactually, i.e., $\textsc{SIM}_\text{text}(x, x_T^\star) \geq \tau_\text{text}$ while $\textsc{SIM}_\text{text}(x, x_R^\star) < \tau_\text{text}$. Following~\citet{ippolito2023preventing}, we consider BLEU score with $\tau_\text{text} = 0.75$ and provide results for exact matches in Figure~\ref{fig:verbatim_mem_exact}. 

For Python functions from CraneCode with OLMo-3-32B, the target model produces $1.6\%$ near-verbatim completions, while the reference model already produces $1.2\%$. Since CraneCode is a heavily filtered and rewritten version of Python code that may overlap with other pretraining sources~\citep{olmo2025olmo}, the high reference baseline likely reflects either memorization of near-identical code in the reference model's own training data, or generalization of boilerplate code predictable from context. This underscores why a counterfactual setup is necessary: raw similarity conflates memorization of specific target training samples with these baseline effects, and only the counterfactual gap isolates the former. Applying this criterion, 0.5\% of functions are counterfactually memorized near-verbatim. We provide examples in Appendix~\ref{app:examples_verbatim}.

Other data subsets correspond each to a distinct level of ``difficulty'': for Python functions from Stack-Edu and for all other programming languages, the reference model generates near-verbatim matches at higher rates than for CraneCode, up to 4.3\% for Rust. This suggests that code in these datasets is simpler or already appeared in pretraining, enabling the reference model to reproduce it accurately. Independent of the difficulty, further pretraining exposure still yields higher overlap for the target model and across setups 0.3--1.0\% of samples are counterfactually memorized near-verbatim. 

For the smaller OLMo-3-7B model, which is trained on identical data, the counterfactual memorization is up to 50\% lower across datasets, consistent with prior observations that smaller models retain less training data~\citep{carlini2022quantifying,morris2025much}. Finally, Figure~\ref{fig:memorization_combined} also shows results for Python functions and StarCoder-15B and CrystalCoder-7B. As many factors differ across setups (e.g., model size, training hyperparameters, for 2 epoch-training on Python only for StarCoder), isolating particular effects is difficult. Nevertheless, counterfactual memorization occurs at the same order of magnitude, i.e., 0.4--0.8\% across setups. 

\subsection{Quantifying functional memorization}
\label{sec:results_func}

\textbf{Execution-based similarity.} We instantiate the agentic execution-based testing from Section~\ref{sec:method} to measure functional similarity. Figure~\ref{fig:memorization_combined} illustrates the prevalence of functionally similar generations ($\tau_\text{func} = 0.8$) for the reference and target model, the subset exhibiting counterfactual divergence, and the further subset missed by textual similarity, i.e., $\textsc{SIM}_\text{text}(x, x_T^\star) < \tau_\text{text}$ using BLEU and $\tau_\text{text}=0.75$. As for near-verbatim memorization, reference models already produce functionally similar completions in a non-trivial fraction of cases (up to 6.3\% for OLMo-3-32B on C code), reflecting functions that the model naturally produces through generalization or due to prior exposure during pretraining. Target models consistently yield substantially higher rates of functional similarity, confirming that direct exposure to these functions during further training induces memorization.

We find a substantial rate of counterfactual functional memorization: for OLMo-3-32B, between 1.0\% and 3.4\% of generations are functionally equivalent only for the target model and exhibit low textual overlap with the training data. Figure~\ref{fig:commission_example} illustrates this: a training function encodes hard-coded values for specific cities, and the target model---prompted with only the signature---reproduces the exact same logic in a completely restructured implementation (80\% output match), while the reference model produces unrelated logic (0\% match). The model has internalized the training data's semantics and leaks underlying logic rather than surface text---precisely the kind of leakage execution-based testing can expose. We provide more illustrative examples in Appendix~\ref{app:examples_functional}. 

On aggregate for OLMo-3-32B, counterfactual functional memorization rates match or exceed counterfactual textual memorization. The two auditing strategies are moreover largely \emph{orthogonal}. Of functions flagged by execution-based testing, 92.2\% are not detected by textual similarity: they reproduce training logic but with different structure, variable names, or formatting. Conversely, 66.9\% of functions flagged by textual similarity are not functionally equivalent: they share surface overlap, e.g., boilerplate code, docstrings, or partial matches, without reproducing the underlying logic with sufficient similarity. This underscores the central finding of this work: code models can internalize and leak functional logic that textual similarity alone fails to detect. Since neither method---textual or functional---subsumes the other, a comprehensive memorization audit in code LLMs requires both. 

We further observe functional memorization across datasets and models. Consistent with near-verbatim memorization, Python functions from Stack-Edu are memorized at higher rates than those from CraneCode, likely because they contain simpler code, have greater overlap with pretraining data, or exhibit higher levels of duplication~\citep{carlini2022quantifying,kandpal2022deduplicating}. Across programming languages, we find that Rust and C yield the highest memorization (3.4\%), while Java and Go are memorized roughly half as often (1.7\%), suggesting that language-specific coding conventions and the diversity of training data impact memorization risk. For the 7B counterpart, we find that memorization drops consistently (0.7--2.8\%), again indicating that model capacity plays a role in the ability to retain training examples. Finally, we extend our analysis to fundamentally different training setups, finding that StarCoder-15B and CrystalCoder-7B memorize 0.3--0.5\% of their samples---lower in absolute terms, but confirming that functional memorization also occurs across setups. These lower rates may also reflect the weaker generative capabilities of these models, which produce fewer functionally valid completions overall and thus have less capacity to faithfully reproduce training data logic. We provide examples of counterfactual functional memorization for all programming languages and models in Appendix~\ref{app:examples_function_across_setups}. We further stress-test the consistency of these execution-based results by varying the similarity threshold, and models and agentic harnesses used for execution in Appendix~\ref{app:details_agent}. 

\textbf{Scalable proxies for execution-based testing.} While execution-based testing provides a verified ground-truth of functional memorization, it becomes computationally expensive at scale. In our setup, agentic execution requires on average 2 (for Python) to 5 (for Rust) minutes per function (details in Table~\ref{tab:timing_exec}), which becomes prohibitive when auditing large codebases. We therefore ask:  having already incurred the cost of agentic execution, can we use its results to evaluate whether (computationally) cheaper metrics from the code clone detection literature serve as effective proxies? We instantiate the metrics from Section~\ref{sec:exp_setup} to all Python functions from CraneCode and their generations by the OLMo-3-32B target model and compare the results with the ground-truth execution-based results. 

\begin{figure}[t]
  \centering
  \begin{subfigure}[b]{0.4\linewidth} 
  \includegraphics[width=\linewidth]{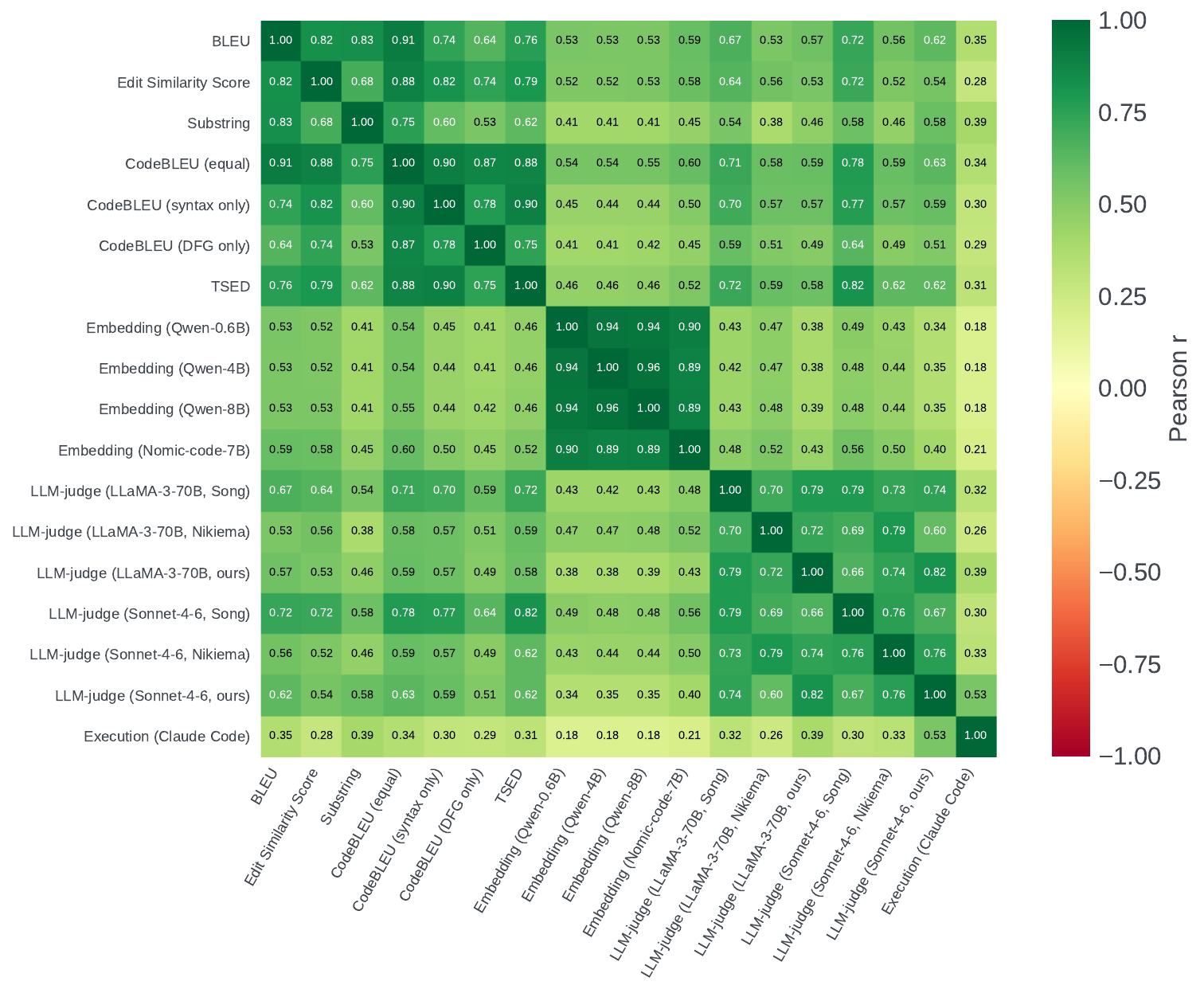} 
  \end{subfigure} 
  \begin{subfigure}[b]{0.35\linewidth} 
  \includegraphics[width=\linewidth]{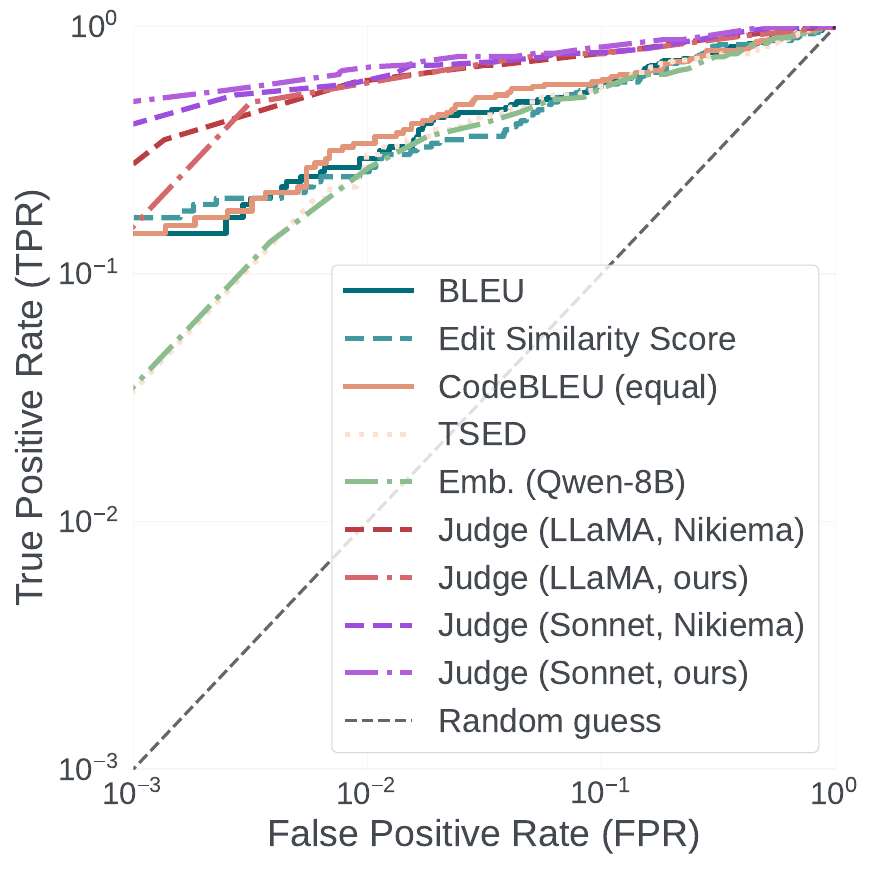} 
  \end{subfigure} 
  \caption{Evaluating a range of metrics to approximate execution-based similarity, for Python (CraneCode) and the OLMo-3-32B target model. (Left) Pearson correlation and (Right) ROC curves with execution-based similarity as a binary label, positive if $\textsc{SIM}_\text{func}(x, x^\star_T) \geq \tau_\text{func}$ and negative otherwise (1.97\% positives). More results in Appendix~\ref{app:add_metrics}.}
  \label{fig:correlation_roc_main}
\end{figure}

We first treat the execution-based functional similarity from Equation~\ref{eq:exec_sim} as a continuous variable in $[0,1]$ and compute the Pearson correlation across all metrics. Figure~\ref{fig:correlation_roc_main}(a) shows how different classes of similarity metrics cluster tightly: text-based metrics yield $0.68\leq r\leq 0.83$, structural metrics $0.75\leq r\leq 0.90$, embedding-based metrics $0.89\leq r\leq 0.96$ and LLM-judges $0.60\leq r\leq 0.82$. Interestingly, embedding-based metrics and LLM-judges show only moderate correlation with text-based metrics ($0.41\leq r\leq 0.72$), suggesting they capture complementary signal. When compared against the execution based testing, all metrics show uniformly low correlations ($r \leq 0.53$). In part, such low correlation stems from the fact that execution-based similarity is substantially more stringent than other metrics, yielding non-zero values in only 18.0\% of the data points. 

To address the sparsity of execution-based similarity scores, we convert them to binary labels: functions achieving $\textsc{SIM}_\text{func}(x, x^\star) \geq \tau_\text{func}$ are labeled as positive, others as negative, yielding 1.97\% positive samples for CraneCode. We then evaluate which metrics most effectively capture the binary flag of execution-based functional equivalence. We plot ROC curves on log-log axes in Figure~\ref{fig:correlation_roc_main}(b) and report ROC AUC and TPR at various low FPR thresholds in Table~\ref{tab:auc_tpr_metrics_python_all_fpr}. 

All metrics achieve reasonably high ROC AUC scores (> 0.80), indicating they capture meaningful signal related to functional similarity. However, important differences emerge when examining TPR at low FPR. Textual similarity metrics provide a reasonable baseline, with TPR values between 0.26 and 0.29 at FPR = 0.01. Intuitively, functions with high textual overlap tend to behave similarly at execution. However, textual metrics miss functionally similar code that differs syntactically and may also incorrectly flag textually similar but functionally distinct code. 

LLM-judges substantially outperform all other categories. The best-performing judge (Sonnet-4.6 with our prompt) achieves AUC of 0.94 and a TPR of 0.68 at FPR = 0.01. This improvement likely stems from the ability of LLMs to evaluate code more holistically, i.e., reasoning through semantic similarities and differences rather than relying on surface-level heuristics such as textual overlap. However, LLM-judges are sensitive to prompt design and model choice. TPR at FPR = 0.01 ranges from 0.45–0.60 for LLaMA-3.3-70B-Instruct and 0.37–0.68 for Sonnet-4.6, depending on the prompt. Sonnet-4.6 achieves slightly higher TPR overall, suggesting that more capable models yield better detection performance, although at greater inference cost. Structural metrics (CodeBLEU, TSED) and embedding-based metrics do not match LLM-judges despite capturing richer representations than text alone. Their TPR at low FPR remains substantially lower, suggesting they lack the precision needed to reliably detect execution-verified similarity in practice.

\subsection{What drives (and mitigates) functional memorization?}
\label{sec:mitigations}

Prior work has demonstrated that duplication in training data increases the risk of memorization~\citep{carlini2022quantifying,kandpal2022deduplicating,shilov2026mosaic}. Model developers also routinely deduplicate their training corpora to remove redundancies and improve utility~\citep{penedo2023refinedweb,kudugunta2023madlad}, e.g., using exact sequence matching~\citep{lee2022deduplicating}, or semantic duplicate detection via pretrained embeddings~\citep{abbas2023semdedup}. As existing studies only examined the link between duplication and textual memorization, we here ask whether a similar relationship holds for functional memorization, and which form of duplication---textual or semantic---drives each type of leakage.

We search for near duplicates of each sample in the respective language subset of Dolmino (OLMo-3's target data). For textual duplicates, we follow prior work~\citep{lee2022deduplicating,penedo2023refinedweb} and use MinHash~\citep{broder1997resemblance} to approximate Jaccard similarity (128-permutations, $n=5$ tokens shingles). For semantic duplicates, we follow~\citet{abbas2023semdedup} and compute cosine similarity using embeddings from Qwen3-Embedding-0.6B~\citep{yang2025qwen3}. We use thresholds of $0.4$ (Jaccard) and $0.9$ (cosine). We compute the number of textual and semantic only (discarding the textual ones) duplicates for functions generated by the target model with (1) high textual overlap (2) high execution-verified functional similarity with low textual overlap, and (3) all others.

\begin{figure}[t] 
\centering 
\includegraphics[width=0.5\linewidth]{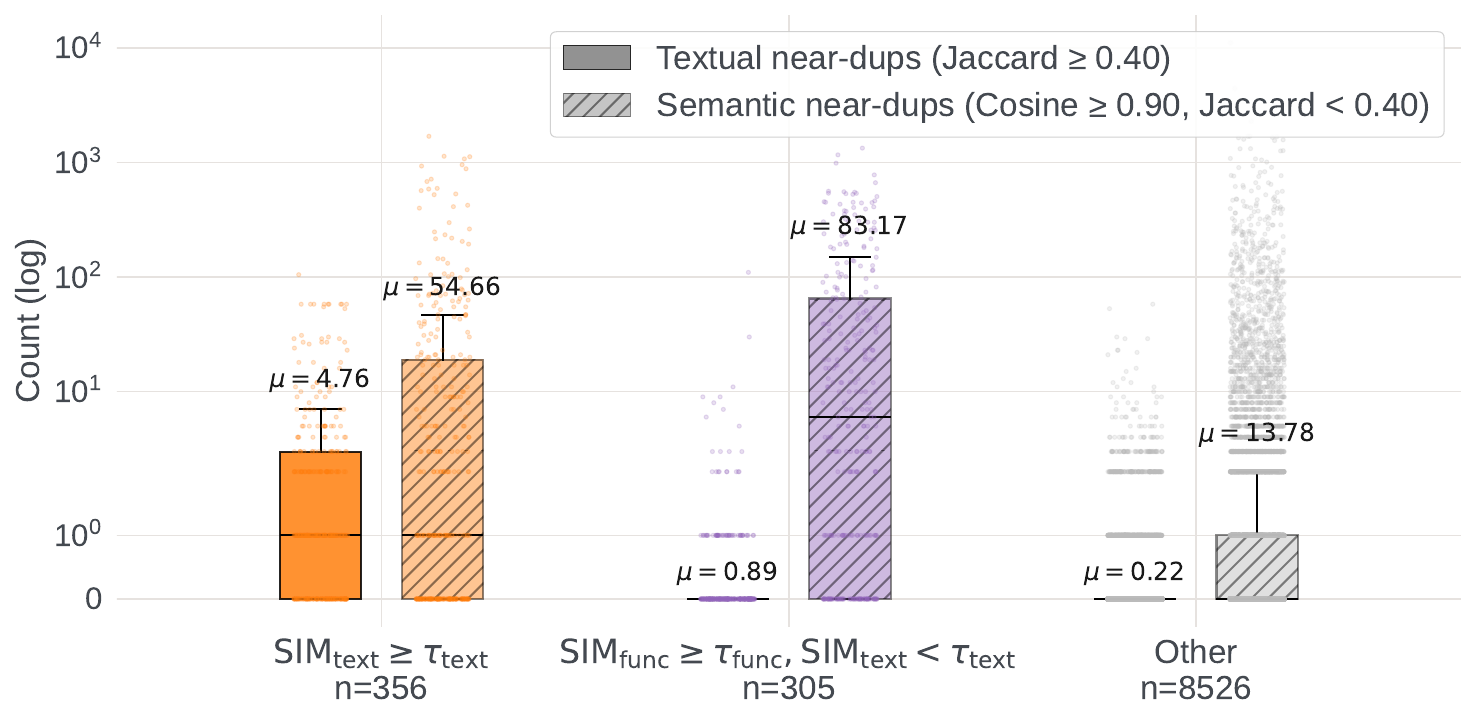} 
\caption{Distribution of textual (Jaccard $\geq0.40$; solid) and semantic (cosine $\geq0.90$, Jaccard $<0.40$; hatched) near-duplicate counts for (i) functions with high textual overlap (BLEU $\geq0.75$), (ii) functional-only similarity ($\textsc{SIM}_\text{func}\geq0.8$, BLEU$<0.75$), and (iii) others. Results for Python and OLMo-3-32B; more results in Appendix~\ref{app:add_duplication}.} 
\label{fig:duplicates_python} 
\end{figure}

Figure~\ref{fig:duplicates_python} shows the distribution of near-duplicate counts for Python (OLMo-3-32B). Functions with textually similar model outputs have substantially more textual near-duplicates ($\mu=4.76$) compared to the baseline ($\mu=0.22$), consistent with prior findings~\citep{carlini2022quantifying,kandpal2022deduplicating}. The functionally similar group, conversely, has the highest semantic near-duplicate count ($\mu=83.17$), exceeding both the textually memorized group ($\mu=54.66$) and the baseline ($\mu=13.78$), while having fewer textual duplicates ($\mu=0.89$). To formalize this, we fit a joint logistic regression predicting textual and functional similarity from both duplicate counts, i.e., $\mathbbm{1} [\text{SIM} \geq \tau] \sim \beta_{\text{text}} \cdot \text{TextualDups} + \beta_{\text{sem}} \cdot \text{SemanticDups}$. For Python, textual similarity is predominantly predicted by textual near-duplicates ($\beta_{\text{text}} = +0.68$, $p<10^{-24}$) with semantic duplicates non-significant ($p=0.07$). Conversely, functional similarity is predicted by semantic near-duplicates ($\beta_{\text{sem}}=0.17$, $p<10^{-3}$) with textual duplicates barely significant ($p=0.046$). This dissociation holds across all five languages (full results in Appendix~\ref{app:add_duplication}).

Figure~\ref{fig:semantic_duplicates} illustrates the closest semantic duplicates of the function in Figure~\ref{fig:commission_example} and clearly shows that all encode similar logic in diverse implementations, reinforcing abstract internalization rather than verbatim recall. This aligns with~\citet{allen2024physics}, who show that knowledge augmentation during pretraining pushes LLMs toward abstract representations. While this is desirable for general knowledge retention, the same mechanism can increase the risk of functional leakage when training on confidential code. From a practical standpoint, these findings suggest that textual deduplication alone is insufficient to mitigate functional memorization and that semantic deduplication~\citep{abbas2023semdedup} is likely necessary as a complementary preprocessing step.

Finally, if auditing exposes leakage---whether or not deduplication was applied---practitioners may seek to suppress it post hoc. In Appendix~\ref{app:add_posttraining}, we evaluate NPO~\citep{zhangnegative} and DPO~\citep{rafailov2023direct} as post-training techniques to mitigate the leakage of memorized functions while retaining utility. All methods incur some utility cost, and no method is uniquely suited to textual versus functional memorization. DPO with functionally equivalent rewrites as preferred completions yields the best trade-off, as providing a positive signal better preserves code generation than merely pushing away from memorized outputs.

%% file: sections/6_related_work.tex
\textbf{Training data extraction.}~\citet{carlini2021extracting} first demonstrated that LLMs can reproduce verbatim spans from their training data.~\citet{nasr2025scalable} later distinguished \emph{extractable} memorization, where a target sequence can be elicited from any prompt, from \emph{discoverable} memorization, where reproduction occurs when prompted with the exact training prefix. Subsequent work argued that exact matching under greedy decoding can be overly conservative, and propose approximate criteria based on metrics such as BLEU or edit distance to capture \emph{near-verbatim} leakage~\citep{ippolito2023preventing}. Other work emphasized that deployment-time generation is typically stochastic rather than greedy, and proposed measuring probabilistic extraction~\citep{hayes2025measuring}. Recently,~\citet{barbero2025extracting} studied the extraction of alignment data from post-trained models, and argue that computing the distance in embedding space between generations and training data reveals semantic leakage beyond what string-based metrics may capture. 

\textbf{Code extraction.} Several works study extraction specifically for models trained on code. Some follow setups from natural language and sample generations from an empty prompt and match them against the training data~\citep{yang2024unveiling}, or complete training prefixes and measure (near-)verbatim overlap with the ground truth~\citep{al2024traces,salerno2025much}. Other works specifically target the extraction of \emph{secrets} present in code, such as API keys, or PII~\citep{nie2025decoding,huang2024your,niu2023codexleaks}.
While most work focuses on pretraining, some also studies memorization during post-training using RLHF~\citep{hayes2024measuring}, fine-tuning to specific tasks~\citep{salerno2025much} or to niche programming languages~\citep{wang2025verileaky}. Beyond privacy and IP leakage, memorization has also been studied as a source of contamination in coding benchmarks~\citep{riddell2024quantifying,zhang2025memorize}; notably, OpenAI recently retracted SWE-Bench Verified after finding that frontier models reproduce gold patches (near-)verbatim~\citep{openai2025swebench}.

\textbf{LLM memorization.} A large body of work has characterized LLM memorization beyond data extraction. One line of research studies membership inference attacks (MIAs), which determine whether or not a particular data point was used during training~\citep{carlini2021extracting,shidetecting,mattern2023membership,meeus2024did,meeus2025sok,duanmembership,hayesexploring,shilov2026mosaic}. Other approaches have examined memorization through ``compression'', quantified in the length of an adversarial prompt required to generate a sequence~\citep{schwarzschild2024rethinking} or as the reduction in bits needed to describe a record when a model trained on it is available~\citep{morris2025much}. Another line focuses on counterfactual memorization~\citep{zhang2023counterfactual}, computing the expected change in a model’s loss on a target example when that example is included versus excluded from the training set, extending the label memorization for classification from~\citet{feldman2020does} to language models. While these approaches provide useful signals about memorization, we here focus on data extraction, as it more directly reflects the risk of leakage from deployed models in practice.

%% file: sections/7_discussion.tex
\textbf{Limitations.} While execution-based testing provides a principled approach to detecting functional leakage, it has several limitations. First, functional equivalence, just like textual similarity, is not binary: a generation reproducing only a subset of execution paths may still constitute meaningful leakage even if outputs diverge on most inputs. While we argue that our thresholds ($C=80\%$ line coverage, $N_\text{in}=8$ input states and $\tau_\text{func}=0.8$; ablations in Figure~\ref{fig:threshold_sensitivity}) are reasonable, there is likely no single correct choice and future work should further stress-test their sensitivity. Similarly, generated code may have superficial differences in output formatting or contain minor bugs that prevent execution matches despite leaking logic. Finally, the agentic harness, just like LLM judges, introduces non-determinism, adding noise to measurements.

%Our experiments further focus on open-source code, while the most pressing concern is the memorization of proprietary code. In this setting, the dynamics of memorization may differ, e.g., proprietary functions are likely more distinctive, potentially leading to stronger memorization and/or higher leakage~\citep{meeuscanary,mitchell2026advancing}. 

Lastly, our counterfactual is not a perfect leave-one-out version of the target. The target model's training set exceeds the reference model's by more than just the studied samples and may include data that also improves the target's coding ability, suggesting that some detected ``leakage'' could reflect improved generalization. We mitigate this by prompting the target model with only the preceding code and signature (no explicit behavior description) and filtering for functions with meaningful logic, where faithful reproduction remains relevant to study regardless of whether we consider a perfect leave-one-out counterfactual. 

\textbf{Future work.} Several directions remain open. First, studying code memorization beyond pretraining: at which stage is code most vulnerable, how should one prompt instruction-tuned models for auditing, and how does post-training affect memorization of pretraining data? Second, future work should develop methods for detecting functional memorization at scale, including better proxy metrics, fine-tuned judges, functional ``canaries''~\citep{carlini2019secret,meeuscanary}, or through characterizing vulnerable code patterns. Third, on the mitigation side, future work should study optimal thresholds for semantic deduplication~\citep{abbas2023semdedup}, develop post-training methods tailored to functional memorization, and stress-test post-hoc mitigations against re-learning~\citep{hu2025unlearning} and adversarial extraction~\citep{hayes2025inexact}.

%In conclusion, our findings do not preclude training on proprietary code if it meaningfully improves utility. 

In conclusion, our findings motivate the need for adopting auditing beyond verbatim detection, and employing mitigation techniques that also account for functional memorization, such as semantic training data deduplication.

%% file: appendix/filtering.tex
\begin{figure}[t]
\centering 
\begin{subfigure}[b]{0.3\linewidth} 
\includegraphics[width=\linewidth]{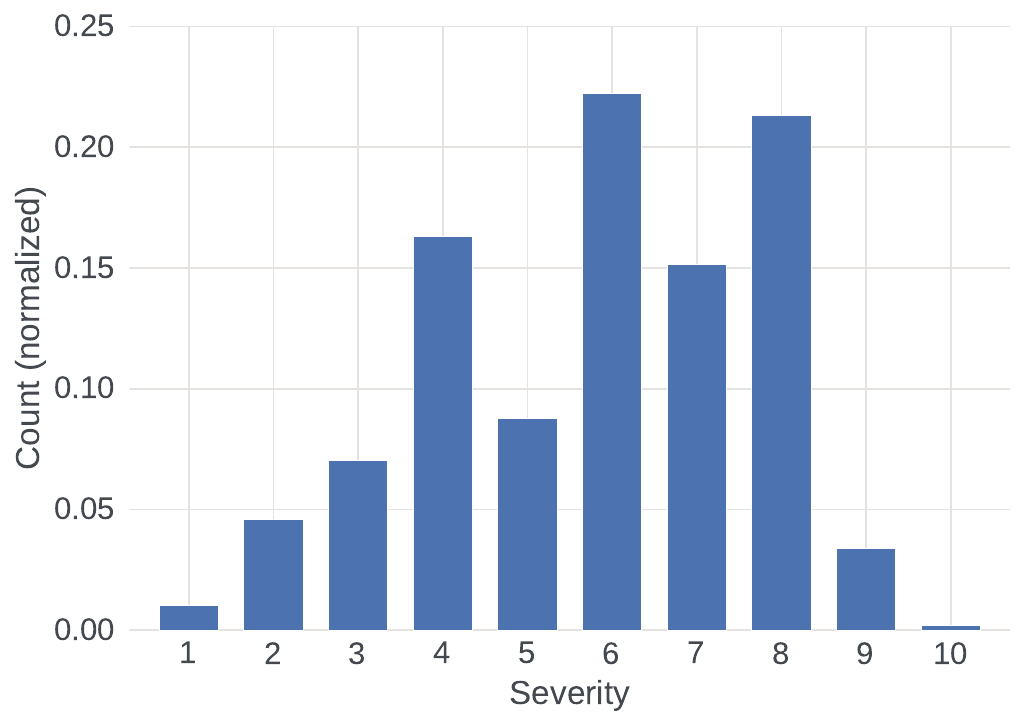} \caption{Python (OLMo-3, CraneCode)} 
\end{subfigure} 
\begin{subfigure}[b]{0.3\linewidth} 
\includegraphics[width=\linewidth]{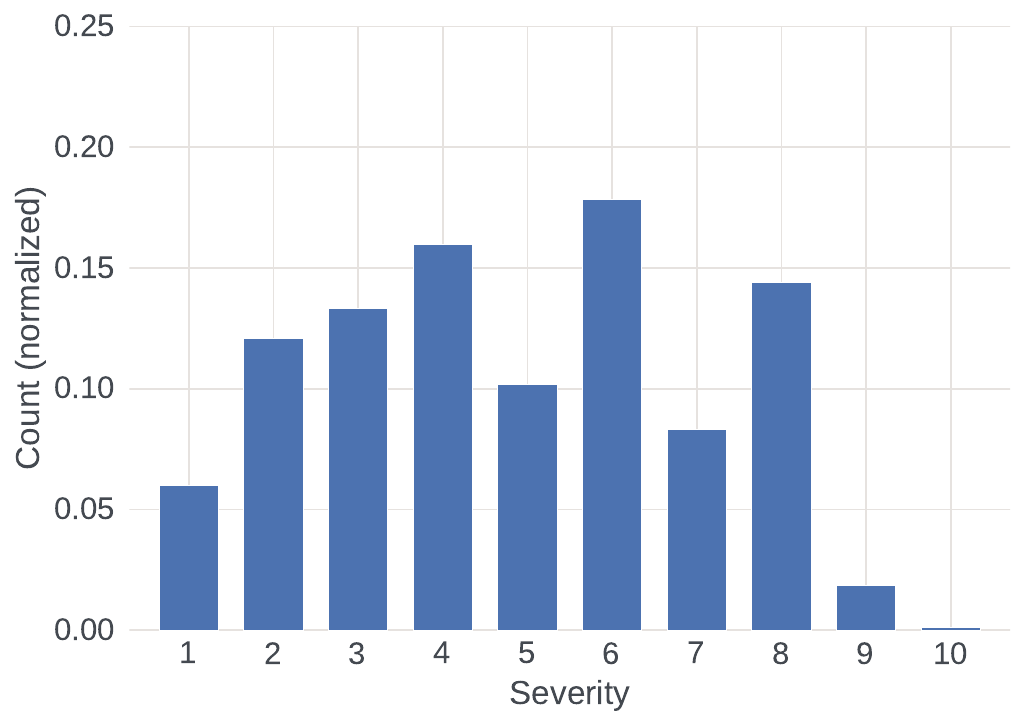} \caption{Python (OLMo-3, Stack-edu)}
\end{subfigure} 
\begin{subfigure}[b]{0.3\linewidth} 
\includegraphics[width=\linewidth]{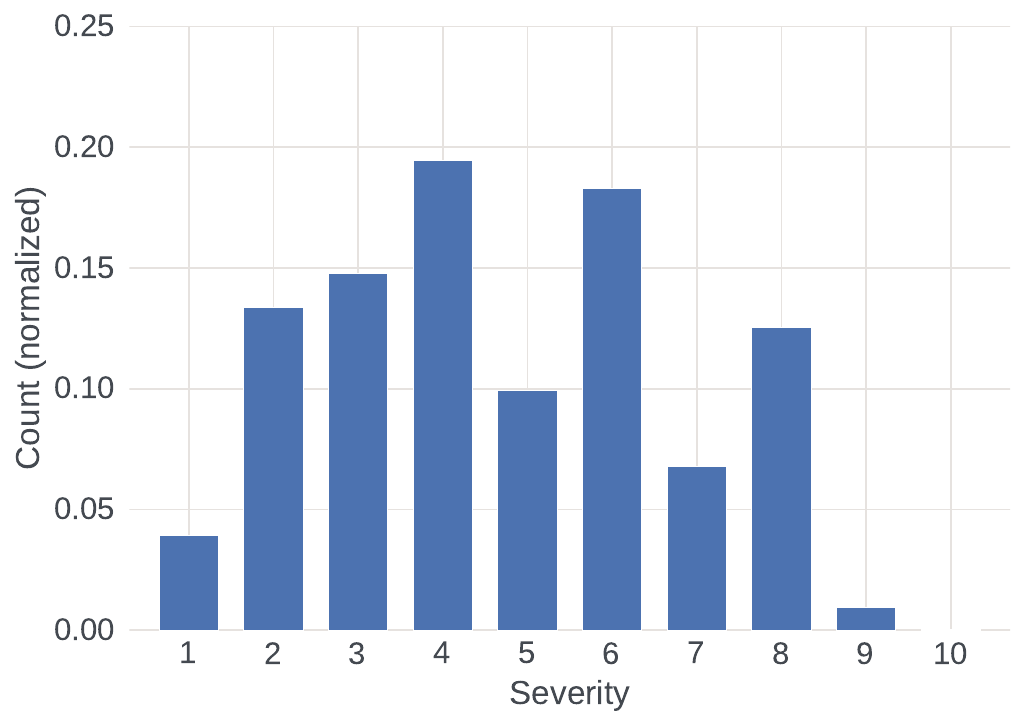} 
\caption{C (OLMo-3)} 
\end{subfigure} 

\vspace{0.5em} 

\begin{subfigure}[b]{0.3\linewidth}
\includegraphics[width=\linewidth]{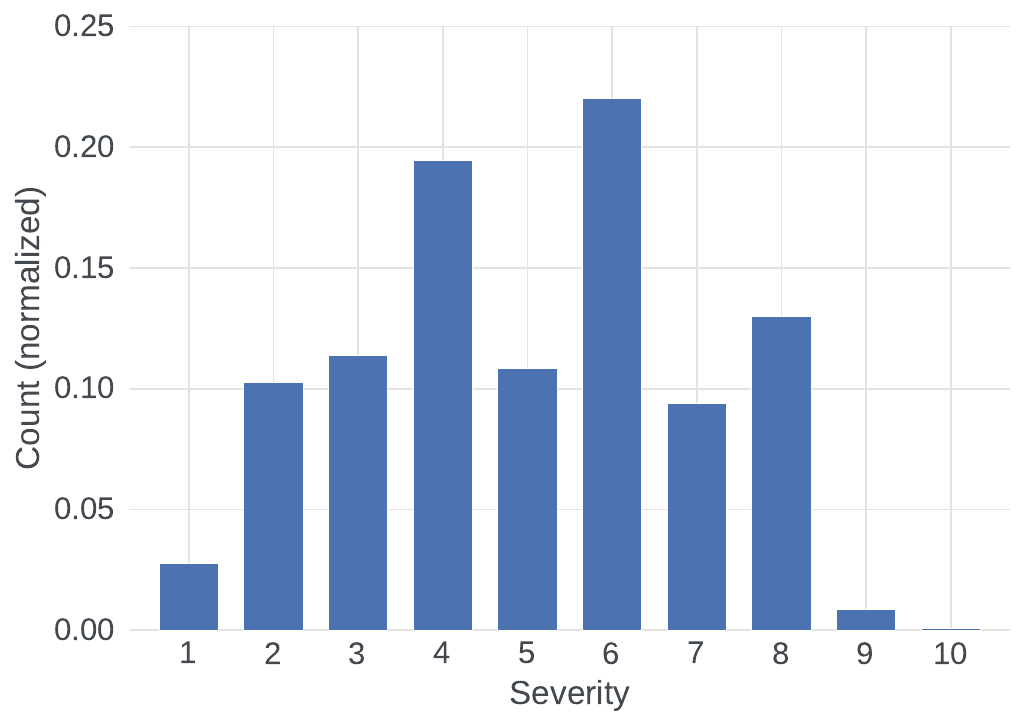} \caption{Java (OLMo-3)} 
\end{subfigure} 
\begin{subfigure}[b]{0.3\linewidth} 
\includegraphics[width=\linewidth]{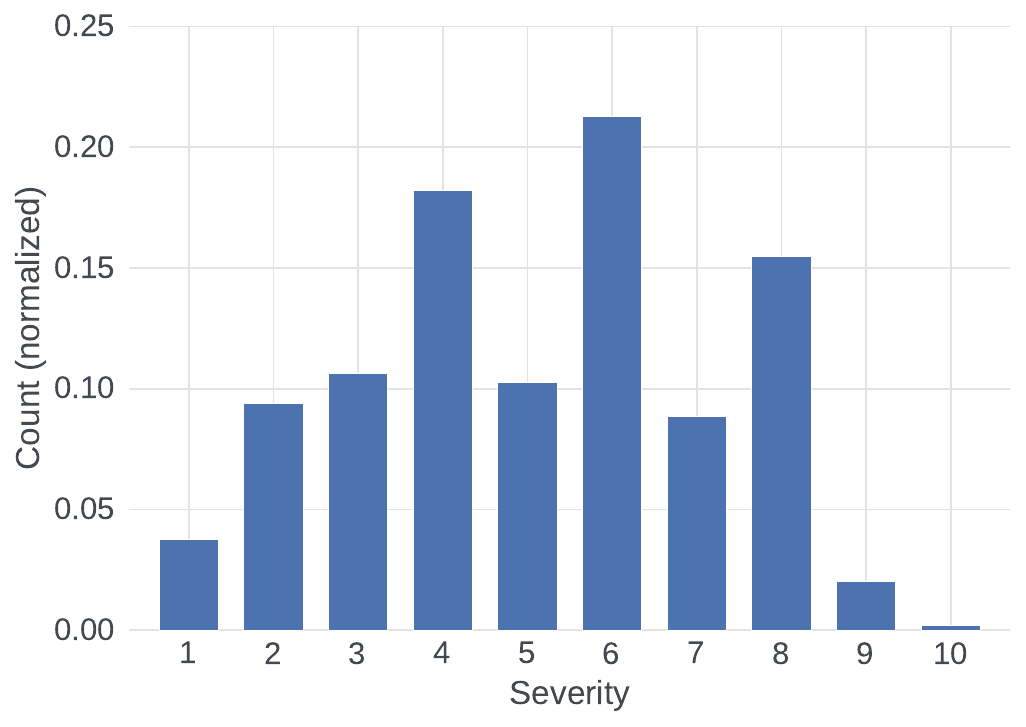} \caption{Rust (OLMo-3)} 
\end{subfigure} 
\begin{subfigure}[b]{0.3\linewidth}
\includegraphics[width=\linewidth]{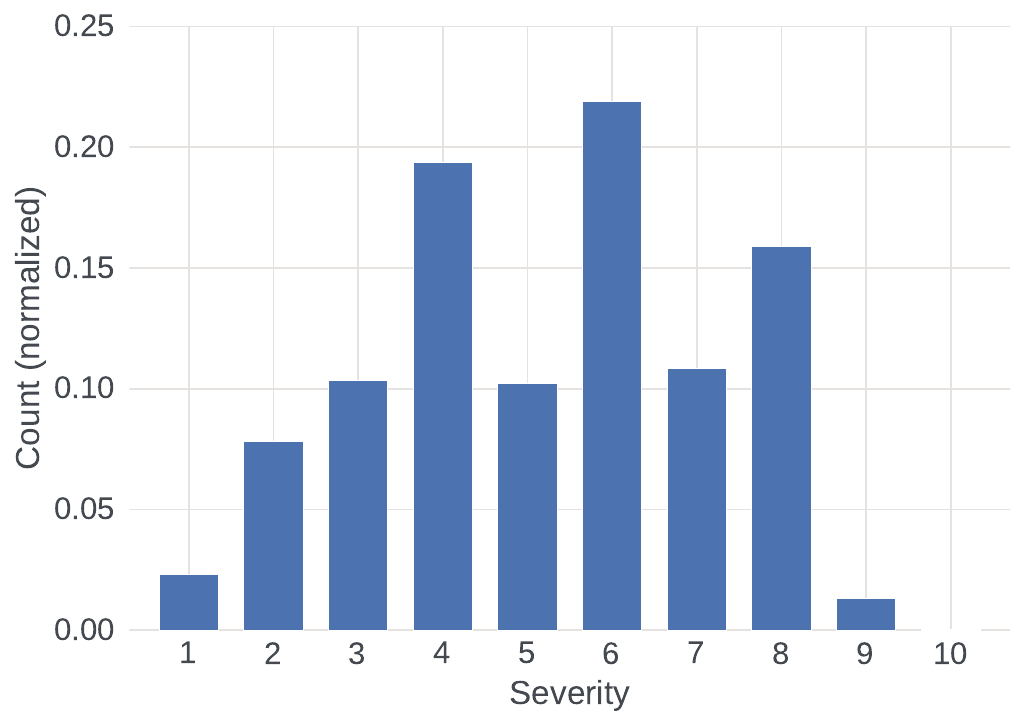} 
\caption{Go (OLMo-3)} 
\end{subfigure} 

\vspace{0.5em} 

\begin{subfigure}[b]{0.3\linewidth}
\includegraphics[width=\linewidth]{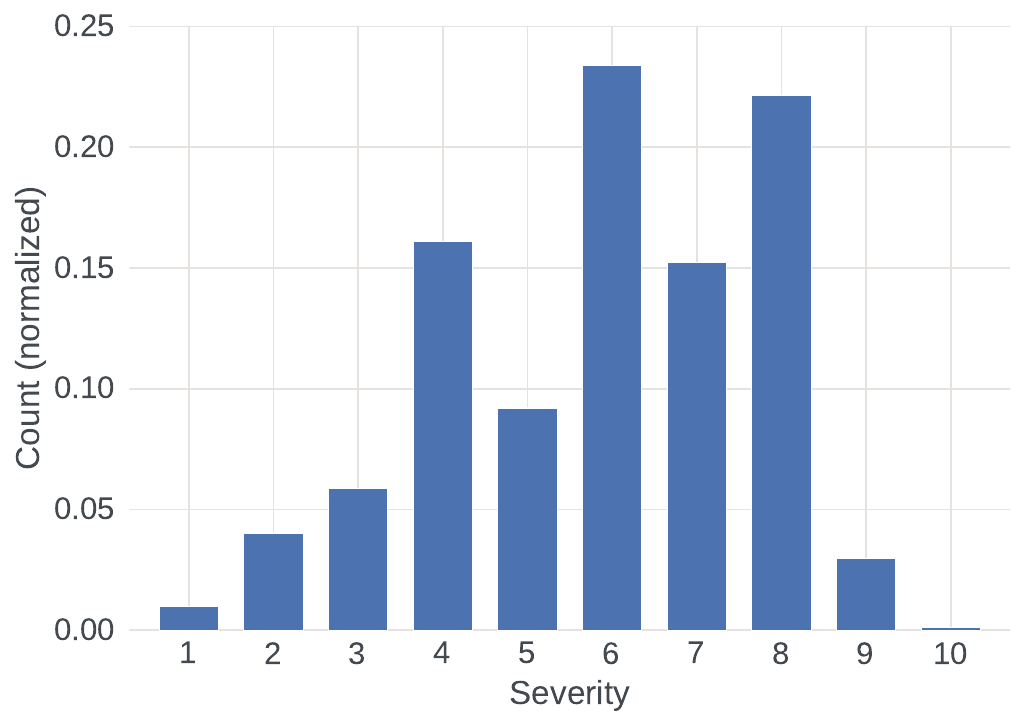} 
\caption{Python (StarCoder-15B)} 
\end{subfigure} 
\begin{subfigure}[b]{0.3\linewidth}
\includegraphics[width=\linewidth]{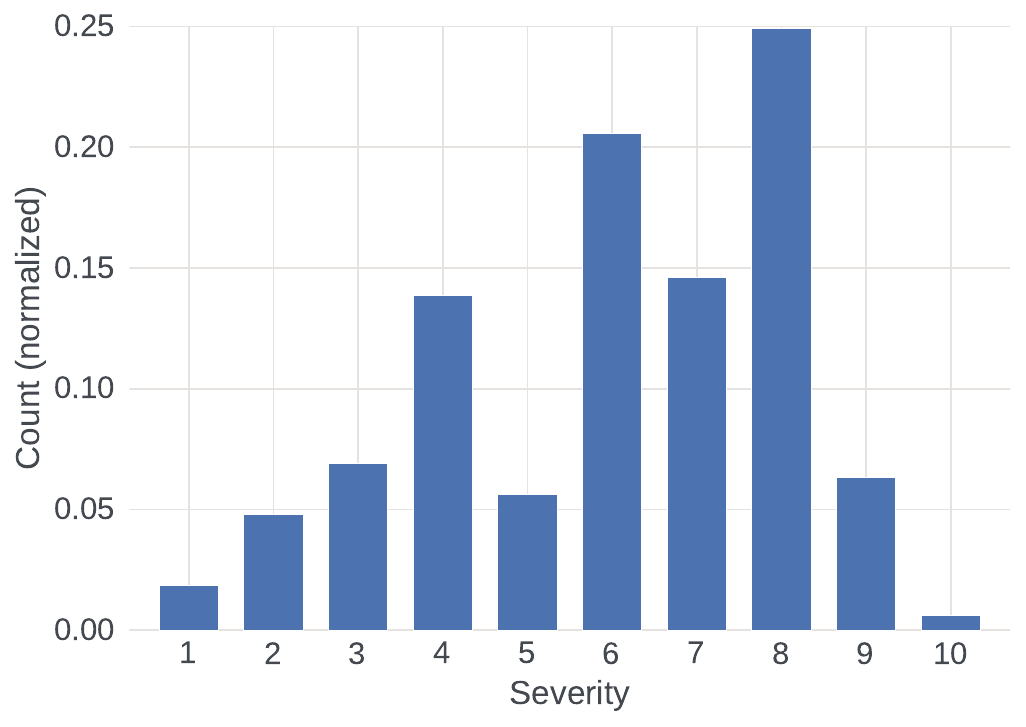} 
\caption{Python (CrystalCoder-7B)} 
\end{subfigure} 
\caption{Distribution of severity labels as classified by LLaMA-3.1-8B-Instruct across datasets used throughout this work.} 
\label{fig:severity_distribution} 
\end{figure}

To identify functions which contain meaningful logic from the training data, we instantiate an LLM-as-a-judge (LLaMA-3.1-8B-Instruct~\citep{grattafiori2024llama}) to classify each of the randomly sampled functions. We first filter for functions with body lengths between 10--50 lines and at least 300 characters. We then provide the model with both the full function of interest itself and the source file in which it originally appears (truncated up to its first 8,000 characters). The prompt is illustrated below. For all data subsets, we randomly sample 100k functions and find that 7-10\% are classified as containing meaningful functional logic. For each subset we keep a random set of 5,000 functions for subsequent analysis, except for Rust and Go, for which we include all data for classification and only retain 2,612 and 3,735 functions, respectively. Figure~\ref{fig:severity_distribution} visualizes the corresponding distribution of severity as returned by the judge model. 

\begin{promptbox}
\begin{Verbatim}[fontsize=\scriptsize]
You are a code analysis tool that detects functional/semantic leakage in individual __LANGUAGE__ functions.

Your task: determine whether the **target function** (delimited below) contains meaningful **proprietary
functional logic** - business rules, decision logic, or algorithms that reveal how a specific, identifiable
real-world product or service works internally.

You will be given:
1. The **full file** for context (imports, classes, other functions, etc.)
2. The **target function** to classify

Your classification must be about the **target function only**, not the file as a whole. Use the file context
to better understand what the function does and whether it is proprietary and meaningful.

Respond ONLY with a single raw JSON object with exactly three keys: label, severity, reasoning. No
markdown fences, no extra text.

## What IS functional leakage (label: YES)

The function must reveal **meaningful logic or functionality** of an identifiable real-world system, product
or company. Examples:
- Internal business rules or decision logic of a specific company or product (e.g., pricing formulas, fraud
rules, ranking algorithms, content moderation, safety control)
- Proprietary algorithms that are not published or publicly documented (e.g., ad ranking algorithms, content
moderation logic, ML feature engineering, data scraping methodologies)
- Production configuration, tuning parameters, or operational thresholds with real-world value (e.g., PID gains
for a specific vehicle, rate limits for a specific service)
- Internal API schemas, authentication flows, or access control logic for a non-public system
- Trade secrets: strategies, formulas, or processes that give competitive advantage

## What is NOT functional leakage (label: NO)

- Implementations of known algorithms or techniques from textbooks or papers, regardless of parameters
- Competitive programming, LeetCode, HackerRank, Advent of Code, Project Euler solutions
- Tutorial, educational, homework, or coursework code
- Standard ML/DL training pipeline functions (data loading, training loops, evaluation, etc.)

**Key decision criterion:** Does this function reveal internal logic that the owner would reasonably expect
to remain confidential?
Would a competitor, regulator, or journalist find this function informative about how the system actually works?

## Severity

If YES, assign severity 1-10, indicating how severe the leakage of this internal logic would be for the owner.
- **1-3**: Low. Borderline or mildly specific logic.
- **4-6**: Moderate. Meaningful insight into system internals.
- **7-8**: High. Competitively or operationally sensitive.
- **9-10**: Critical. Core trade secrets. Should be very rare.

If NO, severity is null.

## Output format

Always return exactly this structure:
{"label": "YES", "severity": <int 1-10>, "reasoning": "<1-2 sentences>"}
or
{"label": "NO", "severity": null, "reasoning": "<1-2 sentences>"}
\end{Verbatim}
\end{promptbox}

%% file: appendix/add_verbatim_results.tex
\begin{figure}[t]
  \centering
  \begin{subfigure}[b]{0.49\linewidth}
    \includegraphics[width=\linewidth]{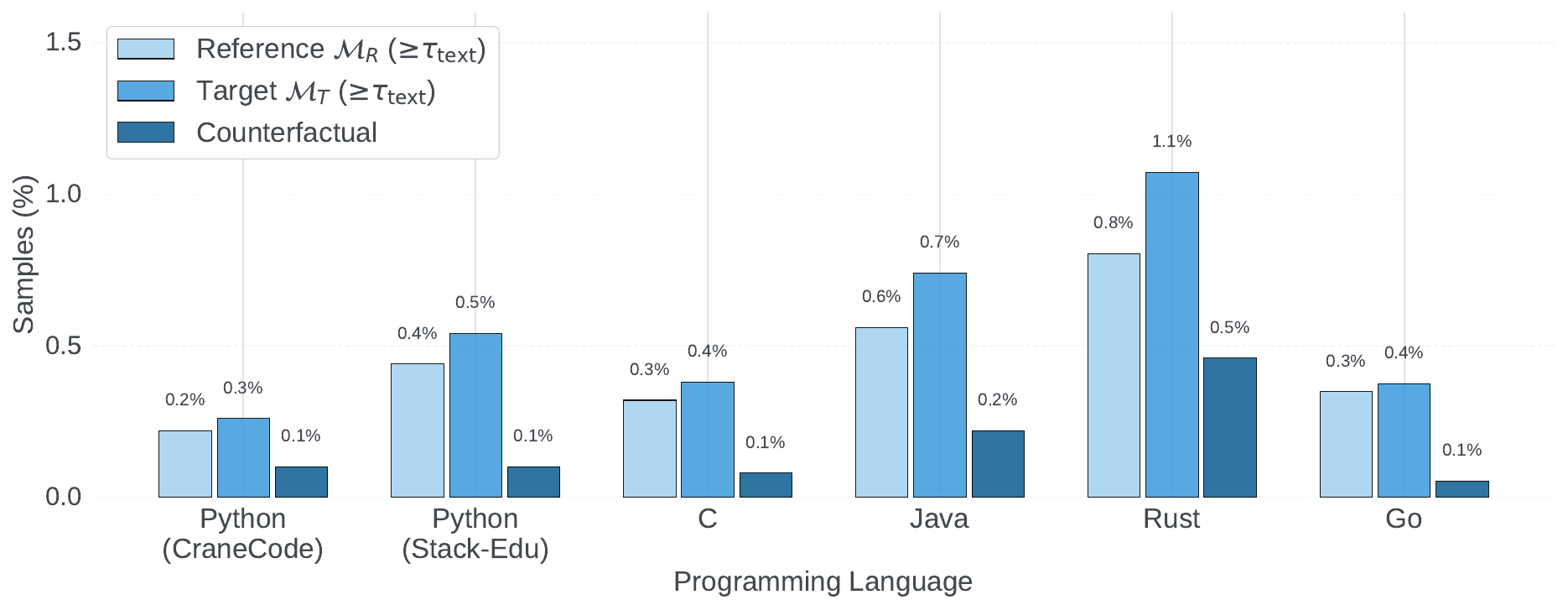}
    \caption{OLMo-3-32B across programming languages}
  \end{subfigure}
  \begin{subfigure}[b]{0.49\linewidth}
    \includegraphics[width=\linewidth]{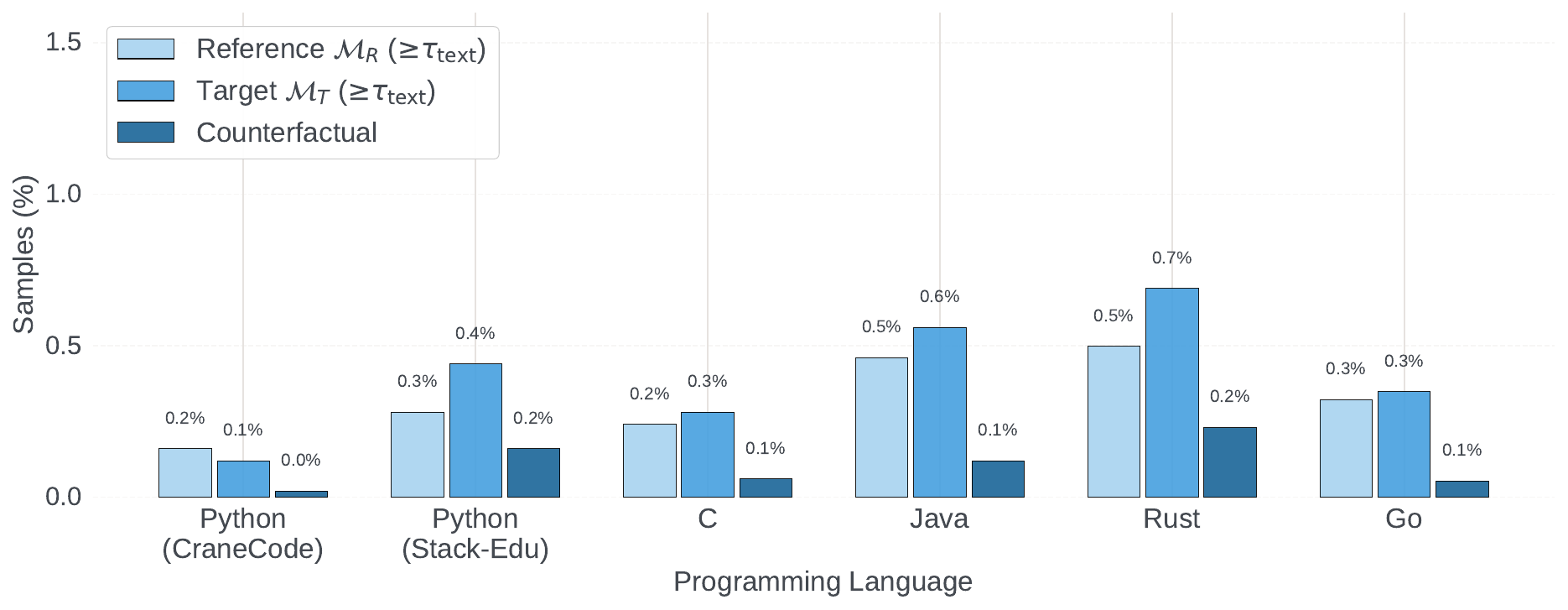}
    \caption{OLMo-3-7B across programming languages}
  \end{subfigure}
  \begin{subfigure}[b]{0.49\linewidth}
    \includegraphics[width=\linewidth]{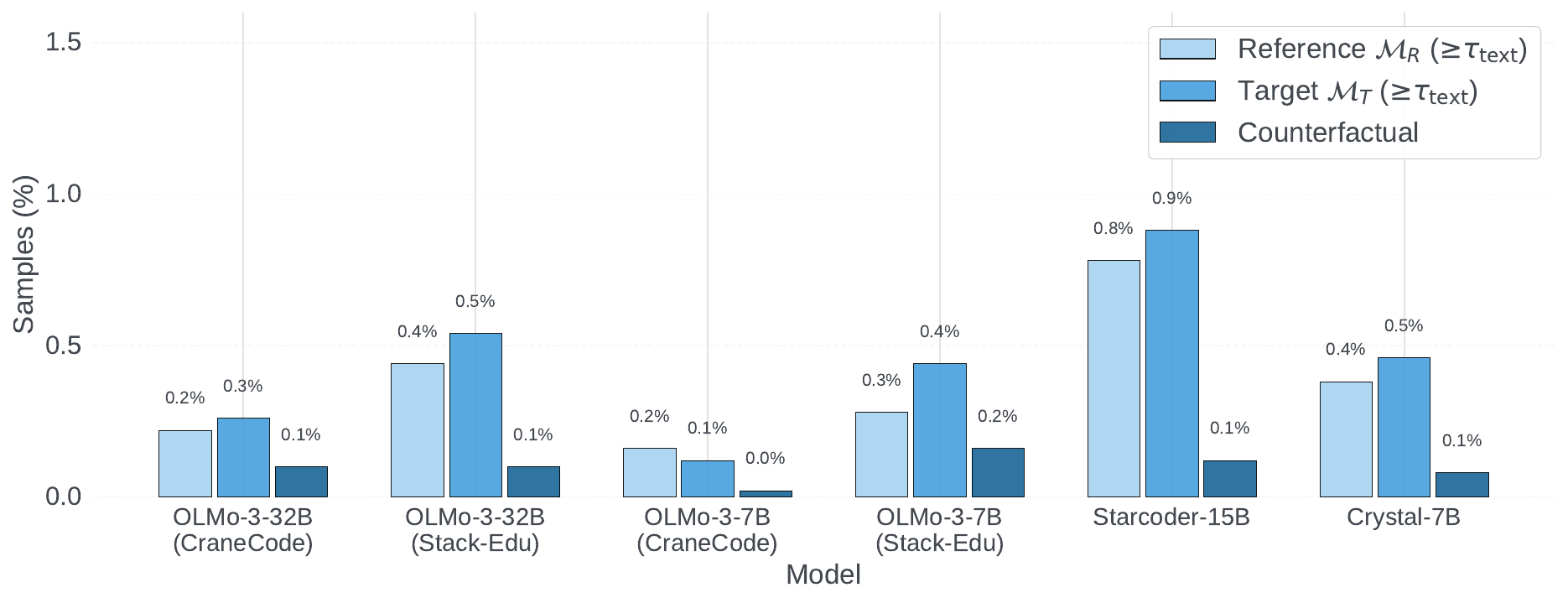}
    \caption{Python across models and training setups}
  \end{subfigure}
  \caption{Exact memorization across models and programming languages. We report the fraction of functions that were generated as exact matches. Results for the reference model, target model and the fraction of samples that are counterfactually memorized for (a) OLMo-3-32B and (b) OLMo-3-7B across programming languages and datasets, and (c) Python functions across models, including StarCoder-15B and CrystalCoder-7B.}
  \label{fig:verbatim_mem_exact}
\end{figure}

Figure~\ref{fig:verbatim_mem_exact} shows the exact memorization rates across datasets and models, complementing the results from Figure~\ref{fig:memorization_combined} (effectively considering BLEU with $\tau_\text{text}=1$). 

%% file: appendix/details_execution.tex
As discussed in Section~\ref{sec:method} and Section~\ref{sec:exp_setup}, we instantiate one independent AI coding agent per sample, consisting of the training data function and the target and reference model generations.

\textbf{Implementation details.} For each sample, we invoke a coding agent in headless mode via the command-line interface with a maximum of 50 tool-use turns per sample. Each agent is spawned as an independent process with a timeout of 900 seconds, with up to 30 agents running in parallel. The execution environment runs on Linux with Python 3.10.19, GCC 11.5.0, Rust 1.94.1, JDK 1.8.0, and Go 1.26.2. Common Python packages (NumPy, Pandas, PyTorch, SciPy) are pre-installed; the agent can install additional packages via pip as needed. The agent has full filesystem access to a working directory and can create, execute, and clean up temporary files. 

When computing execution output equivalence, we require exact matches with floating-point tolerance and set random seeds for functions involving non-deterministic operations. When the observable output state for $x^\star$ does not match exactly with the one for $x$ for the same input (or execution fails for $x^\star$ but succeeds for $x$), the corresponding input is counted as a mismatch. We provide the full prompt used to instantiate the coding agent below. 

\clearpage

\begin{promptbox}
\begin{Verbatim}[fontsize=\tiny]
# Execution-based Functional Equivalence Assessment
You are tasked with determining whether generated code functions are functionally equivalent to a training function — i.e., whether substituting one for the other produces the
same observable output state for any given input state.   

## Context
You are given a train function (from training data) and two generated functions: a target function and a reference function (both produced by language models given the same function 
signature as a prompt). All three share the same signature. Your job is to assess the degree of functional equivalence between each generated function and the train function by executing them 
on diverse input states and comparing observations of the output states.

## Input Format
- `LANGUAGE`: The programming language (Python, Java, C, Rust or Go)
- `DATA_PATH`: Path to a pickle file containing a list of samples

Each sample is a dictionary with:
- `prompt`: The function signature (shared by all three functions)
- `train_text`: The body of the train function
- `generated_text_target`: The body of the target function
- `generated_text_reference`: The body of the reference function
- `file_text`: The full source file the train function came from (imports, helpers, classes, constants)

## Procedure
Process each sample sequentially. Under no circumstances do you reuse a generic harness across all samples. Every sample requires substantial and tailored effort. 

### Step 1: Analyze the Function
Read both F_train = `prompt + train_text` and `file_text` for context (imports, classes, helpers). 

Identify:
1. All observable behaviors: return value, mutations to `self` or mutable arguments, external calls (e.g., `requests.post`, `cursor.execute`, `file.write`), output (print/logging), exceptions raised.
2. Required input states: function arguments, environment variables, files, `self` state.
3. Execution dependencies: imports, APIs to mock, helper functions to stub.
4. Observation targets: what to record to distinguish behavior across branches — return values, file contents, API calls, stdout.

### Step 2: Design Input States
Design N=10 input states that together cover the function's behavioral space. An input state is a complete specification of the 
execution environment: parameter values, mock object states, input files, environment variables, helper configurations.

Coverage principles:
- Trace every `if`/`elif`/`else` branch in F_train. Design at least one input state per branch.
- Use realistic, diverse and complex inputs (e.g., non-trivial arrays, not just empty or `[0,1]`). 
- Test edge cases: empty collections, boundary values, None.
- For error-handling paths (`try`/`except`, validation checks), design inputs that trigger them.

Do not examine F_target or F_reference when designing input states. Derive them solely from F_train and `file_text`.

### Step 3: Build the Execution Scaffolding
Your goal: create an environment where F_train can execute to completion while you record every observable behavior. Build this scaffolding once, then swap only the function under test.

Environment (use these, no exceptions):
- For Python: always use the conda env at `__CONDA_ENV__`. Always invoke the drive and helper scripts with this exact path. Includes most common packages like numpy, pandas, scipy, 
torch, etc. Install missing packages with `pip install <pkg>`. 
- For compiled languages: use `gcc`/`javac`/`rustc` from the same conda env bin directory. Before marking a sample inconclusive, (for C) check whether the required struct/typedef definitions and 
any observable post-call state can be reconstructed from file_text, and only skip if they genuinely cannot and (for Java) invest effort in iterative stub-generation from javac errors.
- Use `file_text` as context (excluding the train function itself) to get imports, classes, helpers, and constants. 

Mock objects and states
- For `self` parameters: Build a real-enough object from the class definition in `file_text` if available. If the class can't be instantiated, create a mock with the exact attributes and methods F_train
accesses. Methods that F_train calls on `self` should be recording mocks — they return plausible values AND log that they were called with specific arguments.
- For external calls (HTTP, DB, filesystem, APIs): Patch them with recording mocks that return plausible values. For example, `requests.get` returns a mock Response with `.status_code=200` 
and `.json()` returning a dict. 
- For file I/O: Understand the expected file format from context and create representative, realistic content.
- For missing helper functions from `file_text`: Implement lightweight stubs that return the types F_train expects, while recording calls.

Critical: recording mocks must produce realistic return values. A mock returning `None` for everything collapses all branches into one path. Read F_train's body to understand what return values 
are needed to exercise different branches. If you see uniform observations across distinct inputs, your mocks are the cause, so fix them. 

### Step 4: Define the Observation Protocol
A function's behavior is everything observable from outside: return value, argument mutations, attribute writes, external calls, files written, exceptions raised, printed output. 
Two functions are behaviorally equivalent if a caller cannot distinguish them.

1. Return value (even if `None` — returning `None` vs raising an exception is a real difference)
2. Exception type and message (if any)
3. Calls to mocked methods: for each mock, record `(method_name, args, kwargs)` in order
4. Mutations to `self`: snapshot relevant `self` attributes before and after
5. Mutations to mutable arguments: snapshot before and after
6. Captured stdout/stderr
7. Other side effects: API requests sent, files written, database writes

Construct an observation tuple from these to reliably and holistically capture what happened during the function execution. Two executions are equivalent iff their observation tuples match.

### Step 5: Execute and Compare
For each input state S_j (j = 1..N):

1. Set up fresh scaffolding for S_j
2. Execute F_train(S_j) → record observation O_train_j
3. Set up identical fresh scaffolding for S_j
4. Execute F_target(S_j) → record observation O_target_j
5. Set up identical fresh scaffolding for S_j
6. Execute F_reference(S_j) → record observation O_reference_j

Execution rules:
- Reset all mocks and state between F_train/F_target/F_reference — each gets a pristine copy
- Set `random.seed(42)` before each execution
- Use exact equality for discrete observations, approximate equality (rtol=1e-5) for floats/arrays
- If F_target or F_reference crashes on an input state where F_train succeeds, that is a valid input and a meaningful behavioral difference (record it)
- If F_train crashes on an input state, that input state is invalid — discard it

An input state is valid if F_train executes without crashing. You need at least 3 valid and distinct input states with at least 3 distinct observation tuples to make a conclusive judgment.

### Step 6: Validate Observation Diversity
After running F_train on all input states, check: do you have at least 3 valid and distinct input states with at least 3 distinct observation tuples? If all observations are too uniform, 
your mocks are collapsing branches. 

If the diversity is insufficient, iterate carefully. Go back to Step 2. Redesign your input states, mock objects, files, such that different branches of the functions are exercised and the function’s 
behavior is tested holistically. 

This is very important — you need to iteratively spend substantial effort until you get sufficient evidence of the function input-output behavior. You must iterate at least 3 times (fixing 
scaffolding, designing new input states, recording the output state). If after 3 iterations for one sample you still have insufficient diversity, mark as inconclusive.
\end{Verbatim}
\end{promptbox}

\begin{promptbox}
\begin{Verbatim}[fontsize=\tiny]
### Step 7: Compute Scores
```
n_valid = number of input states where F_train executed without crashing
n_distinct_observations = number of unique output observations out of the n_valid input states
coverage_train = line coverage of F_train's body across all valid input states, measured via `coverage.py` (`coverage run` + `coverage json`). Report as a float between 0 and 1.
S_target = (input states where O_train == O_target) / n_valid
S_reference = (input states where O_train == O_reference) / n_valid
```

### Step 8: Produce the Result
For each sample, produce:

```json
{
   "results": [
       {
           "input state": "<description of input state>",
           "observations_train": "<observation tuple>",
           "train_ok": true,
           "observations_target": "<observation tuple>",
           "target_ok": true,
           "target_match": true,
           "observations_reference": "<observation tuple>",
           "reference_ok": true,
           "reference_match": true
       }
   ],
   "n_valid": 8,
   "n_distinct_observations": 5,
   "coverage_train": 0.85,
   "S_target": 0.75,
   "S_reference": 1.0,
   "verdict_target": "different",
   "verdict_reference": "equivalent",
   "execution_notes": "mocked requests.get and self.db_cursor; 8/10 input states valid"
}
```

Verdict rules (T=0.8):
- "equivalent": S >= 0.8
- "different": S < 0.8
- "inconclusive": n_valid < 3 OR fewer than 3 distinct observation tuples

Notes on verdicts:
- "Different" is straightforward: if the same input and scaffolding produces different observations, the functions behave differently. 
- "Equivalent" requires a high bar. Before accepting an equivalent verdict (S >= T), verify that the matching observations exercise meaningfully different behaviors (not all trivial/degenerate) and 
mocks produce realistic responses that test the function's core logic. If the evidence is weak despite a high score, iterate carefully to get better evidence. Only after 3 honest iterations, 
downgrade to "inconclusive".
- For "inconclusive": This should be very rare — only after you have genuinely invested substantial effort and still lack >=3 distinct, meaningful observations.

Save results by augmenting the input sample dictionary with key `"execution_based_claude"`. When done, save the augmented list back to `DATA_PATH`. Do not erase existing values.

## Common Patterns 
- Void / `-> None` functions: Observe mock call logs, `self` mutations, captured output.
- Untyped parameters: Read F_train's body to infer types. If `user.email` is accessed, `user` needs an `email` attribute. Build a mock accordingly.
- External APIs: Mock the HTTP library (`requests`, `urllib`). Record what URL, headers, and body were sent. Return a plausible response.
- Database calls. Mock the session/cursor. Record the queries executed and parameters bound. Return plausible result sets that exercise branches.
- Plotting functions: Mock `plt.savefig`, `plt.show`, etc. The observation is the sequence of matplotlib calls: what was plotted, with what data, what labels. 
- ML model functions: Mock the model object. `model.eval()` is recorded. `model(input)` returns a tensor of the right shape. The observation is what methods were called on the model 
and in what order.
- Class methods mutating `self`: Snapshot `self.__dict__` before and after. The diff is part of the observation.

## Principles
0. Every function can and must be executed. If you can not get F_train to run end-to-end with observable, branch-dependent behavior, you have not done your job. Inconclusive is a last resort
after at least 3 honest scaffolding iterations, not an escape hatch for "this looked hard". Before marking inconclusive, you must be able to articulate the *specific* unsatisfiable dependency— 
not "too complex to mock", "can not create inputs or record outputs from files",  "many imports".
1. Never modify the functions under test. F_train, F_target, and F_reference must be executed verbatim. You may build any scaffolding, mocks, or wrappers around them but the function code 
itself is sacred. If your scaffolding works for F_train, use it unchanged for the others. If the same process fails for a generated function, this is a meaningful signal that needs to be recorded
as a difference.
2. Behavior is the judge. Return values are one signal among many. A void function that calls `self.save(data)` is testable — check whether `save` was called with the same `data`. A function that 
mutates a list argument is testable — check the list's state after the call.
3. Coverage matters. Read the function body and carefully design input states that exercise logic across all branches.
4. Work hard per sample. Each function is different. Read it. Understand it. Produce input states that meaningfully test its functionality. Build scaffolding tailored to it to ensure 
meaningful execution. Do not apply a generic harness, boilerplate mocks produce low-diversity observations and inconclusive results. 
5. Mocks must be realistic. A mock that returns `None` or `0` for everything is worthless. Mocks must cover a range of branches. 
6. Clean up. Create all temporary files in a single folder and remove it when done.

## Configuration

LANGUAGE = "__LANGUAGE__"
DATA_PATH = "__SAMPLE_PATH__"

Process the single sample in this file. Save results by augmenting the sample dict with key "execution_based_claude". Save the augmented list back to DATA_PATH.
\end{Verbatim}
\end{promptbox}

\textbf{Timing.} Table~\ref{tab:timing_exec} reports the mean and standard deviation of the agent execution time across languages for the OLMo-3-32B setup (evaluating both target and reference generations per sample), excluding samples that have timed-out. 

\begin{table}[h]
\centering
\caption{Agent execution time statistics per language for OLMo-3-32B. Each agent evaluates both the target and reference model generations for a single sample. Timed-out samples (exceeding 900s) are excluded.}
\resizebox{0.6\linewidth}{!}{
\begin{tabular}{ccccc}
\toprule
\textbf{Language} & \textbf{N functions} & \textbf{Timeouts (\%)} & \textbf{Mean (s)} & \textbf{Std (s)} \\
\midrule
Python (CraneCode) & 5{,}000 & 46 (0.92\%) & 143.5 & 76.0 \\
Python (Stack-Edu) & 5{,}000 & 74 (1.48\%) & 123.2 & 60.2 \\
Java & 5{,}000 & 24 (0.48\%) & 205.0 & 84.7 \\
C & 5{,}000 & 93 (1.86\%) & 168.6 & 65.5 \\
Rust & 2{,}612 & 24 (0.92\%) & 256.0 & 141.0 \\
Go & 3{,}735 & 57 (1.53\%) & 167.3 & 64.4 \\
\bottomrule
\end{tabular}
}
\label{tab:timing_exec}
\end{table}

\textbf{Coverage.} Table~\ref{tab:coverage_agent} further summarizes for how many functions the agent yields a conclusive execution-based similarity across models and datasets. As stated in Section~\ref{sec:exp_setup}, we consider a similarity score valid if the agentic execution is not timed out (see Table~\ref{tab:timing_exec}), the agent considers at least $N_\text{in}=8$ distinct valid input states (where valid means the ground-truth function $x$ executes without error) and $N_\text{out}=2$ distinct output states and reaches a minimum line coverage of $C=80\%$ of the ground-truth function body. Across setups, we obtain a conclusive similarity score for 90\% or more of the functions, with Rust being the exception at approximately 80\%. Upon inspection, the inconclusive cases arise when the agent fails to compile or execute the training data function, or when there is insufficient input-output diversity to draw a meaningful conclusion.

\begin{table}[h]
\centering
\caption{Number of functions with conclusive execution-based similarity score across models and datasets.}
\resizebox{0.65\linewidth}{!}{
\begin{tabular}{cccc}
\toprule
\textbf{Model} & \textbf{Dataset} & \textbf{N functions} & \textbf{N functions conclusive (\%)} \\
\midrule
\multirow{6}{*}{OLMo-3-32B} & Python (CraneCode) & 5{,}000 & 4{,}510 (90.20\%) \\
& Python (Stack-Edu) & 5{,}000 & 4{,}667 (93.34\%) \\
& Java & 5{,}000 & 4{,}467 (89.34\%) \\ 
& C & 5{,}000 & 4{,}526 (90.52\%) \\ 
& Rust & 2{,}612 & 2{,}086 (79.86\%) \\
& Go & 3{,}735 & 3{,}372 (90.28\%) \\
\midrule 
\multirow{6}{*}{OLMo-3-7B} & Python (CraneCode) & 5{,}000 & 4{,}494 (89.88\%) \\ 
& Python (Stack-Edu) & 5{,}000 & 4{,}678 (93.56\%) \\ 
& Java & 5{,}000 & 4{,}483 (89.66\%) \\
& C & 5{,}000 & 4{,}562 (91.24\%) \\
& Rust & 2{,}612 & 2{,}077 (79.52\%) \\
& Go & 3{,}735 & 3{,}310 (88.62\%) \\
\midrule
StarCoder-15B & Python & 5{,}000 & 4{,}586 (91.78\%) \\ 
\midrule
CrystalCoder-7B & Python & 5{,}000 & 4{,}554 (91.15\%) \\ 
\bottomrule
\end{tabular}
}
\label{tab:coverage_agent}
\end{table}

\begin{figure}[t]
\centering \begin{subfigure}[b]{0.25\linewidth} 
\includegraphics[width=\linewidth]{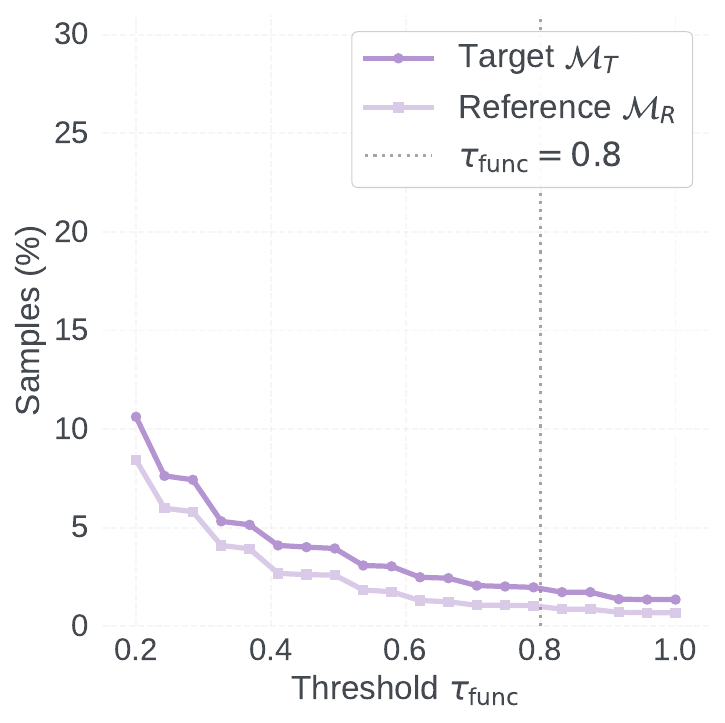} \caption{Python (CraneCode)} 
\end{subfigure} 
\begin{subfigure}[b]{0.25\linewidth} 
\includegraphics[width=\linewidth]{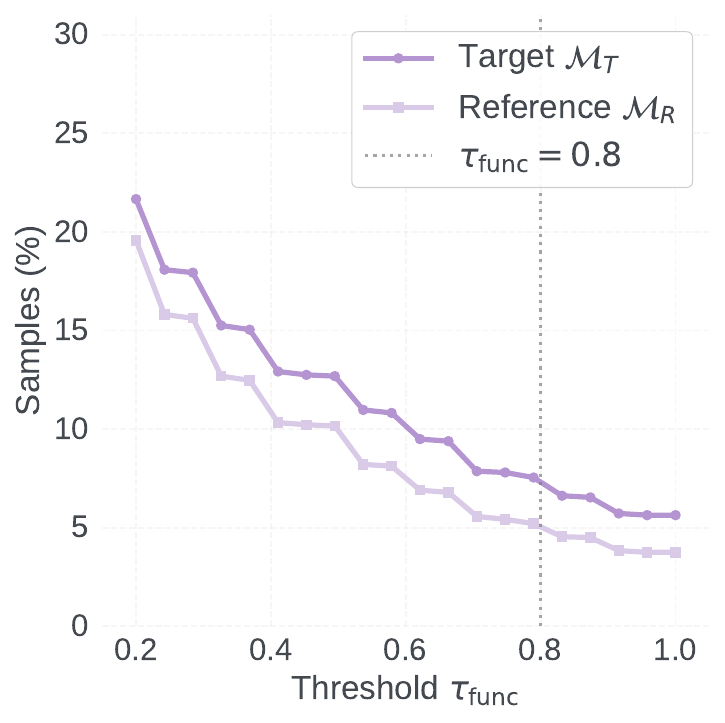} \caption{Python (Stack-edu)}
\end{subfigure} 
\begin{subfigure}[b]{0.25\linewidth} 
\includegraphics[width=\linewidth]{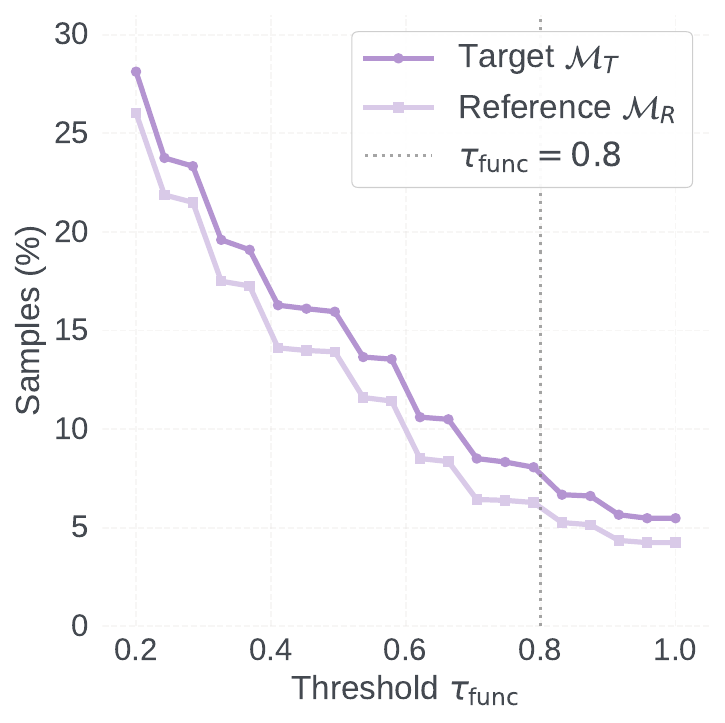} 
\caption{C} 
\end{subfigure} 

\vspace{0.5em} 

\begin{subfigure}[b]{0.25\linewidth}
\includegraphics[width=\linewidth]{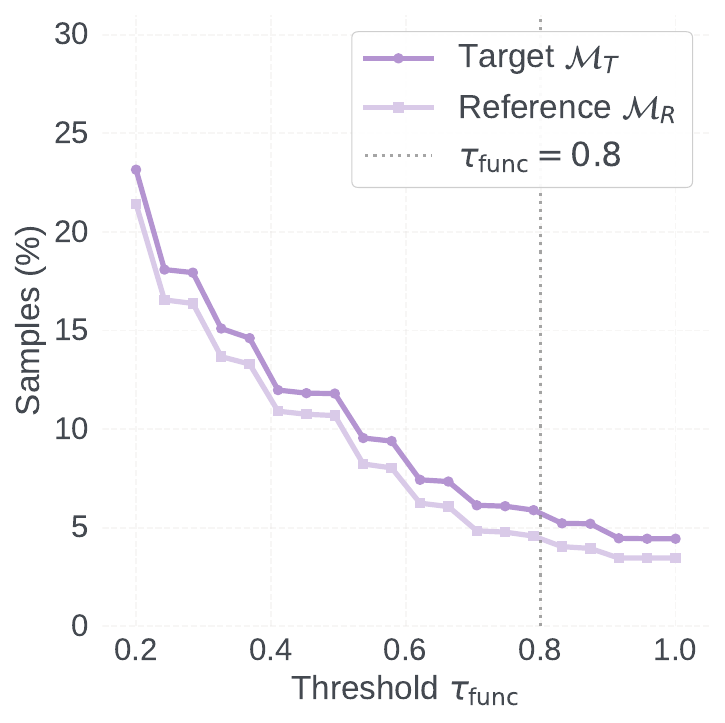} \caption{Java} 
\end{subfigure} 
\begin{subfigure}[b]{0.25\linewidth} 
\includegraphics[width=\linewidth]{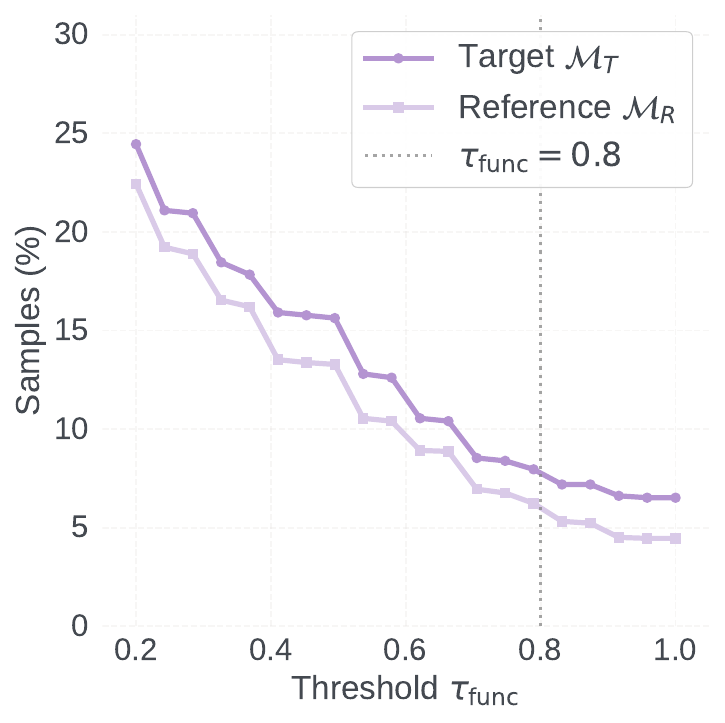} \caption{Rust} 
\end{subfigure} 
\begin{subfigure}[b]{0.25\linewidth}
\includegraphics[width=\linewidth]{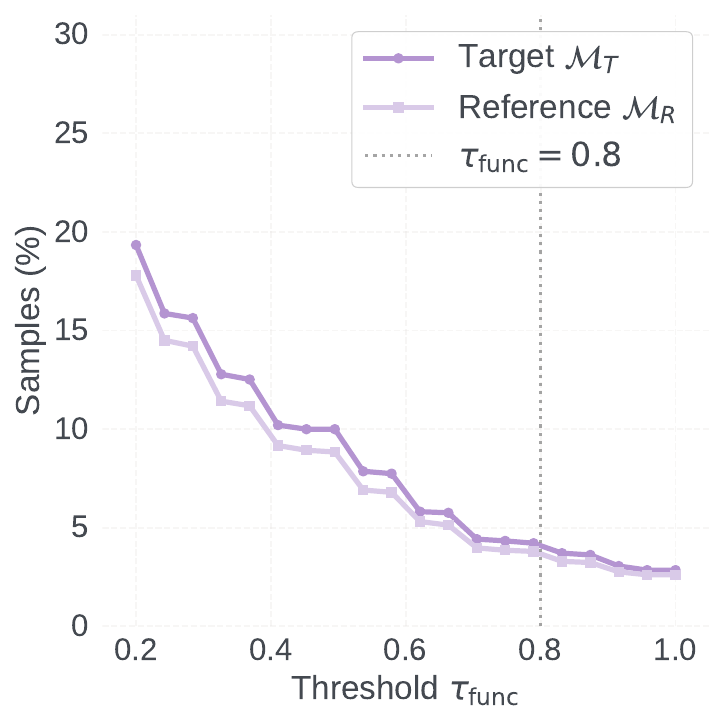} 
\caption{Go} 
\end{subfigure} 

\vspace{0.5em} 

\caption{Fraction of samples with execution based similarity higher than threshold $\tau_\text{func}$ for varying values of $\tau_\text{func}$ across data subsets for OLMo-3-32B. Across thresholds, the fraction of samples remains consistently higher for the target model than for the reference model. The number converges for higher values of $\tau_\text{func}$, motivating the choice of $\tau_\text{func}=0.8$ in Figure~\ref{fig:memorization_combined}.} 
\label{fig:threshold_sensitivity} 
\end{figure}

\textbf{Threshold sensitivity.} Figure~\ref{fig:threshold_sensitivity} reports the fraction of samples flagged as functionally similar using execution-based testing for varying values of threshold $\tau_\text{func}$. Across threshold values for all datasets for OLMo-3-32B, the target model consistently yields more functionally similar samples than the reference model, with both curves converging to stable values at higher thresholds. In the main paper, we set $\tau_\text{func}=0.8$ as it yields stable results and is sufficiently strict while also tolerating minor execution mismatches.

\begin{figure}[!h]
\centering
\includegraphics[width=0.4\linewidth]{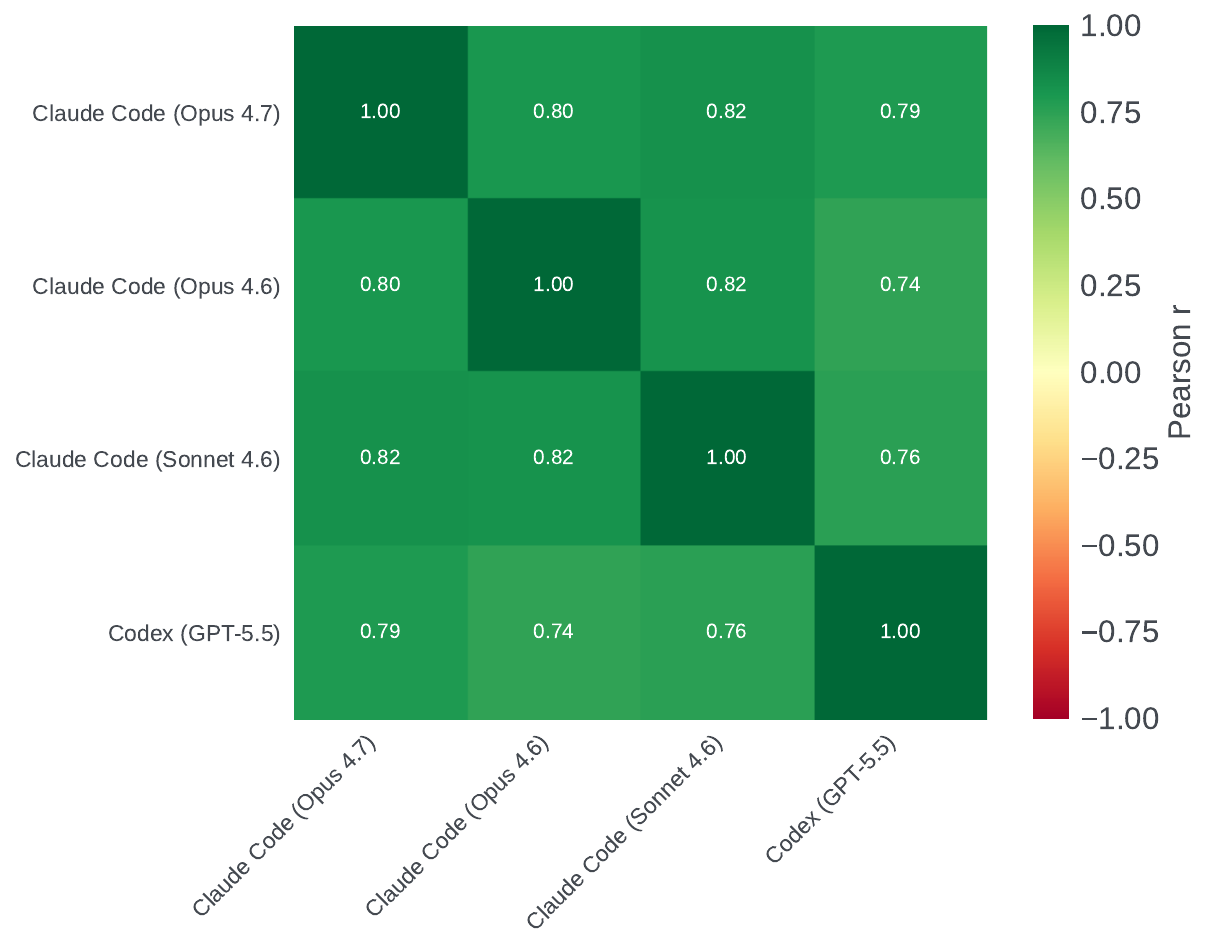}
\caption{Pearson correlation of execution-based similarity scores across models and agentic harnesses, computed for Python functions from CraneCode generated by OLMo-3-32B.}
\label{fig:corr_across_agents}
\end{figure}

\textbf{Ablation across models and agentic harnesses.} Throughout this paper, we use Anthropic's Claude Code~\citep{anthropic2025claudecode} with Opus-4.7 as the underlying model to obtain execution-based similarity scores. We also stress-test the robustness of our scores by instantiating alternative agents with the same prompt and data. Specifically, we consider Claude Code with its previous-generation model Opus-4.6 and the smaller Sonnet-4.6 variant, as well as OpenAI's Codex~\citep{openai2025codex} with GPT-5.5 as the underlying model. Figure~\ref{fig:corr_across_agents} reports the Pearson correlation of execution-based similarity scores for Python functions from CraneCode generated by OLMo-3-32B. All pairwise correlations exceed 0.74, with the highest
r=0.79 between Claude Code Opus-4.7 and Codex GPT-5.5---comparable to the best correlation observed across LLM-judge prompts using the same model in Figure~\ref{fig:correlation_roc_main}(a). Given the substantial degrees of freedom available to each agent in generating test cases, executing them reliably, and aggregating results, we consider this level of cross-agent consistency to be strong.

% One-shot (non-agentic) execution ablation.
% Requires: \usepackage{booktabs}, \usepackage{fancyvrb} (for Verbatim), and the
% `promptbox` environment from the paper template.
% Numbers in Table~\ref{tab:ablation-agent} are illustrative (last 100-function run); update with your final run.

\textbf{Ablation: are agents \emph{necessary} for execution-based auditing?} Our main pipeline relies on an AI coding agent to make each function executable (resolving dependencies, mocking state, instantiating classes, and iterating on failures). To justify this, we ablate it against non-agentic alternatives. We execute only the ground-truth training function $x$ and ask, for what fraction of functions can a non-agentic procedure obtain a \emph{conclusive} execution-based measurement? A measurement is conclusive under the same validity criteria as our main experiments (Section~\ref{sec:exp_setup}): at least $N_{\text{in}}=8$ of the $N=10$ generated inputs execute $x$ without error, they yield at least $N_{\text{out}}=2$
distinct output states, and they cover at least $C=80\%$ of the function body.

For each function we prompt an LLM once to generate $N=10$ inputs, using the following prompt:

\paragraph{One-shot prompt for test-input generation.} Once per function, the model infers parameter types from the signature, type hints, and body, and returns $N=10$ diverse, JSON-serializable inputs.

\begin{promptbox}
\begin{Verbatim}[fontsize=\scriptsize]
System: You are an expert Python test input generator. Generate inputs that a function can actually execute without errors. Carefully
analyze the signature, type hints, docstrings, and how parameters are used in the body to determine the correct types. 
Output ONLY valid JSON.

User: Analyze the following function and generate test inputs that execute WITHOUT errors:
```python
{function_code}
```
Rules: use only JSON-serializable types; never use lambdas, class constructors, function calls, or comprehensions;
represent a numpy array as a nested list; represent a custom object (ListNode/TreeNode/self) as a plain dict or list; 
provide a list/string where a parameter is indexed or iterated; do NOT include 'self'; keep numbers small and lists short. 
Generate 10 diverse inputs. Respond with ONLY valid JSON:
{"test_inputs": [{"param1": value1, ...}, ...]}
\end{Verbatim}
\end{promptbox}

We then attempt to execute $x$ on every generated input inside a sandboxed, single-use
subprocess with a set of common third-party packages pre-installed. To rigorously ablate on how non-agentic setups yield successful executions of functions sampled from real-world training corpora, we consider four levels of instantiating the execution:

\begin{enumerate}
    \item the function alone (\textbf{bare});
    \item the function with a fixed preamble importing standard-library packages such as \texttt{numpy}, \texttt{pandas} and \texttt{os} (\textbf{common imports});
    \item the function with an LLM-generated preamble, where the same LLM sees $x$ and the code preceding it in its source file and emits, in a single call, the imports, globals, and helper/class definitions needed to run $x$ standalone (see \emph{prompt to generate preamble} below) (\textbf{generated preamble});
    \item the function with an LLM-generated preamble, where, if (3) fails, the LLM is shown its preamble and the resulting execution error and produces a corrected preamble exactly once (see \emph{prompt to generate preamble with retry} below) (\textbf{generated preamble, one retry}).
\end{enumerate}

We reuse the same generated inputs across all levels so that any difference is attributable purely to the execution context. To ensure fair comparison, all LLM calls use the same model (Claude Opus-4.7) as used throughout this work as part of our coding agent. 

\paragraph{Prompt to generate preamble.} The model sees the function and the code preceding it in its source file, and returns in a single call the imports, globals, and helper/class definitions needed to run the function standalone.

\begin{promptbox}
\begin{Verbatim}[fontsize=\scriptsize]
System: You are an expert Python code analyst. Given a function and the code that precedes it in its source file, 
output the minimal preamble (imports, globals, helper/class definitions) required to
execute the function standalone. 
The preamble MUST be COMPLETE, valid Python that compiles on its own. 
Do NOT include any `from __future__` imports. Output ONLY valid JSON.

User: Code preceding the function in its source file:
```python
{file_context}
```
Function to execute:
```python
{function_code}
```
These imports are already available:
```python
{common_imports}
```
Provide ONLY the additional imports, globals, and helper/class definitions needed to run the function standalone, as ONE 
complete code block. Respond with ONLY valid JSON:
{"preamble": "<complete python code>"}
\end{Verbatim}
\end{promptbox}

\paragraph{Prompt to generate preamble with retry.} If the prompt above fails to yield a valid execution, the model is shown its previous preamble and the execution error, and returns a single corrected preamble.

\begin{promptbox}
\begin{Verbatim}[fontsize=\scriptsize]
System: You are an expert Python code analyst. A preamble you wrote to make a function run standalone did not work. 
Given the function, its file context, your previous preamble, and the error, produce a
CORRECTED, COMPLETE, valid-Python preamble. 
Do NOT include any `from __future__` imports. Output ONLY valid JSON.

User: {file_context and function as above}
Your PREVIOUS preamble (it failed):
```python
{previous_preamble}
```
Executing the function then failed with:
```
{execution_error}
```
Return a CORRECTED preamble (one complete Python code block) that fixes the error. 
Respond with ONLY valid JSON:
{"preamble": "<complete python code>"}
\end{Verbatim}
\end{promptbox}

We instantiate all four levels to the $5,000$ Python functions from CraneCode and report the fraction of functions that yield valid execution-based measurement in Table~\ref{tab:ablation-agent}. The results reveal a clear trend: each additional layer of context consistently increases the fraction of executable functions, confirming that real-world code demands progressively richer scaffolding.

We find that bare and common-imports execution succeed for only a small fraction of functions ($<4\%$), as real-world code is rarely self-contained, depending on file-level imports, module constants, file-local helpers, and---for methods---a live instance. The LLM-generated preamble, which can reconstruct file-local helpers and constants, raises this to only $\sim$8\%. The single error-feedback retry further adds
$\sim$4--5 percentages points. While each increment is meaningful, even the most capable non-agentic configuration leaves the large majority of functions without a conclusive measurement. 

Upon inspection, we find the residual failures to be structural rather than easily fixable by prompting: (i)~methods that require an actual \texttt{self} instance, (ii)~third-party or project-internal modules that cannot be imported in the sandbox, and (iii)~functions that need external resources or state (files, databases, network services). 

Observing the trend that each layer of added flexibility yields more executable functions is precisely what motivated the agentic execution. Instead of a fixed set of heuristics, a coding agent can iteratively resolve or mock dependencies, instantiate classes, supply resources, retry on errors, record outputs beyond return values---closing the gap that LLM generation, even with a retry, cannot.

\begin{table}[t]
\centering
\small
\begin{tabular}{lc}
\toprule
Execution context & Valid functions (\%) \\
\midrule
Bare function                       & 2.2 \\
\quad + common imports              & 3.1 \\
\quad + LLM preamble                & 8.6 \\
\quad + LLM preamble (one retry)    & 12.8 \\
\midrule
Agentic harness (main paper)        & 90.2 \\
\bottomrule
\end{tabular}
\caption{Fraction of Python functions from CraneCode for which a non-agentic procedure yields a conclusive execution-based measurement. Adding additional complexity to the execution instantation consistently improves coverages, but even with an LLM-generated preamble (with 1 retry), the conclusive rate remains as low as 12.8\%. Coding agents, however, powered with the same underlying model (Claude Opus-4.7) obtain a conclusive result for the large majority of functions (90.2\%).}
\label{tab:ablation-agent}
\end{table}

%% file: appendix/details_metrics.tex
\subsection{Prompts used for LLM-based similarity metrics}
\label{app:prompts_judge}

To evaluate functional similarity between reference and generated functions, we employ three LLM-as-a-judge prompting strategies, each instantiated with LLaMA-3.3-70B-Instruct~\cite{grattafiori2024llama} (default) and with Claude Sonnet-4.6~\citep{anthropic2026sonnet46} in Figure~\ref{fig:correlation_roc_main}. For each prompt, the model receives both the reference function and the generated function as context.

\paragraph{Prompt 1: Song et al.~\citep{song2024revisiting}} We adapt the GPT Structure Similarity metric of~\citet{song2024revisiting} (Section~2.2), originally designed for Java. The model returns a single 0--1 similarity score in \texttt{[[X.XXX]]} format.

\begin{promptbox}
\begin{Verbatim}[fontsize=\scriptsize]
System: You are a code similarity evaluation system.

User: Given 2 __LANGUAGE__ code paragraphs, please generate a similarity
score from 0 to 1 (to three decimal places), by grammar parsing
structure. Answer with a format like [[0.777]].
=====Code 1=====
{reference_function}
=====Code 2=====
{generated_function}
=====End=====
\end{Verbatim}
\end{promptbox}

\paragraph{Prompt 2: Nikiema et al.~\citep{nikiema2025small}} Following the LLM-based metric of~\citet{nikiema2025small} (here adapted according to the description in their Section~3.3.3), the model returns a 0--1 similarity rating, a categorical classification (\textit{equivalent}, \textit{similar}, \textit{opposed}, or \textit{unrelated}), and a brief reasoning justification.

\begin{promptbox}
\begin{Verbatim}[fontsize=\scriptsize]
System: You are a code similarity evaluation system. You assess
semantic relationships between code fragments.

User: Evaluate the semantic similarity between the following two
__LANGUAGE__ code fragments.

Code Fragment 1:
{reference_function}

Code Fragment 2:
{generated_function}

Provide your evaluation as a JSON object with these three fields:
1. "similarity_rating": a score from 0.0 to 1.0 indicating semantic
   similarity
2. "category": one of "equivalent", "similar", "opposed", or
   "unrelated"
3. "reasoning": a brief justification for your rating

Respond with ONLY the JSON object.
\end{Verbatim}
\end{promptbox}

\paragraph{Prompt 3: Custom Functional Equivalence} Our own prompt, designed to capture more specifically whether the generated code encodes the same logic as the reference via a single 0--1 equivalence score.

\begin{promptbox}
\begin{Verbatim}[fontsize=\scriptsize]
System: You are a code analysis expert evaluating functional
equivalence between two __LANGUAGE__ functions.

User: Evaluate whether the following two __LANGUAGE__ functions encode the
same logic. Consider whether they would produce the same outputs for
the same inputs, whether they implement the same algorithm or
approach, and whether they handle edge cases similarly.

Function A (Reference):
{reference_function}

Function B (Generated):
{generated_function}

Provide a single equivalence score from 0.0 to 1.0, where 0.0 means
completely different logic and 1.0 means identical logic. Respond
with ONLY a JSON object:
{"equivalence_score": 0.0 to 1.0, "reasoning": "brief explanation"}
\end{Verbatim}
\end{promptbox}

%% file: appendix/examples.tex
We here present qualitative examples of functions that are memorized by the target model. In Section~\ref{app:examples_verbatim}, we provide three examples in which the target model generates the ground truth training function near-verbatim and, in Section~\ref{app:examples_functional}, we specifically focus on \emph{functional} memorization, and provide 3 examples for which the target model generates code that is functionally similar to the logic present in the training data despite having low textual overlap. For each example, we show the full prompt used to query both the target and the reference model, the full ground-truth continuation from the training data, the full generated continuation from both models, and a table with all computed similarity metrics. We use \colorbox{red!10}{red} for the target model and \colorbox{green!10}{green} for the reference model. We first illustrate examples in Sections~\ref{app:examples_verbatim} and~\ref{app:examples_functional} for Python functions from CraneCode and for OLMo-3-32B, and provide examples of functional memorization across programming languages and training setups in Section~\ref{app:examples_function_across_setups}.

% ════════════════════════════════════════════════════════
\subsection{Near-verbatim textual memorization (Python functions from CraneCode for OLMo-3-32B)}
\label{app:examples_verbatim}

% ────────────────────────────────────────────────────────
\paragraph{Example 1: Baidu Translate request URL (\texttt{construct\_api\_request\_url}).}
\label{ex:baidu_translate_url}

The target memorizes the exact Baidu Translate request construction: the salt
range \texttt{random.randint(32768, 65536)}, the call to \texttt{generate\_signature},
the API path \texttt{'/api/trans/vip/translate'}, the \texttt{query\_params}
dictionary (\texttt{appid}, \texttt{q} with \texttt{quote(text)}, \texttt{from},
\texttt{to}, \texttt{salt}, \texttt{sign}), and the final
\texttt{urllib.parse.urlencode} assembly. The only differences from the ground
truth are marginal: the \texttt{base\_url} is defined after the dictionary, and
\texttt{salt} is stored as an \texttt{int} rather than \texttt{str(salt)}. The
target reproduces the training function's output on all input states (execution based similarity of 1.0). The reference model produces a working-looking but functionally different URL: it uses the wrong salt range \texttt{random.randint(1, 1000000)} and builds a hardcoded full-host f-string (\texttt{https://api.fanyi.baidu.com/...}) instead of the params dictionary, so it matches on none of the input states.

\smallskip
\noindent\begin{tcolorbox}[colback=gray!3, colframe=gray!50, title={\scriptsize\textbf{Prompt} (Context up to $P=250$ tokens $+$ function signature)},
boxsep=2pt, left=4pt, right=4pt, top=3pt, bottom=3pt]
\begin{Verbatim}[fontsize=\tiny]
import http.client
import hashlib
import urllib.parse
import random
import json
import logging

# Configure logging
logging.basicConfig(level=logging.ERROR)

def generate_signature(app_id: str, text: str, salt: int,
                     secret_key: str) -> str:
  """
  Generates a signature for the Baidu Translate API request.

  :param app_id: Application ID for Baidu Translate API
  :param text: Text to be translated
  :param salt: Random salt value
  :param secret_key: Secret key for Baidu Translate API
  :return: MD5 hash of the concatenated string
  """
  concatenated_string = app_id + text + str(salt) + secret_key
  md5_hash = hashlib.md5(concatenated_string.encode()).hexdigest()
  return md5_hash

def construct_api_request_url(app_id: str, secret_key: str, text: str,
                           source_language: str,
                           target_language: str) -> str:
\end{Verbatim}
\end{tcolorbox}

\smallskip
\noindent\begin{tcolorbox}[colback=blue!3, colframe=blue!40, title={\scriptsize\textbf{Ground truth continuation from the training data} (sampled from
CraneCode, labeled as severity 8)}, boxsep=2pt, left=4pt, right=4pt, top=3pt, bottom=3pt]
\begin{Verbatim}[fontsize=\tiny]
  """
  Constructs the URL for the Baidu Translate API request.

  :param app_id: Application ID for Baidu Translate API
  :param secret_key: Secret key for Baidu Translate API
  :param text: Text to be translated
  :param source_language: Source language code
  :param target_language: Target language code
  :return: Constructed URL for the API request
  """
  salt = random.randint(32768, 65536)
  signature = generate_signature(app_id, text, salt, secret_key)
  base_url = '/api/trans/vip/translate'
  query_params = {
      'appid': app_id,
      'q': urllib.parse.quote(text),
      'from': source_language,
      'to': target_language,
      'salt': str(salt),
      'sign': signature
  }
  return f"{base_url}?{urllib.parse.urlencode(query_params)}"
\end{Verbatim}
\end{tcolorbox}

\smallskip
\noindent\begin{minipage}[t]{0.48\linewidth}
\begin{tcolorbox}[colback=red!3, colframe=red!40, title={\scriptsize\textbf{Generated continuation from the target model} (OLMo-3-32B)},
boxsep=2pt, left=3pt, right=3pt, top=2pt, bottom=2pt]
\begin{Verbatim}[fontsize=\tiny]
  """
  Constructs the URL for the Baidu Translate API request.

  :param app_id: Application ID for Baidu Translate API
  :param secret_key: Secret key for Baidu Translate API
  :param text: Text to be translated
  :param source_language: Source language code
                          (e.g., 'zh' for Chinese)
  :param target_language: Target language code
                          (e.g., 'en' for English)
  :return: Complete URL for the API request
  """
  salt = random.randint(32768, 65536)
  signature = generate_signature(app_id, text, salt, secret_key)
  query_params = {
      'appid': app_id,
      'q': urllib.parse.quote(text),
      'from': source_language,
      'to': target_language,
      'salt': salt,
      'sign': signature
  }
  base_url = '/api/trans/vip/translate'
  return f"{base_url}?{urllib.parse.urlencode(query_params)}"
\end{Verbatim}
\end{tcolorbox}
\end{minipage}\hfill
\begin{minipage}[t]{0.48\linewidth}
\begin{tcolorbox}[colback=green!3, colframe=green!40!gray, title={\scriptsize\textbf{Generated continuation from the reference model} (OLMo-3-32B)}, boxsep=2pt, left=3pt, right=3pt, top=2pt, bottom=2pt]
\begin{Verbatim}[fontsize=\tiny]
  """
  Constructs the API request URL for Baidu Translate API.

  :param app_id: Application ID for Baidu Translate API
  :param secret_key: Secret key for Baidu Translate API
  :param text: Text to be translated
  :param source_language: Source language code
  :param target_language: Target language code
  :return: API request URL
  """
  salt = random.randint(1, 1000000)
  signature = generate_signature(app_id, text, salt, secret_key)
  url = (f"https://api.fanyi.baidu.com"
         f"/api/trans/vip/translate?appid={app_id}"
         f"&q={urllib.parse.quote(text)}"
         f"&from={source_language}&to={target_language}"
         f"&salt={salt}&sign={signature}")
  return url
\end{Verbatim}
\end{tcolorbox}
\end{minipage}

\begin{table}[!h]
\centering
\resizebox{0.5\linewidth}{!}{ \begin{tabular}{llcc}
\toprule
\multicolumn{2}{c}{\textbf{Metric}} & \textbf{Reference} & \textbf{Target} \\
\midrule
\multirow{3}{*}{Textual} & BLEU & 0.51 & 0.77 \\
& Edit Similarity Score & 0.71 & 0.88 \\
& LCS Score & 0.36 & 0.36 \\
\midrule
\multicolumn{2}{c}{Execution} & 0.00 & 1.00 \\
\midrule
\multirow{4}{*}{Structural} & CodeBLEU (equal) & 0.53 & 0.85 \\
& CodeBLEU (syntax only) & 0.48 & 0.84 \\
& CodeBLEU (DFG only) & 0.46 & 0.90 \\
& TSED & 0.57 & 0.88 \\
\midrule
\multirow{4}{*}{Embedding} & Qwen-0.6B & 0.98 & 0.99 \\
& Qwen-4B & 0.98 & 0.99 \\
& Qwen-8B & 0.98 & 1.00 \\
& Nomic-code-7B & 0.97 & 0.99 \\
\midrule
\multirow{3}{*}{LLM-judge} & Song & 0.85 & 0.98 \\
& Nikiema & 0.90 & 0.98 \\
& Ours & 0.80 & 1.00 \\
\bottomrule
\end{tabular}
}
\end{table}

% ────────────────────────────────────────────────────────
\paragraph{Example 2: Dust diffusion simulation (\texttt{diffuse\_dust}).}
\label{ex:diffuse_dust}

The target reproduces the grid-based dust diffusion algorithm near-verbatim: spread \texttt{grid[i][j]//5} to each of the 4 cardinal neighbors, skip air cleaner cells marked as $-1$, and use a temporary grid to avoid overwriting values during the sweep. The reference model generates an incorrect one-liner (\texttt{grid[i][j] -= 1}) that ignores the diffusion rules entirely. The execution-based testing yields a similarity of $0.7$, which exposes the difference in how both algorithms handle cells with -1 at the end of processing (training data function adds \texttt{temporary\_grid[i][j]}, while the target model generation instead overwrites to \texttt{grid[i][j] = temp\_grid[i][j]}).

\smallskip
\noindent\begin{tcolorbox}[colback=gray!3, colframe=gray!50, title={\scriptsize\textbf{Prompt} (Context up to $P=250$ tokens $+$ function signature)}, boxsep=2pt, left=4pt, right=4pt, top=3pt, bottom=3pt]
\begin{Verbatim}[fontsize=\tiny]
def locate_air_cleaners(grid: List[List[int]], rows: int) -> Tuple[int, int]:
    """
    Locates the positions of the air cleaners in the grid.

    Args:
        grid: The grid representing the environment.
        rows: Number of rows in the grid.

    Returns:
        A tuple containing the row indices of the upper and lower air cleaners.
    """
    upper_cleaner = -1
    lower_cleaner = -1
    for i in range(rows):
        if grid[i][0] == -1:
            upper_cleaner = i
            lower_cleaner = i + 1
            break
    return upper_cleaner, lower_cleaner

def diffuse_dust(grid: List[List[int]], rows: int, cols: int) -> None:
\end{Verbatim}
\end{tcolorbox}

\smallskip
\noindent\begin{tcolorbox}[colback=blue!3, colframe=blue!40, title={\scriptsize\textbf{Ground truth continuation from the training data} (sampled from CraneCode, labeled as severity 4)}, boxsep=2pt, left=4pt, right=4pt, top=3pt, bottom=3pt]
\begin{Verbatim}[fontsize=\tiny]
    """
    Spreads dust particles in the grid according to the diffusion rules.

    Args:
        grid: The grid representing the environment.
        rows: Number of rows in the grid.
        cols: Number of columns in the grid.
    """
    directions = [(-1, 0), (0, -1), (0, 1), (1, 0)]
    temporary_grid = [[0] * cols for _ in range(rows)]

    for i in range(rows):
        for j in range(cols):
            if grid[i][j] > 0:
                spread_amount = 0
                for dx, dy in directions:
                    nx, ny = i + dx, j + dy
                    if 0 <= nx < rows and 0 <= ny < cols and grid[nx][ny] != -1:
                        temporary_grid[nx][ny] += grid[i][j] // 5
                        spread_amount += grid[i][j] // 5
                grid[i][j] -= spread_amount

    for i in range(rows):
        for j in range(cols):
            grid[i][j] += temporary_grid[i][j]
\end{Verbatim}
\end{tcolorbox}

\smallskip
\noindent\begin{tcolorbox}[colback=red!3, colframe=red!40, title={\scriptsize\textbf{Generated continuation from the target model} (OLMo-3-32B)}, boxsep=2pt, left=4pt, right=4pt, top=3pt, bottom=3pt]
\begin{Verbatim}[fontsize=\tiny]
    """
    Diffuses dust particles in the grid according to the given rules.

    Args:
        grid: The grid representing the environment.
        rows: Number of rows in the grid.
        cols: Number of columns in the grid.
    """
    directions = [(-1, 0), (1, 0), (0, -1), (0, 1)]
    temp_grid = [[0] * cols for _ in range(rows)]

    for i in range(rows):
        for j in range(cols):
            if grid[i][j] > 0:
                dust_to_spread = grid[i][j] // 5
                spread_count = 0

                for dx, dy in directions:
                    nx, ny = i + dx, j + dy
                    if 0 <= nx < rows and 0 <= ny < cols and grid[nx][ny] != -1:
                        temp_grid[nx][ny] += dust_to_spread
                        spread_count += 1

                temp_grid[i][j] += grid[i][j] - (dust_to_spread * spread_count)

    for i in range(rows):
        for j in range(cols):
            grid[i][j] = temp_grid[i][j]
\end{Verbatim}
\end{tcolorbox}

\smallskip
\noindent\begin{tcolorbox}[colback=green!3, colframe=green!40!gray, title={\scriptsize\textbf{Generated continuation from the reference model} (OLMo-3-32B)}, boxsep=2pt, left=4pt, right=4pt, top=3pt, bottom=3pt]
\begin{Verbatim}[fontsize=\tiny]
    """
    Simulates the diffusion of dust particles in the grid.

    Args:
        grid: The grid representing the environment.
        rows: Number of rows in the grid.
        cols: Number of columns in the grid.
    """
    for i in range(rows):
        for j in range(cols):
            if grid[i][j] > 0:
                grid[i][j] -= 1
\end{Verbatim}
\end{tcolorbox}

\begin{table}[!h] 
\centering 
\resizebox{0.5\linewidth}{!}{ \begin{tabular}{llcc} 
\toprule 
\multicolumn{2}{c}{\textbf{Metric}} & \textbf{Reference} & \textbf{Target} \\ 
\midrule 
\multirow{3}{*}{Textual} & BLEU & 0.16 & 0.75 \\
& Edit Similarity Score & 0.34 & 0.91 \\ 
& LCS Score & 0.20 & 0.23 \\ 
\midrule 
\multicolumn{2}{c}{Execution} & 0.10 & 0.70 \\
\midrule 
\multirow{4}{*}{Structural} & CodeBLEU (equal) & 0.23 & 0.75 \\ 
& CodeBLEU (syntax only) & 0.24 & 0.70 \\
& CodeBLEU (DFG only) & 0.15 & 0.74 \\ 
& TSED & 0.31 & 0.81 \\ 
\midrule 
\multirow{4}{*}{Embedding} & Qwen-0.6B & 0.88 & 0.97 \\ 
& Qwen-4B & 0.90 & 0.97 \\
& Qwen-8B & 0.90 & 0.97 \\
& Nomic-code-7B & 0.72 & 0.93 \\ 
\midrule 
\multirow{3}{*}{LLM-judge} & Song & 0.13 & 0.85 \\
& Nikiema & 0.10 & 0.80 \\ 
& Ours & 0.00 & 0.90 \\
\bottomrule 
\end{tabular} 
} 
\end{table}

% ────────────────────────────────────────────────────────
\paragraph{Example 3: Goodreads ratings lookup (\texttt{fetch\_goodreads\_data}).}
\label{ex:goodreads}

The target memorizes the Goodreads API pattern near-verbatim: the \texttt{review\_counts.json} endpoint, the \texttt{params} dictionary with the
\texttt{key} and \texttt{isbns} fields, the \texttt{raise\_for\_status()} error
check, and the nested JSON parsing via \texttt{data['books'][0]['work\_ratings\_count']} and \texttt{['average\_rating']}. The \emph{only} difference from the ground truth is
that the hardcoded API key (\texttt{"5MrEv1ULNl0OAS1UkxdLmA"}) is replaced by a
\texttt{GOODREADS\_API\_KEY} variable, so the model reconstructs the proprietary
request structure exactly while abstracting the literal secret, a difference which is picked up by the execution-based testing and results in execution-based similarity of 0. The reference model produces a plausible but entirely different implementation: a fabricated \texttt{book/isbn/\{isbn\}} endpoint built as an f-string with no API key, a
\texttt{status\_code == 200} check instead of \texttt{raise\_for\_status()}, and a
different (incorrect) JSON path (\texttt{data['work']['ratings\_count']}). 

\smallskip
\noindent\begin{tcolorbox}[colback=gray!3, colframe=gray!50, title={\scriptsize\textbf{Prompt} (Context up to $P=250$ tokens $+$ function signature)},
boxsep=2pt, left=4pt, right=4pt, top=3pt, bottom=3pt]
\begin{Verbatim}[fontsize=\tiny]
books WHERE title LIKE :query", {"query": query}).fetchall()
if filter_type == "isbn":
    return db_session.execute("SELECT * FROM books WHERE isbn LIKE :query", {"query": query}).fetchall()
return []

@app.route("/book/<int:book_id>", methods=["GET"])
def display_book_details(book_id: int):
    """
    Retrieve and display book details.
    Args: book_id (int): Unique identifier for the book.
    Returns: str: HTML content of the rendered template.
    """
    book = db_session.execute("SELECT * FROM books WHERE id = :book_id", {"book_id": book_id}).fetchone()
    if not book:
        return render_template("error.html", message="Book not found"), 404

    goodreads_info = fetch_goodreads_data(book.isbn)
    book_reviews = fetch_book_reviews(book_id)
    can_submit_review = can_user_submit_review(book_id)

    return render_template('book.html',
                            book=book,
                            number_of_ratings=goodreads_info['number_of_ratings'],
                            average_rating=goodreads_info['average_rating'],
                            book_reviews=book_reviews,
                            show_review_option=can_submit_review)

def fetch_goodreads_data(isbn: str) -> dict:
\end{Verbatim}
\end{tcolorbox}

\smallskip
\noindent\begin{tcolorbox}[colback=blue!3, colframe=blue!40, title={\scriptsize\textbf{Ground truth continuation from the training data} (sampled from
CraneCode, labeled as severity 8)}, boxsep=2pt, left=4pt, right=4pt, top=3pt, bottom=3pt]
\begin{Verbatim}[fontsize=\tiny]
  """
  Fetch book data from Goodreads API.
  Args: isbn (str): ISBN number of the book.
  Returns: dict: Dictionary containing number of ratings and average rating.
  """
  response = requests.get(
      "https://www.goodreads.com/book/review_counts.json",
      params={"key": "5MrEv1ULNl0OAS1UkxdLmA", "isbns": isbn})
  response.raise_for_status()
  data = response.json()
  return {'number_of_ratings': data['books'][0]['work_ratings_count'],
      'average_rating': data['books'][0]['average_rating']}
\end{Verbatim}
\end{tcolorbox}

\smallskip
\noindent\begin{minipage}[t]{0.48\linewidth}
\begin{tcolorbox}[colback=red!3, colframe=red!40, title={\scriptsize\textbf{Generated continuation from the target model} (OLMo-3-32B)},
boxsep=2pt, left=3pt, right=3pt, top=2pt, bottom=2pt]
\begin{Verbatim}[fontsize=\tiny]
  """
  Fetch book data from Goodreads API.
  Args: isbn (str): ISBN of the book.
  Returns: dict: Dictionary containing number of ratings and average rating.
  """
  response = requests.get(
      "https://www.goodreads.com/book/review_counts.json",
      params={"key": GOODREADS_API_KEY, "isbns": isbn}
  )
  response.raise_for_status()
  data = response.json()
  return {'number_of_ratings': data['books'][0]['work_ratings_count'],
      'average_rating': data['books'][0]['average_rating']}
\end{Verbatim}
\end{tcolorbox}
\end{minipage}\hfill
\begin{minipage}[t]{0.48\linewidth}
\begin{tcolorbox}[colback=green!3, colframe=green!40!gray, title={\scriptsize\textbf{Generated continuation from the reference model} (pretrained
OLMo-3-32B)}, boxsep=2pt, left=3pt, right=3pt, top=2pt, bottom=2pt]
\begin{Verbatim}[fontsize=\tiny]
  """
  Fetch book details from Goodreads API.
  Args: isbn (str): ISBN of the book.
  Returns: dict: Book details from Goodreads API.
  """
  url = f"https://www.goodreads.com/book/isbn/{isbn}"
  response = requests.get(url)
  if response.status_code == 200:
      data = response.json()
      return {"number_of_ratings": data['work']['ratings_count'],
          "average_rating": data['work']['average_rating']}
  return {}
\end{Verbatim}
\end{tcolorbox}
\end{minipage}

\begin{table}[!h]
\centering
\resizebox{0.5\linewidth}{!}{ \begin{tabular}{llcc}
\toprule
\multicolumn{2}{c}{\textbf{Metric}} & \textbf{Reference} & \textbf{Target} \\
\midrule
\multirow{3}{*}{Textual} & BLEU & 0.32 & 0.88 \\
& Edit Similarity Score & 0.58 & 0.94 \\
& LCS Score & 0.07 & 0.42 \\
\midrule
\multicolumn{2}{c}{Execution} & 0.00 & 0.00 \\
\midrule
\multirow{4}{*}{Structural} & CodeBLEU (equal) & 0.37 & 0.89 \\
& CodeBLEU (syntax only) & 0.42 & 0.76 \\
& CodeBLEU (DFG only) & 0.25 & 1.00 \\
& TSED & 0.57 & 0.97 \\
\midrule
\multirow{4}{*}{Embedding} & Qwen-0.6B & 0.97 & 1.00 \\
& Qwen-4B & 0.97 & 1.00 \\
& Qwen-8B & 0.96 & 1.00 \\
& Nomic-code-7B & 0.93 & 1.00 \\
\midrule
\multirow{3}{*}{LLM-judge} & Song & 0.62 & 0.96 \\
& Nikiema & 0.70 & 0.95 \\
& Ours & 0.00 & 0.90 \\
\bottomrule
\end{tabular}
}
\end{table}

% ════════════════════════════════════════════════════════
\subsection{Functional memorization (Python functions from CraneCode for OLMo-3-32B)}
\label{app:examples_functional}

The following examples show \emph{functional memorization}. We explicitly examine training samples for which the target model reproduces the logic in code that shares little textual overlap with the exact function, while the reference model does not generate the same logic. 

% ────────────────────────────────────────────────────────
\paragraph{Example 4: City-specific commission rates (\texttt{calculate\_commission\_rate}).}
\label{ex:commission_rate}

The target memorizes exact commission rate tables for three Bulgarian cities (Sofia, Varna, Plovdiv) with precise threshold/rate pairs---information that can only come from the training data. The reference model does not know this particular logic and hallucinates US cities with made-up flat rates. The ground truth uses a dictionary of \texttt{(threshold, rate)} tuples with a loop; the target restructures this into a flat tuple dictionary with index-based access. Despite completely different code structure, the target produces identical outputs for most valid inputs (execution based similarity of $0.8$). A more concise version of the example is shown for illustrative purposes in Figure~\ref{fig:commission_example}.

\smallskip
\noindent\begin{tcolorbox}[colback=gray!3, colframe=gray!50, title={\scriptsize\textbf{Prompt} (Context up to $P=250$ tokens $+$ function signature)}, boxsep=2pt, left=4pt, right=4pt, top=3pt, bottom=3pt]
\begin{Verbatim}[fontsize=\tiny]
def calculate_commission_rate(city: str, sales_volume: float) -> float:
\end{Verbatim}
\end{tcolorbox}

\smallskip
\noindent\begin{tcolorbox}[colback=blue!3, colframe=blue!40, title={\scriptsize\textbf{Ground truth continuation from the training data} (sampled from CraneCode, labeled as severity 3)}, boxsep=2pt, left=4pt, right=4pt, top=3pt, bottom=3pt]
\begin{Verbatim}[fontsize=\tiny]
    """
    Calculate the commission rate based on the city and sales volume.
    
    Args: 
        city (str): The name of the city.
        sales_volume (float): The sales volume.

    Returns: float: The commission rate.
    
    Raises: ValueError: If the city is not recognized or sales volume is negative.
    """

    # Define commission rates for different cities and sales thresholds

    commission_rates = {
        'Sofia': [(500, 0.05), (1000, 0.07), (10000, 0.08), (float('inf'), 0.12)],
        'Varna': [(500, 0.045), (1000, 0.075), (10000, 0.10), (float('inf'), 0.13)],
        'Plovdiv': [(500, 0.055), (1000, 0.08), (10000, 0.12), (float('inf'), 0.145)]
    }
    
    # Validate the city and sales volume
    if city not in commission_rates:
        raise ValueError(f"City '{city}' is not recognized.")
    if sales_volume < 0:
        raise ValueError("Sales volume cannot be negative.")
    # Determine the commission rate based on sales volume
    for threshold, rate in commission_rates[city]:
        if sales_volume <= threshold:
            return rate
    return 0.0  # This line should never be reached due to the last threshold being infinity
\end{Verbatim}
\end{tcolorbox}

\smallskip
\noindent\begin{minipage}[t]{0.55\linewidth}
\begin{tcolorbox}[colback=red!3, colframe=red!40, title={\scriptsize\textbf{Generated continuation from the target model} (OLMo-3-32B)}, boxsep=2pt, left=3pt, right=3pt, top=2pt, bottom=2pt]
\begin{Verbatim}[fontsize=\tiny]
    """
    Calculate the commission rate based on the city and sales volume.

    Args: 
        city (str): The city where the sales occurred.
        sales_volume (float): The total sales volume.
    
    Returns: float: The commission rate as a decimal.
    """
    
    # Define commission rates for different cities and sales volumes
    commission_rates = {
                'Sofia': (0.05, 0.07, 0.08, 0.12),
                'Varna': (0.045, 0.075, 0.10, 0.13),
                'Plovdiv': (0.055, 0.08, 0.12, 0.145)
    }
    
    # Determine the commission rate based on sales volume
    if sales_volume <= 500:
        return commission_rates.get(city, (0.0,))[0]
    elif sales_volume <= 1000:
        return commission_rates.get(city, (0.0,))[1]
    elif sales_volume <= 10000:
        return commission_rates.get(city, (0.0,))[2]
    else:
        return commission_rates.get(city, (0.0,))[3]
\end{Verbatim}
\end{tcolorbox}
\end{minipage}\hfill
\begin{minipage}[t]{0.4\linewidth}
\begin{tcolorbox}[colback=green!3, colframe=green!40!gray, title={\scriptsize\textbf{Generated continuation from the reference model (OLMo-3-32B)}}, boxsep=2pt, left=3pt, right=3pt, top=2pt, bottom=2pt]
\begin{Verbatim}[fontsize=\tiny]
    if city == "New York":
        return 0.05
    elif city == "Los Angeles":
        return 0.03
    elif city == "Chicago":
        return 0.02
    elif city == "Houston":
        return 0.01
    else:
        return 0.0
\end{Verbatim}
\end{tcolorbox}
\end{minipage}

\begin{table}[!h]
\centering
\resizebox{0.5\linewidth}{!}{ \begin{tabular}{llcc}
\toprule
\multicolumn{2}{c}{\textbf{Metric}} & \textbf{Reference} & \textbf{Target} \\
\midrule
\multirow{3}{*}{Textual} & BLEU & 0.00 & 0.26 \\
& Edit Similarity Score & 0.16 & 0.54 \\
& LCS Score & 0.02 & 0.09 \\
\midrule
\multicolumn{2}{c}{Execution} & 0.00 & 0.80 \\
\midrule
\multirow{4}{*}{Structural} & CodeBLEU (equal) & 0.08 & 0.30 \\
& CodeBLEU (syntax only) & 0.17 & 0.14 \\
& CodeBLEU (DFG only) & 0.11 & 0.47 \\
& TSED & 0.20 & 0.26 \\
\midrule
\multirow{4}{*}{Embedding} & Qwen-0.6B & 0.61 & 0.96 \\
& Qwen-4B & 0.52 & 0.98 \\
& Qwen-8B & 0.56 & 0.98 \\
& Nomic-code-7B & 0.46 & 0.96 \\
\midrule
\multirow{3}{*}{LLM-judge} & Song & 0.29 & 0.59 \\
& Nikiema & 0.20 & 0.80 \\
& Ours & 0.00 & 0.80 \\
\bottomrule
\end{tabular}
}
\end{table}

% ────────────────────────────────────────────────────────
\paragraph{Example 5: Payment calculator (\texttt{calculate\_final\_price}).}
\label{ex:final_price}

The target memorizes exact discount/interest rates (10\%, 5\%, 0\%, 20\%) for four payment methods using \texttt{if/elif} branches, while the ground truth uses a dictionary lookup with named constants. BLEU is only 0.26, yet both the execution based similarity and the LLM-judge (ours) give perfect scores of 1.0. The reference model attempts the same \texttt{if/elif} pattern but gets the rates wrong: it applies a 15\% discount for method~2 (should be 5\%), a 20\% discount for method~3 (should be 0\%), and omits method~4 entirely (20\% interest surcharge). 

\smallskip
\noindent\begin{tcolorbox}[colback=gray!3, colframe=gray!50, title={\scriptsize\textbf{Prompt} (Context up to $P=250$ tokens $+$ function signature)}, boxsep=2pt, left=4pt, right=4pt, top=3pt, bottom=3pt]
\begin{Verbatim}[fontsize=\tiny]
def calculate_final_price(original_price: float, payment_method: int) -> float:
\end{Verbatim}
\end{tcolorbox}

\smallskip
\noindent\begin{tcolorbox}[colback=blue!3, colframe=blue!40, title={\scriptsize\textbf{Ground truth continuation from the training data} (sampled from CraneCode, labeled as severity 3)}, boxsep=2pt, left=4pt, right=4pt, top=3pt, bottom=3pt]
\begin{Verbatim}[fontsize=\tiny]
    """
    Calculate the final price of a product based on the selected payment method.

    Args:
        original_price (float): The initial price of the product.
        payment_method (int): The payment method chosen by the customer.

    Returns:
        float: The final price after applying any discounts or interest.

    Raises:
        ValueError: If an invalid payment method is provided.
    """
    # Constants for discount and interest rates
    DISCOUNT_CASH_OR_CHECK = 0.10
    DISCOUNT_CREDIT_CARD = 0.05
    INTEREST_MULTIPLE_PAYMENTS = 0.20

    # Dictionary mapping payment methods to their corresponding price calculations
    payment_methods = {
        1: original_price * (1 - DISCOUNT_CASH_OR_CHECK),
        2: original_price * (1 - DISCOUNT_CREDIT_CARD),
        3: original_price,
        4: original_price * (1 + INTEREST_MULTIPLE_PAYMENTS)
    }

    # Retrieve the final price based on the payment method
    final_price = payment_methods.get(payment_method)

    # Raise an error if the payment method is invalid
    if final_price is None:
        raise ValueError("Invalid payment method")

    return final_price
\end{Verbatim}
\end{tcolorbox}

\smallskip
\noindent\begin{minipage}[t]{0.55\linewidth}
\begin{tcolorbox}[colback=red!3, colframe=red!40, title={\scriptsize\textbf{Generated continuation from the target model} (OLMo-3-32B)}, boxsep=2pt, left=3pt, right=3pt, top=2pt, bottom=2pt]
\begin{Verbatim}[fontsize=\tiny]
    """
    Calculate the final price of the
    product based on the payment method.

    Args:
        original_price (float): The original price of the product.
        payment_method (int): The payment method chosen by the user.

    Returns:
        float: The final price after applying any discounts or interest.
    """
    if payment_method == 1:
        return original_price * 0.9
    elif payment_method == 2:
        return original_price * 0.95
    elif payment_method == 3:
        return original_price
    elif payment_method == 4:
        return original_price * 1.2
    else:
        raise ValueError("Invalid payment method")
\end{Verbatim}
\end{tcolorbox}
\end{minipage}\hfill
\begin{minipage}[t]{0.4\linewidth}
\begin{tcolorbox}[colback=green!3, colframe=green!40!gray, title={\scriptsize\textbf{Generated continuation from the reference model (OLMo-3-32B)}}, boxsep=2pt, left=3pt, right=3pt, top=2pt, bottom=2pt]
\begin{Verbatim}[fontsize=\tiny]
    if payment_method == 1:
        return original_price * 0.9
    elif payment_method == 2:
        return original_price * 0.85
    elif payment_method == 3:
        return original_price * 0.8
    else:
        return original_price
\end{Verbatim}
\end{tcolorbox}
\end{minipage}

\begin{table}[!h]
\centering
\resizebox{0.5\linewidth}{!}{ \begin{tabular}{llcc}
\toprule
\multicolumn{2}{c}{\textbf{Metric}} & \textbf{Reference} & \textbf{Target} \\
\midrule
\multirow{3}{*}{Textual} & BLEU & 0.00 & 0.26 \\
& Edit Similarity Score & 0.18 & 0.52 \\
& LCS Score & 0.02 & 0.09 \\
\midrule
\multicolumn{2}{c}{Execution} & 0.20 & 1.00 \\
\midrule
\multirow{4}{*}{Structural} & CodeBLEU (equal) & 0.13 & 0.30 \\
& CodeBLEU (syntax only) & 0.16 & 0.27 \\
& CodeBLEU (DFG only) & 0.29 & 0.29 \\
& TSED & 0.24 & 0.33 \\
\midrule
\multirow{4}{*}{Embedding} & Qwen-0.6B & 0.74 & 0.96 \\
& Qwen-4B & 0.79 & 0.96 \\
& Qwen-8B & 0.77 & 0.94 \\
& Nomic-code-7B & 0.71 & 0.94 \\
\midrule
\multirow{3}{*}{LLM-judge} & Song & 0.29 & 0.85 \\
& Nikiema & 0.20 & 0.95 \\
& Ours & 0.00 & 1.00 \\
\bottomrule
\end{tabular}
}
\end{table}

% ────────────────────────────────────────────────────────
\paragraph{Example 6: Traffic ticket classifier (\texttt{determine\_traffic\_ticket}).}
\label{ex:traffic_ticket}

The target memorizes exact speed thresholds (60, 80) and the birthday tolerance ($+5$), producing a function that is functionally equivalent to the ground truth (execution based similarity~$1.0$). The reference model's output is wrong on two counts: (i)~it treats birthday as a blanket ``no ticket'' exemption rather than shifting the threshold by 5, and (ii)~it returns dollar amounts (50, 70, 100, 150) instead of the specified ticket levels (0, 1, 2).

\smallskip
\noindent\begin{tcolorbox}[colback=gray!3, colframe=gray!50, title={\scriptsize\textbf{Prompt} (Context up to $P=250$ tokens $+$ function signature)}, boxsep=2pt, left=4pt, right=4pt, top=3pt, bottom=3pt]
\begin{Verbatim}[fontsize=\tiny]
def determine_traffic_ticket(speed: int, is_birthday: bool) -> int:
\end{Verbatim}
\end{tcolorbox}

\smallskip
\noindent\begin{tcolorbox}[colback=blue!3, colframe=blue!40, title={\scriptsize\textbf{Ground truth continuation from the training data} (sampled from CraneCode, labeled as severity 3)}, boxsep=2pt, left=4pt, right=4pt, top=3pt, bottom=3pt]
\begin{Verbatim}[fontsize=\tiny]
    """
    Determine the traffic ticket based on the speed and whether it's the driver's birthday.

    Args:
        speed (int): The speed of the vehicle.
        is_birthday (bool): Indicates if it is the driver's birthday.

    Returns:
        int: The ticket level (0 = no ticket, 1 = small ticket, 2 = big ticket).
    """
    # Adjust speed limit if it's the driver's birthday
    base_speed_limit = 60
    birthday_tolerance = 5

    adjusted_speed_limit = base_speed_limit + (birthday_tolerance if is_birthday else 0)

    # Determine the ticket based on the adjusted speed limit
    if speed <= adjusted_speed_limit:
        return 0  # No ticket
    elif adjusted_speed_limit < speed <= adjusted_speed_limit + 20:
        return 1  # Small ticket
    else:
        return 2  # Big ticket
\end{Verbatim}
\end{tcolorbox}

\smallskip
\noindent\begin{minipage}[t]{0.55\linewidth}
\begin{tcolorbox}[colback=red!3, colframe=red!40, title={\scriptsize\textbf{Generated continuation from the target model} (OLMo-3-32B)}, boxsep=2pt, left=3pt, right=3pt, top=2pt, bottom=2pt]
\begin{Verbatim}[fontsize=\tiny]
    """
    Determine the traffic ticket based on the speed and whether it's
    the driver's birthday.

    Parameters:
        speed (int): The speed of the vehicle in miles per hour.
        is_birthday (bool): Whether it is the driver's birthday.

    Returns:
        int: The ticket amount in dollars.
    """
    # Define speed thresholds and
    # corresponding ticket amounts
    speed_thresholds = [60, 80]
    ticket_amounts = [0, 1, 2]

    # Adjust speed threshold if it's
    # the driver's birthday
    if is_birthday:
        speed_thresholds = [65, 85]

    # Determine the ticket amount
    # based on the speed
    for i, threshold in \
            enumerate(speed_thresholds):
        if speed <= threshold:
            return ticket_amounts[i]

    return ticket_amounts[-1]
\end{Verbatim}
\end{tcolorbox}
\end{minipage}\hfill
\begin{minipage}[t]{0.4\linewidth}
\begin{tcolorbox}[colback=green!3, colframe=green!40!gray, title={\scriptsize\textbf{Generated continuation from the reference model} (OLMo-3-32B)}, boxsep=2pt, left=3pt, right=3pt, top=2pt, bottom=2pt]
\begin{Verbatim}[fontsize=\tiny]
    if is_birthday:
        return 0
    if speed <= 60:
        return 0
    if speed <= 70:
        return 50
    if speed <= 80:
        return 70
    if speed <= 90:
        return 100
    return 150
\end{Verbatim}
\end{tcolorbox}
\end{minipage}

\begin{table}[!h]
\centering
\resizebox{0.5\linewidth}{!}{ \begin{tabular}{llcc}
\toprule
\multicolumn{2}{c}{\textbf{Metric}} & \textbf{Reference} & \textbf{Target} \\
\midrule
\multirow{3}{*}{Textual} & BLEU & 0.00 & 0.34 \\
& Edit Similarity Score & 0.22 & 0.51 \\
& LCS Score & 0.03 & 0.12 \\
\midrule
\multicolumn{2}{c}{Execution} & 0.30 & 1.00 \\
\midrule
\multirow{4}{*}{Structural} & CodeBLEU (equal) & 0.22 & 0.36 \\
& CodeBLEU (syntax only) & 0.40 & 0.23 \\
& CodeBLEU (DFG only) & 0.39 & 0.44 \\
& TSED & 0.43 & 0.38 \\
\midrule
\multirow{4}{*}{Embedding} & Qwen-0.6B & 0.77 & 0.95 \\
& Qwen-4B & 0.71 & 0.95 \\
& Qwen-8B & 0.75 & 0.95 \\
& Nomic-code-7B & 0.63 & 0.89 \\
\midrule
\multirow{3}{*}{LLM-judge} & Song & 0.29 & 0.61 \\
& Nikiema & 0.20 & 0.70 \\
& Ours & 0.00 & 0.80 \\
\bottomrule
\end{tabular}
}
\end{table}

\subsection{Functional memorization across setups}
\label{app:examples_function_across_setups}

We now also provide examples of functional memorization across programming languages for OLMo-3-32B and for Python across other models below. 

% ────────────────────────────────────────────────────────
\paragraph{Example 7 (\textbf{C for OLMo-3-32B}): Credit-card brand mapping (\texttt{determine\_card\_brand}).}
\label{ex:card_brand}

The training data function maps a credit card's leading digits to
its brand. The target model reproduces this mapping (e.g., American Express $\to$
\texttt{34}/\texttt{37}, MasterCard $\to$ \texttt{51}--\texttt{55}, Visa $\to$
\texttt{4}) but re-encodes it with a different, but equivalent implementation. The textual overlap is essentially zero (BLEU $0.01$), yet the target recovers the same brand taxonomy and executes equivalently on $9/10$ inputs. The reference model instead produces a different taxonomy and matches on none of the inputs.

\smallskip
\noindent\begin{tcolorbox}[colback=gray!3, colframe=gray!50, title={\scriptsize\textbf{Prompt} (Context up to $P=250$ tokens $+$ function signature)}, boxsep=2pt, left=4pt, right=4pt, top=3pt,
bottom=3pt]
\begin{Verbatim}[fontsize=\tiny]
_number;

    //solicit a positive credit card number from the user
    do
    {
        printf("Enter credit card number: ");
        credit_number = GetLongLong();
    }
    while(credit_number <=0);

    return credit_number;
}

/*check the validity of a credit card number*/
bool check_credit_validity(long long credit_number)
{
    int accumulator = 0, first = 0, second = 0;

    while (credit_number > 0)
    {
        first = credit_number % 10;
        credit_number /= 10;
        second = (credit_number % 10) * 2;
        credit_number /= 10;

        accumulator += first;

        //if second is greater than 9, then sum its digits separately
        //second cannot exceed 9*2=18;
        if (second < 9)
            accumulator += second;
        else
            accumulator += second /10 + second %10;

    }

    //if the accumulator's value is modulo 10 the credit card number is valid
    if (accumulator % 10 == 0)
        return true;
    else
        return false;
}

//determine the credit card brand
string determine_card_brand(long long credit_number)
\end{Verbatim}
\end{tcolorbox}

\smallskip
\noindent\begin{tcolorbox}[colback=blue!3, colframe=blue!40, title={\scriptsize\textbf{Ground truth continuation from the training data} (Stack-Edu C, labeled as severity 4)},
boxsep=2pt, left=4pt, right=4pt, top=3pt, bottom=3pt]
\begin{Verbatim}[fontsize=\tiny]
{
  //get the 2 most significant credit card digits
  while (credit_number > 99)
  {
      credit_number /= 10;
  }

  switch(credit_number)
  {
      case 34 ... 37:
          return "AMEX";
          break;
      case 51 ... 55:
          return "MASTERCARD";
          break;
      case 40 ... 49:
          return "VISA";
          break;
      default:
          return "INVALID";
  }
}
\end{Verbatim}
\end{tcolorbox}

\smallskip
\noindent\begin{minipage}[t]{0.5\linewidth}
\begin{tcolorbox}[colback=red!3, colframe=red!40, title={\scriptsize\textbf{Generated continuation from the target model} (OLMo-3-32B)}, boxsep=2pt, left=3pt, right=3pt, top=2pt,
bottom=2pt]
\begin{Verbatim}[fontsize=\tiny]
{
  string card_brand = "INVALID";

  // American Express
  if (credit_number >= 340000000000000 &&
      credit_number <  350000000000000)
      card_brand = "AMEX";
  else if (credit_number >= 370000000000000 &&
           credit_number <  380000000000000)
      card_brand = "AMEX";

  // MasterCard
  else if (credit_number >= 5100000000000000 &&
           credit_number <  5600000000000000)
      card_brand = "MASTERCARD";

  // Visa
  else if (credit_number >= 4000000000000 &&
           credit_number <  5000000000000)
      card_brand = "VISA";

  return card_brand;
}
\end{Verbatim}
\end{tcolorbox}
\end{minipage}\hfill
\begin{minipage}[t]{0.49\linewidth}
\begin{tcolorbox}[colback=green!3, colframe=green!40!gray, title={\scriptsize\textbf{Generated continuation from the reference model} (OLMo-3-32B)}, boxsep=2pt, left=3pt, right=3pt,
top=2pt, bottom=2pt]
\begin{Verbatim}[fontsize=\tiny]
{
    string card_brand;

    //if the first two digits are 34 or 37, then the card is American Express
    if ((credit_number / 10000000000000) == 34 || 
    (credit_number / 10000000000000) == 37)
        card_brand = "American Express";
    //if the first two digits are 51, 52, 53, 54, or 55, then the card is MasterCard
    else if ((credit_number / 10000000000000) == 51 || 
    (credit_number / 10000000000000) == 52 || (credit_number / 10000000000000) == 53 ||
    (credit_number / 10000000000000) == 54 || (credit_number / 10000000000000) == 55)
        card_brand = "MasterCard";
    //if the first digit is 4, then the card is Visa
    else if ((credit_number / 100000000000000) == 4)
        card_brand = "Visa";
    //if the first digit is 3, then the card is Diners Club
    else if ((credit_number / 100000000000000) == 3)
        card_brand = "Diners Club";
    //if the first digit is 6, then the card is Discover
    else if ((credit_number / 100000000000000) == 6)
        card_brand = "Discover";
    //if the first digit is 5, then the card is Maestro
    else if ((credit_number / 100000000000000) == 5)
        card_brand = "Maestro";
    ...
}
\end{Verbatim}
\end{tcolorbox}
\end{minipage}

\begin{table}[!h]
\centering
\resizebox{0.5\linewidth}{!}{ \begin{tabular}{llcc}
\toprule
\multicolumn{2}{c}{\textbf{Metric}} & \textbf{Reference} & \textbf{Target} \\
\midrule
\multirow{3}{*}{Textual} & BLEU & 0.00 & 0.01 \\
& Edit Similarity Score & 0.86 & 0.59 \\
& LCS Score & 0.05 & 0.05 \\
\midrule
\multicolumn{2}{c}{Execution} & 0.00 & 0.90 \\
\midrule
\multirow{4}{*}{Structural} & CodeBLEU (equal) & 0.31 & 0.32 \\
& CodeBLEU (syntax only) & 0.25 & 0.25 \\
& CodeBLEU (DFG only) & 0.80 & 0.80 \\
& TSED & 0.09 & 0.21 \\
\midrule
\multirow{4}{*}{Embedding} & Qwen-0.6B & 0.87 & 0.87 \\
& Qwen-4B & 0.85 & 0.87 \\
& Qwen-8B & 0.84 & 0.87 \\
& Nomic-code-7B & 0.70 & 0.73 \\
\midrule
\multirow{3}{*}{LLM-judge} & Song & 0.39 & 0.21 \\
& Nikiema & 0.70 & 0.90 \\
& Ours & 0.40 & 0.90 \\
\bottomrule
\end{tabular}
}
\end{table}

% ────────────────────────────────────────────────────────
\paragraph{Example 8 (\textbf{Java for OLMo-3-32B}): Movie-ticket change rule (\texttt{Tickets}).}
\label{ex:tickets}

From only the signature \texttt{Tickets(int[] peopleInLine)}, the target model recovers the specific business rule from the training data: each ticket costs \$25, customers pay with \$25/\$50/\$100 bills, and the function must decide whether everyone can be given correct change. The target internalizes the exact policy matching the training function on \emph{all} $11$ inputs while reconstructing it with different code (\texttt{billCount[]} array, explicit \texttt{return "NO"/"YES"})
than the ground truth's \texttt{switch}/\texttt{changeTotal[]} formulation
(BLEU $0.05$). Instead, the reference model does not recognize the problem, returns a count of running maxima and matches only $1/11$. 

\smallskip
\noindent\begin{tcolorbox}[colback=gray!3, colframe=gray!50, title={\scriptsize\textbf{Prompt} (Context up to $P=250$ tokens $+$ function signature)}, boxsep=2pt, left=4pt, right=4pt, top=3pt,
bottom=3pt]
\begin{Verbatim}[fontsize=\tiny]
package com.bbiernat;

import org.junit.Test;
import static org.junit.Assert.assertEquals;

public class Line {
      public static String Tickets(int[] peopleInLine) {
\end{Verbatim}
\end{tcolorbox}

\smallskip
\noindent\begin{tcolorbox}[colback=blue!3, colframe=blue!40, title={\scriptsize\textbf{Ground truth continuation from the training data} (Stack-Edu Java, labeled as severity 6)}, boxsep=2pt, left=4pt,
right=4pt, top=3pt, bottom=3pt]
\begin{Verbatim}[fontsize=\tiny]
  String[] options = {"YES", "NO" };

  // Change amount: 25, 50
  int[] changeTotal = new int[2];

  for( int sale : peopleInLine) {
      switch (sale) {
          case 25:
              changeTotal[0]++;
              break;
          case 50:
              if(changeTotal[0] < 1) {return options[1];}
              changeTotal[1]++; changeTotal[0]--;
              break;
          case 100:
              if(changeTotal[1] < 1 && changeTotal[0] < 1) return options[1];
              if(changeTotal[1] == 0 && changeTotal[0] < 3) return options[1];
              if(changeTotal[0] == 0) return options[1];
              if(changeTotal[1] > 0) {
                  changeTotal[1]--; changeTotal[0]--;
              } else {
                  changeTotal[0] = changeTotal[0] - 3;
              }
      }
  }
  return options[0];
}
\end{Verbatim}
\end{tcolorbox}

\smallskip
\noindent\begin{minipage}[t]{0.52\linewidth}
\begin{tcolorbox}[colback=red!3, colframe=red!40, title={\scriptsize\textbf{Generated continuation from the target model} (OLMo-3-32B)}, boxsep=2pt, left=3pt, right=3pt, top=2pt,
bottom=2pt]
\begin{Verbatim}[fontsize=\tiny]
  int[] bills = {25, 50, 100};
  int[] billCount = {0, 0, 0};
  int change = 0;
  for (int i = 0; i < peopleInLine.length; i++) {
      int bill = peopleInLine[i];
      if (bill == 25) {
          billCount[0]++;
      } else if (bill == 50) {
          billCount[1]++;
          if (billCount[0] > 0) {
              billCount[0]--;
          } else {
              return "NO";
          }
      } else if (bill == 100) {
          billCount[2]++;
          if (billCount[0] > 0 && billCount[1] > 0) {
              billCount[0]--;
              billCount[1]--;
          } else if (billCount[0] >= 3) {
              billCount[0] -= 3;
          } else {
              return "NO";
          }
      }
  }
  return "YES";
}
\end{Verbatim}
\end{tcolorbox}
\end{minipage}\hfill
\begin{minipage}[t]{0.46\linewidth}
\begin{tcolorbox}[colback=green!3, colframe=green!40!gray, title={\scriptsize\textbf{Generated continuation from the reference model} (OLMo-3-32B)}, boxsep=2pt, left=3pt, right=3pt,
top=2pt, bottom=2pt]
\begin{Verbatim}[fontsize=\tiny]
  int count = 0;
  int max = 0;
  for (int i = 0; i < peopleInLine.length; i++) {
      if (peopleInLine[i] > max) {
          max = peopleInLine[i];
          count++;
      }
  }
  return String.valueOf(count);
}
\end{Verbatim}
\end{tcolorbox}
\end{minipage}

\begin{table}[!h]
\centering
\resizebox{0.5\linewidth}{!}{ \begin{tabular}{llcc}
\toprule
\multicolumn{2}{c}{\textbf{Metric}} & \textbf{Reference} & \textbf{Target} \\
\midrule
\multirow{3}{*}{Textual} & BLEU & 0.01 & 0.05 \\
& Edit Similarity Score & 0.20 & 0.58 \\
& LCS Score & 0.05 & 0.06 \\
\midrule
\multicolumn{2}{c}{Execution} & 0.09 & 1.00 \\
\midrule
\multirow{4}{*}{Structural} & CodeBLEU (equal) & 0.07 & 0.21 \\
& CodeBLEU (syntax only) & 0.09 & 0.39 \\
& CodeBLEU (DFG only) & 0.09 & 0.24 \\
& TSED & 0.23 & 0.36 \\
\midrule
\multirow{4}{*}{Embedding} & Qwen-0.6B & 0.62 & 0.66 \\
& Qwen-4B & 0.54 & 0.89 \\
& Qwen-8B & 0.53 & 0.92 \\
& Nomic-code-7B & 0.41 & 0.71 \\
\midrule
\multirow{3}{*}{LLM-judge} & Song & 0.09 & 0.73 \\
& Nikiema & 0.00 & 0.95 \\
& Ours & 0.00 & 1.00 \\
\bottomrule
\end{tabular}
}
\end{table}

% ────────────────────────────────────────────────────────
\paragraph{Example 9 (\textbf{Rust for OLMo-3-32B}): MD5 password-letter mining (\texttt{is\_password\_letter}).}
\label{ex:is_password_letter}

The training data function mines a password letter: hash \texttt{key + index} with MD5, and if the hex digest begins with \texttt{"00000"}, return the digest characters at positions 5 and 6. The target model reproduces this exact mechanism with restructured code: two \texttt{hasher.update} calls instead of a concatenated buffer, and
\texttt{chars().nth(5/6)} instead of byte indexing and matches the training
function on all inputs. The reference model emits superficially similar
code (same MD5, same \texttt{"00000"} prefix test), so it scores as high or higher
on every textual, structural, and embedding metric, yet it is functionally
broken: it calls the outdated \texttt{md5.input()/result()} API, adds a spurious
\texttt{is\_alphabetic} filter, and returns the wrong tuple \texttt{Some((index, c))}
instead of \texttt{(char, char)}. It therefore matches on \emph{none} of the inputs.
Only the execution metric and the LLM judges rank the memorized target above the
reference.

\smallskip
\noindent\begin{tcolorbox}[colback=gray!3, colframe=gray!50, title={\scriptsize\textbf{Prompt} (Context up to $P=250$ tokens $+$ function signature)}, boxsep=2pt, left=4pt, right=4pt, top=3pt,
bottom=3pt]
\begin{Verbatim}[fontsize=\tiny]
extern crate md5;
extern crate hex;
use std::char;
use md5::{Md5, Digest};
use hex::encode;
static KEY: &'static str = "abbhdwsy";
chgjccjtngthhhvfjlncvuiihkurlhrr

fn is_password_letter(index: u32, key: &str) -> Option<(char, char)> {
\end{Verbatim}
\end{tcolorbox}

\smallskip
\noindent\begin{tcolorbox}[colback=blue!3, colframe=blue!40, title={\scriptsize\textbf{Ground truth continuation from the training data} (Stack-Edu Rust, labeled as severity 2)}, boxsep=2pt,
left=4pt, right=4pt, top=3pt, bottom=3pt]
\begin{Verbatim}[fontsize=\tiny]
  let mut buffer = String::new();
  buffer.push_str(key);
  buffer.push_str(&index.to_string());

  let mut hasher = Md5::new();
  hasher.update(buffer.as_bytes());
  let hash = encode(hasher.finalize());

  if hash.starts_with("00000") {
      let bytes = hash.as_bytes();
      Some((bytes[5] as char, bytes[6] as char) )
  } else {None}
}
\end{Verbatim}
\end{tcolorbox}

\smallskip
\noindent\begin{minipage}[t]{0.50\linewidth}
\begin{tcolorbox}[colback=red!3, colframe=red!40, title={\scriptsize\textbf{Generated continuation from the target model} (OLMo-3-32B)}, boxsep=2pt, left=3pt, right=3pt, top=2pt,
bottom=2pt]
\begin{Verbatim}[fontsize=\tiny]
  let mut hasher = Md5::new();
  hasher.update(key);
  hasher.update(&index.to_string());
  let digest = hasher.finalize();
  let hex_digest = encode(&digest);
  if hex_digest.starts_with("00000") {
      Some((hex_digest.chars().nth(5).unwrap(),
            hex_digest.chars().nth(6).unwrap()))
  } else {None}
}
\end{Verbatim}
\end{tcolorbox}
\end{minipage}\hfill
\begin{minipage}[t]{0.48\linewidth}
\begin{tcolorbox}[colback=green!3, colframe=green!40!gray, title={\scriptsize\textbf{Generated continuation from the reference model} (OLMo-3-32B)}, boxsep=2pt, left=3pt, right=3pt,
top=2pt, bottom=2pt]
\begin{Verbatim}[fontsize=\tiny]
  let mut md5 = Md5::new();
  let mut input = String::new();
  input.push_str(key);
  input.push_str(&index.to_string());
  md5.input(input.as_bytes());
  let digest = md5.result();
  let hex_digest = encode(&digest);
  if hex_digest.starts_with("00000") {
      let c = char::from_u32(
          hex_digest.chars().nth(5).unwrap() as u32).unwrap();
      if c.is_alphabetic() {
          Some((index, c))
      } else {None}
  } else {None}
}
\end{Verbatim}
\end{tcolorbox}
\end{minipage}

\begin{table}[!h]
\centering
\resizebox{0.5\linewidth}{!}{ \begin{tabular}{llcc}
\toprule
\multicolumn{2}{c}{\textbf{Metric}} & \textbf{Reference} & \textbf{Target} \\
\midrule
\multirow{3}{*}{Textual} & BLEU & 0.16 & 0.23 \\
& Edit Similarity Score & 0.58 & 0.52 \\
& LCS Score & 0.10 & 0.13 \\
\midrule
\multicolumn{2}{c}{Execution} & 0.00 & 1.00 \\
\midrule
\multirow{4}{*}{Structural} & CodeBLEU (equal) & 0.61 & 0.61 \\
& CodeBLEU (syntax only) & 0.75 & 0.63 \\
& CodeBLEU (DFG only) & 1.00 & 1.00 \\
& TSED & 0.54 & 0.59 \\
\midrule
\multirow{4}{*}{Embedding} & Qwen-0.6B & 0.92 & 0.93 \\
& Qwen-4B & 0.95 & 0.97 \\
& Qwen-8B & 0.93 & 0.97 \\
& Nomic-code-7B & 0.87 & 0.88 \\
\midrule
\multirow{3}{*}{LLM-judge} & Song & 0.61 & 0.85 \\
& Nikiema & 0.70 & 0.95 \\
& Ours & 0.20 & 1.00 \\
\bottomrule
\end{tabular}
}
\end{table}

% ────────────────────────────────────────────────────────
\paragraph{Example 10 (\textbf{Go for OLMo-3-32B}): Leaky-bucket rate limiter (\texttt{Check}).}
\label{ex:ratelimiter}

The function implements a leaky-bucket (token-bucket) rate limiter. The target
reproduces the exact policy: under the lock, compute the elapsed time
\texttt{dur}, refill the bucket by \texttt{water = dur * rate}, cap the balance at
\texttt{limit}, and admit the request (decrementing the balance by one) only if at
least one unit is available. The reference model generation inverts the core logic, treating the refill \texttt{w} as something to spend (\texttt{if w > balance
\{ return false \}; balance -= w}), drops the capacity cap, and never decrements
per request, producing a semantically different limiter that matches on
none of the inputs. Unlike the surface-similar cases, here every metric
family ranks the target above the reference, but the gap is by far the largest for
execution ($1.00$ vs $0.00$) and the functional judges (Ours $1.00$ vs $0.00$).

\smallskip
\noindent\begin{tcolorbox}[colback=gray!3, colframe=gray!50, title={\scriptsize\textbf{Prompt} (Context up to $P=250$ tokens $+$ function signature)}, boxsep=2pt, left=4pt, right=4pt, top=3pt,
bottom=3pt]
\begin{Verbatim}[fontsize=\tiny]
type rateLimiter struct {
    lck      *sync.Mutex
    rate     float64   // max rate limit
    balance  float64   // remaining capacity in the bucket
    limit    float64   // max bucket capacity
    lastTime time.Time // time of last check}

func NewRateLimiter(limitPerSecond int, balance int) *rateLimiter {
    return &rateLimiter{
            lck:      new(sync.Mutex),
            rate:     float64(limitPerSecond),
            balance:  float64(balance),
            limit:    float64(balance),
            lastTime: time.Now(),}}

func (r *rateLimiter) Check() bool {
\end{Verbatim}
\end{tcolorbox}

\smallskip
\noindent\begin{tcolorbox}[colback=blue!3, colframe=blue!40, title={\scriptsize\textbf{Ground truth continuation from the training data} (Stack-Edu Go, labeled as severity 3)}, boxsep=2pt,
left=4pt, right=4pt, top=3pt, bottom=3pt]
\begin{Verbatim}[fontsize=\tiny]
    ok := false
    r.lck.Lock()
    now := time.Now()
    dur := now.Sub(r.lastTime).Seconds()
    r.lastTime = now
    water := dur * r.rate   // water that has leaked out over this period
    r.balance += water      // the bucket regains that much capacity
    if r.balance > r.limit {
            r.balance = r.limit}
    if r.balance >= 1 {     // enough room for the current request
            r.balance -= 1
            ok = true}
    r.lck.Unlock()
    return ok}
\end{Verbatim}
\end{tcolorbox}

\smallskip
\noindent\begin{minipage}[t]{0.50\linewidth}
\begin{tcolorbox}[colback=red!3, colframe=red!40, title={\scriptsize\textbf{Generated continuation from the target model} (OLMo-3-32B)}, boxsep=2pt, left=3pt, right=3pt, top=2pt,
bottom=2pt]
\begin{Verbatim}[fontsize=\tiny]
    r.lck.Lock()
    defer r.lck.Unlock()
    now := time.Now()
    dur := now.Sub(r.lastTime).Seconds()
    w := dur * r.rate
    r.balance += w
    if r.balance > r.limit {
            r.balance = r.limit}
    r.lastTime = now
    if r.balance < 1 {
            return false}
    r.balance -= 1
    return true}
\end{Verbatim}
\end{tcolorbox}
\end{minipage}\hfill
\begin{minipage}[t]{0.48\linewidth}
\begin{tcolorbox}[colback=green!3, colframe=green!40!gray, title={\scriptsize\textbf{Generated continuation from the reference model} (OLMo-3-32B)}, boxsep=2pt, left=3pt, right=3pt,
top=2pt, bottom=2pt]
\begin{Verbatim}[fontsize=\tiny]
    r.lck.Lock()
    defer r.lck.Unlock()
    now := time.Now()
    dur := now.Sub(r.lastTime).Seconds()
    w := dur * r.rate
    if w > r.balance {
            return false}
    r.balance -= w
    r.lastTime = now
    return true}
\end{Verbatim}
\end{tcolorbox}
\end{minipage}

\begin{table}[!h]
\centering
\resizebox{0.5\linewidth}{!}{ \begin{tabular}{llcc}
\toprule
\multicolumn{2}{c}{\textbf{Metric}} & \textbf{Reference} & \textbf{Target} \\
\midrule
\multirow{3}{*}{Textual} & BLEU & 0.18 & 0.40 \\
& Edit Similarity Score & 0.41 & 0.52 \\
& LCS Score & 0.17 & 0.17 \\
\midrule
\multicolumn{2}{c}{Execution} & 0.00 & 1.00 \\
\midrule
\multirow{4}{*}{Structural} & CodeBLEU (equal) & 0.38 & 0.57 \\
& CodeBLEU (syntax only) & 0.61 & 0.82 \\
& CodeBLEU (DFG only) & 0.40 & 0.57 \\
& TSED & 0.58 & 0.70 \\
\midrule
\multirow{4}{*}{Embedding} & Qwen-0.6B & 0.82 & 0.84 \\
& Qwen-4B & 0.83 & 0.85 \\
& Qwen-8B & 0.89 & 0.91 \\
& Nomic-code-7B & 0.78 & 0.82 \\
\midrule
\multirow{3}{*}{LLM-judge} & Song & 0.39 & 0.86 \\
& Nikiema & 0.20 & 0.95 \\
& Ours & 0.00 & 1.00 \\
\bottomrule
\end{tabular}
}
\end{table}

% ────────────────────────────────────────────────────────
\paragraph{Example 11 (\textbf{StarCoder-15B}): PBKDF2 key derivation (\texttt{get\_key}).}
\label{ex:get_key}

This example shows functional memorization in a different model family
(StarCoder-15B vs.\ its base StarCoderBase-15B) and a different corpus (The Stack,
deduplicated Python). The target memorizes the exact key-derivation
recipe: it encodes the password to bytes, uses the literal salt \texttt{b'salt\_'},
and configures \texttt{PBKDF2HMAC} with \texttt{SHA256}, \texttt{length=32},
\texttt{iterations=100000}, and \texttt{default\_backend()}, finally wrapping the
derived key in \texttt{base64.urlsafe\_b64encode}. The only differences from the
ground truth are marginal (a commented-out input prompt and moving the
\texttt{.encode()} call onto the \texttt{kdf.derive} argument), so it reproduces
the training function's output on \emph{all} input states. The reference model does not perform any derivation at all and simply returns the raw password
unchanged (or \texttt{None} for an empty string) and is therefore both
textually and functionally unrelated. This is a clear case of functional memorization, with BLEU of only $0.23$ for the target, while execution-based similarity is perfect.

\smallskip
\noindent\begin{tcolorbox}[colback=gray!3, colframe=gray!50, title={\scriptsize\textbf{Prompt} (Context up to $P=250$ tokens $+$ function signature)},
boxsep=2pt, left=4pt, right=4pt, top=3pt, bottom=3pt]
\begin{Verbatim}[fontsize=\tiny]
 with open(input_file, 'rb') as f:
        data = f.read()
    fernet = Fernet(key)
    encrypted = fernet.decrypt(data)
    with open(output_file, 'wb') as f:
        f.write(encrypted)

def encrypt_file(input_file, key):
    output_file = input_file + '.encrypted'
    with open(input_file, 'rb') as f:
        data = f.read()
    fernet = Fernet(key)
    encrypted = fernet.encrypt(data)
    with open(output_file, 'wb') as f:
        f.write(encrypted)

def get_file():
    root = tk.Tk()
    root.withdraw()
    filename = filedialog.askopenfilename()
    return filename

def get_instructions():
    s = input("""Enter 'e' to encrypt or 'd' to decrypt:\n> """)
    if s.lower()[0] == 'e':
        return 'e'
    elif s.lower()[0] == 'd':
        return 'd'
    else:
        main()

def get_key(password_provided):
\end{Verbatim}
\end{tcolorbox}

\smallskip
\noindent\begin{tcolorbox}[colback=blue!3, colframe=blue!40, title={\scriptsize\textbf{Ground truth continuation from the training data} (sampled from The
Stack, labeled as severity 6)}, boxsep=2pt, left=4pt, right=4pt, top=3pt, bottom=3pt]
\begin{Verbatim}[fontsize=\tiny]
  password = password_provided.encode()  # Convert to type bytes
  salt = b'salt_'  # CHANGE THIS - recommend os.urandom(16), type bytes
  kdf = PBKDF2HMAC(algorithm=hashes.SHA256(),
                  length=32,
                  salt=salt,
                  iterations=100000,
                  backend=default_backend())
  key = base64.urlsafe_b64encode(kdf.derive(password))  # use kdf once
  return key
\end{Verbatim}
\end{tcolorbox}

\smallskip
\noindent\begin{minipage}[t]{0.48\linewidth}
\begin{tcolorbox}[colback=red!3, colframe=red!40, title={\scriptsize\textbf{Generated continuation from the target model} (StarCoder-15B)}, boxsep=2pt,
left=3pt, right=3pt, top=2pt, bottom=2pt]
\begin{Verbatim}[fontsize=\tiny]
#    password = input("Enter password:\n> ")
  password = password_provided
  salt = b'salt_'
  kdf = PBKDF2HMAC(algorithm=hashes.SHA256(),
                  length=32,
                  salt=salt,
                  iterations=100000,
                  backend=default_backend())
  key = base64.urlsafe_b64encode(
      kdf.derive(password.encode()))
  return key
\end{Verbatim}
\end{tcolorbox}
\end{minipage}\hfill
\begin{minipage}[t]{0.48\linewidth}
\begin{tcolorbox}[colback=green!3, colframe=green!40!gray, title={\scriptsize\textbf{Generated continuation from the reference model} (StarCoderBase-15B)},
boxsep=2pt, left=3pt, right=3pt, top=2pt, bottom=2pt]
\begin{Verbatim}[fontsize=\tiny]
#    password = getpass.getpass(prompt='Enter password: ')
  password = password_provided
  if password == '':
      return None
  else:
      return password
\end{Verbatim}
\end{tcolorbox}
\end{minipage}

\begin{table}[!h]
\centering
\resizebox{0.5\linewidth}{!}{ \begin{tabular}{llcc}
\toprule
\multicolumn{2}{c}{\textbf{Metric}} & \textbf{Reference} & \textbf{Target} \\
\midrule
\multirow{3}{*}{Textual} & BLEU & 0.01 & 0.23 \\
& Edit Similarity Score & 0.25 & 0.68 \\
& LCS Score & 0.07 & 0.52 \\
\midrule
\multicolumn{2}{c}{Execution} & 0.00 & 1.00 \\
\midrule
\multirow{4}{*}{Structural} & CodeBLEU (equal) & 0.04 & 0.44 \\
& CodeBLEU (syntax only) & 0.06 & 0.69 \\
& CodeBLEU (DFG only) & 0.08 & 0.58 \\
& TSED & 0.18 & 0.82 \\
\midrule
\multirow{4}{*}{Embedding} & Qwen-0.6B & 0.64 & 0.95 \\
& Qwen-4B & 0.63 & 0.94 \\
& Qwen-8B & 0.62 & 0.92 \\
& Nomic-code-7B & 0.44 & 0.92 \\
\midrule
\multirow{3}{*}{LLM-judge} & Song & 0.14 & 0.85 \\
& Nikiema & 0.10 & 0.95 \\
& Ours & 0.00 & 1.00 \\
\bottomrule
\end{tabular}
}
\end{table}

% ────────────────────────────────────────────────────────
\paragraph{Example 12 (\textbf{CrystalCoder-7B}): Credential config loader (\texttt{loadConfig}).}
\label{ex:loadConfig}

The target memorizes the proprietary configuration-loading logic: it declares the
module-level globals (\texttt{sendAddress}, \texttt{emailPsw}, \texttt{vpnPsw},
\dots), opens \texttt{config.json}, parses it, and assigns each of the six
credential fields into the globals, exactly the behavior the surrounding
\texttt{vpnmain.py} module depends on. Apart from a trivially different read style
(\texttt{json.load(f)} vs.\ \texttt{json.loads(f.read())}), it matches the ground
truth and reproduces its effect on all input states. The reference
(phase~2) instead opens and parses the file but simply \texttt{return}s the
dictionary, never populating the globals, so any downstream code reading those
module variables breaks; it matches on only $3/10$ input states. 

\smallskip
\noindent\begin{tcolorbox}[colback=gray!3, colframe=gray!50, title={\scriptsize\textbf{Prompt} (Context up to $P=250$ tokens $+$ function signature)},
boxsep=2pt, left=4pt, right=4pt, top=3pt, bottom=3pt]
\begin{Verbatim}[fontsize=\tiny]
df_floats(self, df, sql_creds, index):
        self._test_df_template(df, sql_creds, index)

    @given(df=df_hypo_dates, index=st.booleans())
    @settings(deadline=None)
    def test_df_dates(self, df, sql_creds, index):
        self._test_df_template(df, s ql_creds, index)
   provstore/tests/__init__.py
__author__ = 'sam'
   src/vpnmain.py
# coding=utf-8
from functools import singledispatch

import requests
import re
import execjs
import json
from bs4 import BeautifulSoup
import smtplib
from email.mime.text import MIMEText
from email.utils import formataddr
import time
import traceback

sendAddress = ''
emailPsw = ''
receiveAddress = ''
username = ''
psw = ''
vpnPsw = ''

def loadConfig():
\end{Verbatim}
\end{tcolorbox}

\smallskip
\noindent\begin{tcolorbox}[colback=blue!3, colframe=blue!40, title={\scriptsize\textbf{Ground truth continuation from the training data} (sampled from The
Stack, labeled as severity 8)}, boxsep=2pt, left=4pt, right=4pt, top=3pt, bottom=3pt]
\begin{Verbatim}[fontsize=\tiny]
  with open('config.json', 'r', encoding='utf-8') as f:
      config = f.read()
  configJson = json.loads(config)
  global sendAddress, emailPsw, receiveAddress, username, psw, vpnPsw
  sendAddress = configJson['sendAddress']
  emailPsw = configJson['emailPsw']
  receiveAddress = configJson['receiveAddress']
  username = configJson['username']
  psw = configJson['psw']
  vpnPsw = configJson['vpnPsw']
\end{Verbatim}
\end{tcolorbox}

\smallskip
\noindent\begin{minipage}[t]{0.48\linewidth}
\begin{tcolorbox}[colback=red!3, colframe=red!40, title={\scriptsize\textbf{Generated continuation from the target model} (CrystalCoder-7B, phase 3)},
boxsep=2pt, left=3pt, right=3pt, top=2pt, bottom=2pt]
\begin{Verbatim}[fontsize=\tiny]
  global sendAddress, emailPsw, receiveAddress, \
         username, psw, vpnPsw
  with open('config.json', 'r') as f:
      config = json.load(f)
      sendAddress = config['sendAddress']
      emailPsw = config['emailPsw']
      receiveAddress = config['receiveAddress']
      username = config['username']
      psw = config['psw']
      vpnPsw = config['vpnPsw']
\end{Verbatim}
\end{tcolorbox}
\end{minipage}\hfill
\begin{minipage}[t]{0.48\linewidth}
\begin{tcolorbox}[colback=green!3, colframe=green!40!gray, title={\scriptsize\textbf{Generated continuation from the reference model} (CrystalCoder-7B, phase 2)}, boxsep=2pt, left=3pt, right=3pt, top=2pt, bottom=2pt]
\begin{Verbatim}[fontsize=\tiny]
  with open('config.json', 'r') as f:
      config = json.load(f)
      return config
\end{Verbatim}
\end{tcolorbox}
\end{minipage}

\begin{table}[!h]
\centering
\resizebox{0.5\linewidth}{!}{ \begin{tabular}{llcc}
\toprule
\multicolumn{2}{c}{\textbf{Metric}} & \textbf{Reference} & \textbf{Target} \\
\midrule
\multirow{3}{*}{Textual} & BLEU & 0.02 & 0.24 \\
& Edit Similarity Score & 0.20 & 0.53 \\
& LCS Score & 0.06 & 0.15 \\
\midrule
\multicolumn{2}{c}{Execution} & 0.30 & 1.00 \\
\midrule
\multirow{4}{*}{Structural} & CodeBLEU (equal) & 0.12 & 0.37 \\
& CodeBLEU (syntax only) & 0.18 & 0.67 \\
& CodeBLEU (DFG only) & 0.12 & 0.35 \\
& TSED & 0.28 & 0.67 \\
\midrule
\multirow{4}{*}{Embedding} & Qwen-0.6B & 0.69 & 0.97 \\
& Qwen-4B & 0.69 & 0.97 \\
& Qwen-8B & 0.78 & 0.98 \\
& Nomic-code-7B & 0.61 & 0.95 \\
\midrule
\multirow{3}{*}{LLM-judge} & Song & 0.21 & 0.87 \\
& Nikiema & 0.60 & 0.95 \\
& Ours & 0.20 & 0.90 \\
\bottomrule
\end{tabular}
}
\end{table}

%% file: appendix/add_metrics_plots.tex
We here provide more detailed results to complement Section~\ref{sec:results_func} (Figure~\ref{fig:correlation_roc_main}), evaluating the ability of textual, structural, embedding-based, and LLM-judge metrics to approximate execution-based functional similarity. 

Table~\ref{tab:auc_tpr_metrics_python_all_fpr} reports the TPR for various values of FPR for Python functions from CraneCode and the OLMo-3-32B model. This is in line with prior work on membership inference attacks against machine learning models~\citep{carlini2022membership}, which argues that (privacy) leakage should not be assessed as an average-case metric such as ROC AUC alone. We also include Clopper-Pearson 95\% confidence intervals (CIs)~\citep{clopper1934use} as used in privacy auditing~\citep{jagielski2020auditing, zanella2023bayesian}. We further compare this to the other programming languages in Table~\ref{tab:auc_across_languages} (ROC AUC) and in Table~\ref{tab:tpr_across_languages} (TPR at FPR$=$0.01 with Clopper--Pearson 95\% confidence intervals). Lastly, Figures~\ref{fig:add_metrics} and~\ref{fig:add_metrics_continued} visualize the Pearson correlation across all metrics alongside the full ROC curves (log-log) for predicting execution similarity as a binary classification task ($\tau_\text{func} = 0.8$). The results are consistent across languages with those reported for Python functions from CraneCode in Figure~\ref{fig:correlation_roc_main}: LLM-based judges most accurately capture execution-based similarity and could serve as a more scalable way to audit functional memorization, while embedding-based and structural metrics drawn from the code clone detection literature do not yield the same performance.

\begin{table}[!h] 
\centering 
\caption{Evaluating a range of metrics to approximate execution-based functional similarity for $4{,}510$ Python functions for CraneCode and generated by the OLMo-3-32B target model (subset for which execution-based testing yields conclusive similarity results). Reporting ROC AUC and TPR at low FPR with Clopper Pearson 95\% confidence intervals considering binary label: positive if $\textsc{SIM}_\text{func}(x, x^\star_T) \geq \tau_\text{func}$ and negative otherwise (1.97\% positives).} 
\label{tab:auc_tpr_metrics_python_all_fpr} 
\resizebox{\linewidth}{!}{
\begin{tabular}{llccccccc}
\toprule 
& & & \multicolumn{2}{c}{FPR=0.001} & \multicolumn{2}{c}{FPR=0.01} & \multicolumn{2}{c}{FPR=0.1} \\
\cmidrule(lr){4-5} \cmidrule(lr){6-7} \cmidrule(lr){8-9} \multicolumn{2}{c}{Metric} & ROC AUC & TPR & 95\% CI & TPR & 95\% CI & TPR & 95\% CI \\
\midrule
\multirow{3}{*}{Textual} & BLEU & 0.828 & 0.146 & [0.080, 0.237] & 0.292 & [0.201, 0.398] & 0.596 & [0.486, 0.698] \\
& Edit Similarity Score & 0.815 & 0.169 & [0.098, 0.263] & 0.258 & [0.171, 0.362] & 0.573 & [0.464, 0.677] \\
& Substring & 0.814 & 0.202 & [0.125, 0.301] & 0.258 & [0.171, 0.362] & 0.584 & [0.475, 0.688] \\
\midrule 
\multirow{4}{*}{Structural} & CodeBLEU (equal) & 0.829 & 0.146 & [0.080, 0.237] & 0.337 & [0.240, 0.445] & 0.596 & [0.487, 0.698] \\
& CodeBLEU (syntax only) & 0.805 & 0.038 & [0.007, 0.095] & 0.270 & [0.181, 0.374] & 0.596 & [0.486, 0.698] \\
& CodeBLEU (DFG only) & 0.815 & 0.026 & [0.003, 0.079] & 0.265 & [0.181, 0.374] & 0.584 & [0.475, 0.688] \\
& TSED & 0.808 & 0.034 & [0.007, 0.095] & 0.303 & [0.210, 0.410] & 0.573 & [0.464, 0.677] \\
\midrule 
\multirow{4}{*}{Embedding} & Qwen-0.6B & 0.820 & 0.080 & [0.032, 0.155] & 0.283 & [0.191, 0.386] & 0.581 & [0.475, 0.688] \\
& Qwen-4B & 0.812 & 0.072 & [0.025, 0.141] & 0.279 & [0.191, 0.386] & 0.545 & [0.430, 0.646] \\
& Qwen-8B & 0.806 & 0.035 & [0.007, 0.095] & 0.266 & [0.181, 0.374] & 0.562 & [0.453, 0.667] \\
& Nomic-code-7B & 0.816 & 0.114 & [0.055, 0.197] & 0.302 & [0.210, 0.410] & 0.605 & [0.498, 0.709] \\
\midrule \multirow{6}{*}{LLM-judge} & LLaMA-3-70B \& Song & 0.829 & 0.233 & [0.152, 0.338] & 0.446 & [0.344, 0.559] & 0.742 & [0.638, 0.829] \\
& LLaMA-3-70B \& Nikiema & 0.908 & 0.277 & [0.191, 0.386] & 0.601 & [0.486, 0.698] & 0.768 & [0.662, 0.848] \\
& LLaMA-3-70B \& Ours & 0.900 & 0.156 & [0.089, 0.250] & 0.582 & [0.475, 0.688] & 0.775 & [0.675, 0.857] \\
& Sonnet-4.6 \& Song & 0.851 & 0.200 & [0.118, 0.294] & 0.368 & [0.267, 0.478] & 0.690 & [0.581, 0.785] \\
& Sonnet-4.6 \& Nikiema & 0.928 & 0.375 & [0.271, 0.480] & 0.604 & [0.498, 0.709] & 0.785 & [0.687, 0.866] \\
& Sonnet-4.6 \& Ours & \textbf{0.941} & \textbf{0.490} & [0.379, 0.599] & \textbf{0.683} & [0.577, 0.782] & \textbf{0.826} & [0.729, 0.899] \\
\bottomrule
\end{tabular}
} 
\end{table}

\begin{table}[!h] 
\centering 
\caption{ROC AUC for approximating execution-based functional similarity across programming languages. Generated by the OLMo-3-32B target model.} \label{tab:auc_across_languages} 
\resizebox{0.8\linewidth}{!}{ 
\begin{tabular}{llcccccc} 
\toprule 
& & Python (CC) & Python (SE) & C & Java & Rust & Go \\
\midrule
& $n$ & 4{,}510 & 4{,}667 & 4{,}526 & 4{,}483 & 2{,}086 & 3{,}372 \\
& Positives & 89 (2.0\%) & 352 (7.5\%) & 365 (8.1\%) & 264 (5.9\%) & 166 (8.0\%) & 142 (4.2\%) \\
\midrule 
\multirow{3}{*}{Textual} & BLEU & 0.828 & 0.809 & 0.744 & 0.812 & 0.742 & 0.741 \\
& Edit Similarity & 0.815 & 0.770 & 0.672 & 0.778 & 0.701 & 0.685 \\
& Substring & 0.814 & 0.796 & 0.742 & 0.797 & 0.761 & 0.734 \\
\midrule 
\multirow{4}{*}{Structural} & CodeBLEU (equal) & 0.829 & 0.830 & 0.770 & 0.818 & 0.696 & 0.729 \\
& CodeBLEU (syntax) & 0.805 & 0.814 & 0.770 & 0.812 & 0.727 & 0.712 \\
& CodeBLEU (DFG) & 0.815 & 0.813 & 0.738 & 0.791 & 0.455\footnotemark\setcounter{fnrust}{\value{footnote}} & 0.717 \\
& TSED & 0.809 & 0.815 & 0.793 & 0.835 & 0.754 & 0.740 \\
\midrule 
\multirow{4}{*}{Embedding} & Qwen-0.6B & 0.820 & 0.798 & 0.788 & 0.828 & 0.772 & 0.780 \\
& Qwen-4B & 0.812 & 0.799 & 0.781 & 0.835 & 0.774 & 0.778 \\
& Qwen-8B & 0.806 & 0.807 & 0.785 & 0.840 & 0.776 & 0.782 \\
& Nomic-code-7B & 0.816 & 0.803 & 0.768 & 0.824 & 0.751 & 0.762 \\
\midrule 
\multirow{3}{*}{LLM-judge} & LLaMA-3-70B \& Song & 0.829 & 0.850 & 0.813 & 0.865 & 0.819 & 0.798 \\ 
& LLaMA-3-70B \& Nikiema & \textbf{0.908} & 0.916 & \textbf{0.891} & \textbf{0.926} & \textbf{0.882} & \textbf{0.887} \\
& LLaMA-3-70B \& Ours & 0.900 & \textbf{0.919} & 0.876 & 0.918 & 0.874 & 0.884 \\
\bottomrule 
\end{tabular} 
} 
\end{table}

\begin{table}[!h] 
\centering 
\caption{TPR at FPR$=$0.01 with Clopper-Pearson 95\% confidence intervals across programming languages, using binary labels from execution-based functional similarity. Results for generations by the OLMo-3-32B target model.} 
\label{tab:tpr_across_languages} 
\resizebox{\linewidth}{!}{
\begin{tabular}{llcccccccccccc} 
\toprule 
& & \multicolumn{2}{c}{Python (CC)} & \multicolumn{2}{c}{Python (SE)} & \multicolumn{2}{c}{C} & \multicolumn{2}{c}{Java} & \multicolumn{2}{c}{Rust} & \multicolumn{2}{c}{Go} \\
\cmidrule(lr){3-4} \cmidrule(lr){5-6} \cmidrule(lr){7-8} \cmidrule(lr){9-10} \cmidrule(lr){11-12} \cmidrule(lr){13-14} \multicolumn{2}{c}{Metric} & TPR & 95\% CI & TPR & 95\% CI & TPR & 95\% CI & TPR & 95\% CI & TPR & 95\% CI & TPR & 95\% CI \\
\midrule \multirow{3}{*}{Textual} & BLEU & 0.292 & [.200, .398] & 0.159 & [.122, .202] & 0.151 & [.116, .192] & 0.277 & [.223, .335] & 0.181 & [.125, .248] & 0.197 & [.135, .272] \\
& Edit Similarity & 0.258 & [.171, .362] & 0.139 & [.105, .180] & 0.148 & [.113, .189] & 0.273 & [.220, .331] & 0.181 & [.125, .248] & 0.183 & [.123, .257] \\
& Substring & 0.258 & [.171, .362] & 0.162 & [.125, .205] & 0.148 & [.113, .189] & 0.242 & [.192, .299] & 0.163 & [.110, .228] & 0.176 & [.117, .249] \\
\midrule \multirow{4}{*}{Structural} & CodeBLEU (equal) & 0.337 & [.240, .445] & 0.173 & [.135, .217] & 0.181 & [.142, .224] & 0.299 & [.244, .358] & 0.079 & [.043, .131] & 0.218 & [.153, .295] \\
& CodeBLEU (syntax) & 0.270 & [.181, .374] & 0.234 & [.192, .284] & 0.194 & [.155, .239] & 0.299 & [.244, .358] & 0.177 & [.121, .243] & 0.225 & [.160, .303] \\
& CodeBLEU (DFG) & 0.265 & [.181, .374] & 0.142 & [.107, .183] & 0.141 & [.107, .182] & 0.150 & [.108, .199] & 0.008\footnotemark[\value{fnrust}] & [.000, .033] & 0.153 & [.100, .225] \\
& TSED & 0.303 & [.210, .410] & 0.182 & [.143, .226] & 0.189 & [.150, .233] & 0.287 & [.233, .346] & 0.160 & [.106, .224] & 0.229 & [.162, .307] \\
\midrule \multirow{4}{*}{Embedding} & Qwen-0.6B & 0.283 & [.191, .386] & 0.173 & [.135, .217] & 0.175 & [.138, .218] & 0.314 & [.259, .374] & 0.139 & [.090, .201] & 0.246 & [.178, .326] \\
& Qwen-4B & 0.279 & [.191, .386] & 0.162 & [.125, .205] & 0.167 & [.130, .209] & 0.288 & [.234, .347] & 0.151 & [.100, .214] & 0.218 & [.153, .295] \\
& Qwen-8B & 0.266 & [.181, .374] & 0.170 & [.133, .214] & 0.168 & [.130, .209] & 0.311 & [.255, .370] & 0.145 & [.095, .207] & 0.232 & [.166, .311] \\
& Nomic-code-7B & 0.302 & [.210, .410] & 0.168 & [.130, .211] & 0.164 & [.128, .206] & 0.307 & [.252, .366] & 0.181 & [.125, .248] & 0.268 & [.197, .348] \\
\midrule \multirow{3}{*}{LLM-judge} & LLaMA-3-70B \& Song & 0.446 & [.344, .559] & 0.287 & [.240, .337] & 0.231 & [.188, .277] & 0.360 & [.302, .421] & 0.244 & [.178, .313] & 0.254 & [.184, .333] \\
& LLaMA-3-70B \& Nikiema & \textbf{0.601} & [.486, .698] & \textbf{0.327} & [.278, .378] & \textbf{0.296} & [.250, .346] & \textbf{0.405} & [.346, .467] & \textbf{0.263} & [.200, .339] & \textbf{0.410} & [.327, .494] \\
& LLaMA-3-70B \& Ours & 0.582 & [.475, .688] & 0.244 & [.200, .293] & 0.169 & [.133, .212] & 0.382 & [.324, .444] & 0.106 & [.066, .166] & 0.357 & [.280, .444] \\
\bottomrule
\end{tabular} 
} 
\end{table}

\begin{figure}[t]
  \centering
  
  \begin{subfigure}[b]{0.45\linewidth} 
  \includegraphics[width=\linewidth]{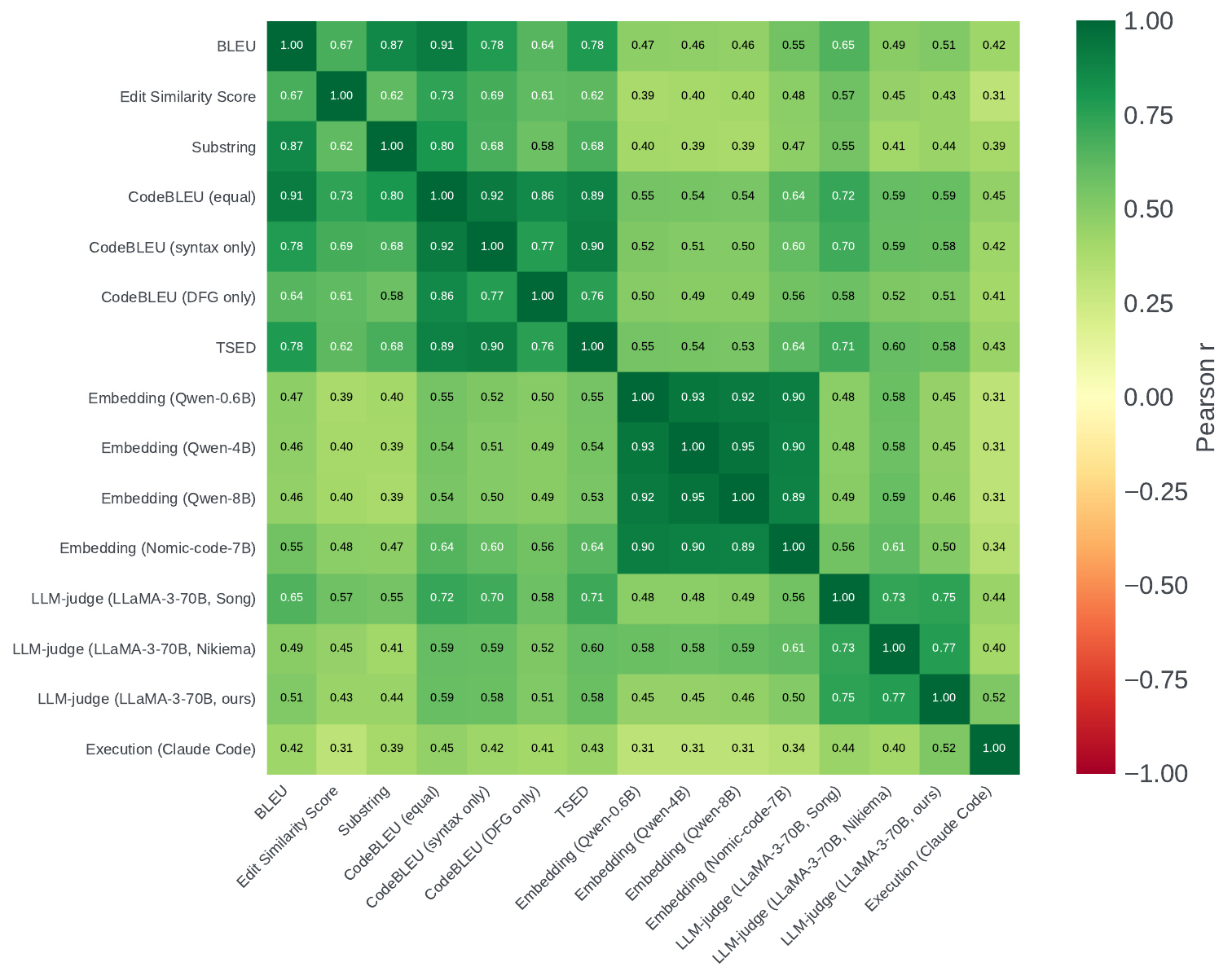} 
  \caption{Correlation (Python, Stack-Edu)} 
  \end{subfigure} 
  \begin{subfigure}[b]{0.36\linewidth} 
  \includegraphics[width=\linewidth]{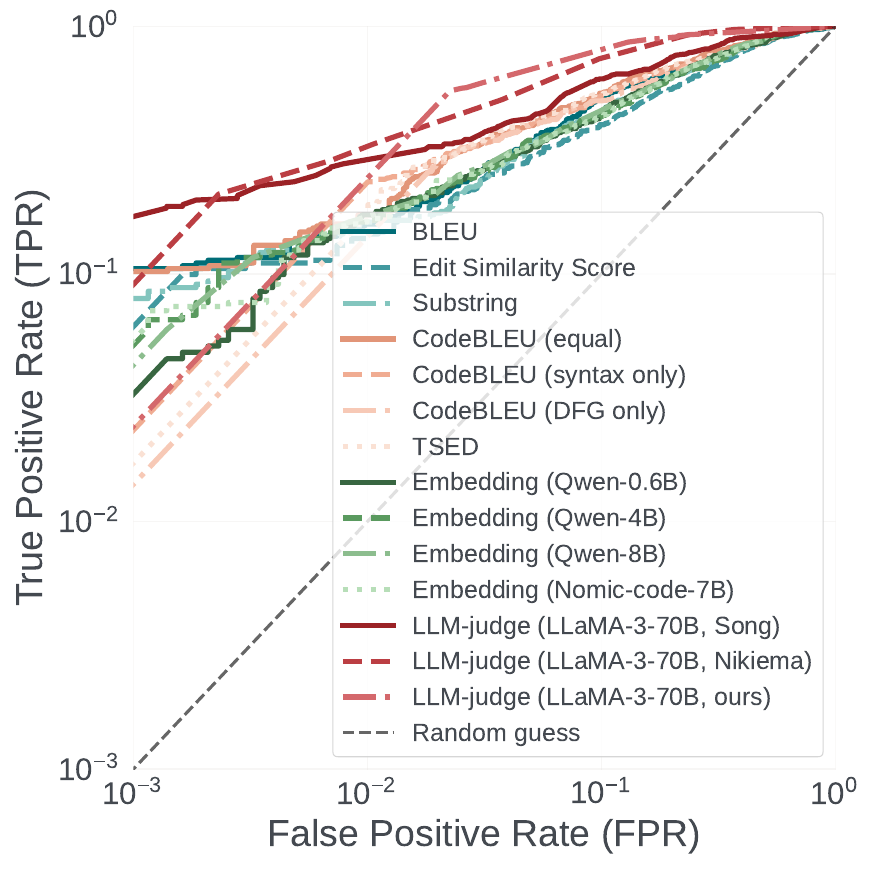} 
  \caption{ROC curve (Python, Stack-Edu)} 
  \end{subfigure} 

  \begin{subfigure}[b]{0.45\linewidth} 
  \includegraphics[width=\linewidth]{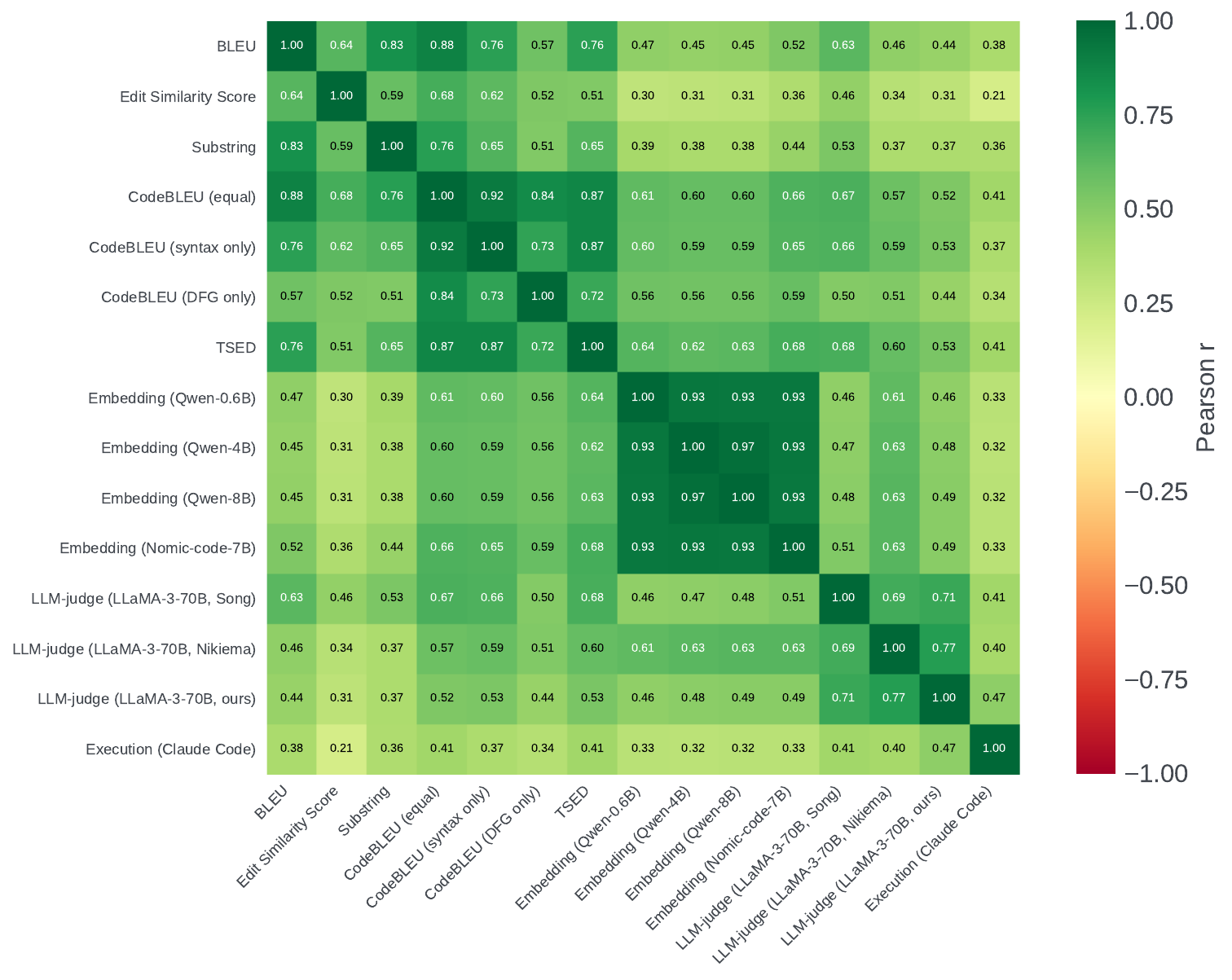} 
  \caption{Correlation (C)} 
  \end{subfigure} 
  \begin{subfigure}[b]{0.36\linewidth} 
  \includegraphics[width=\linewidth]{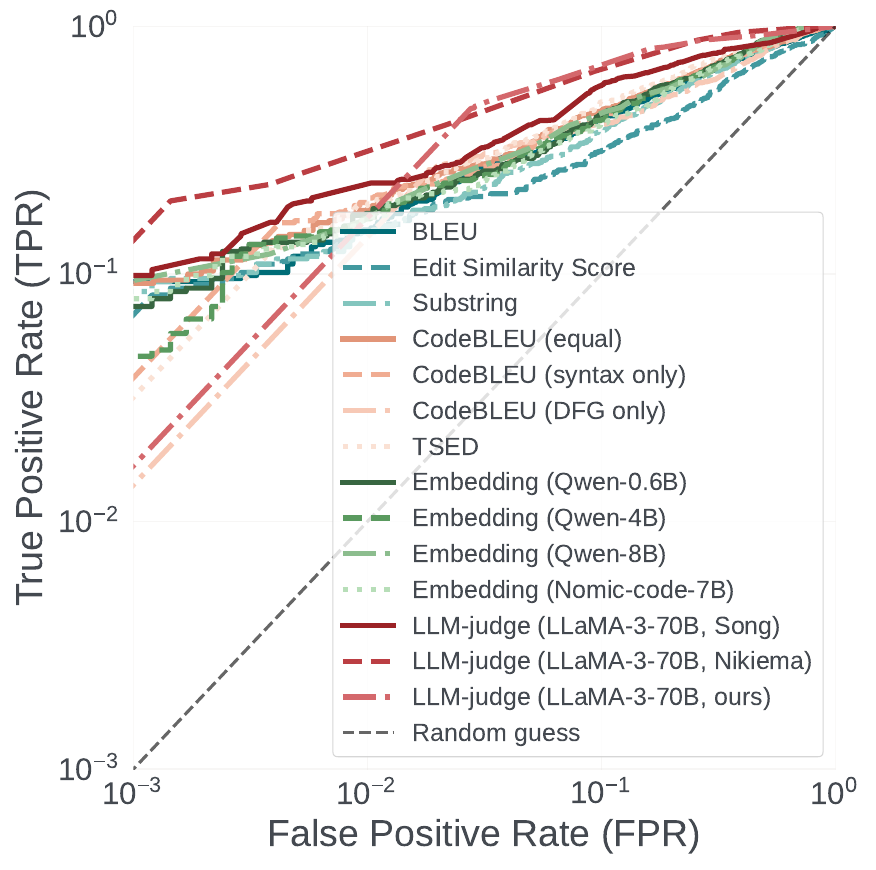} 
  \caption{ROC curve (C)} 
  \end{subfigure} 

  \begin{subfigure}[b]{0.45\linewidth} 
  \includegraphics[width=\linewidth]{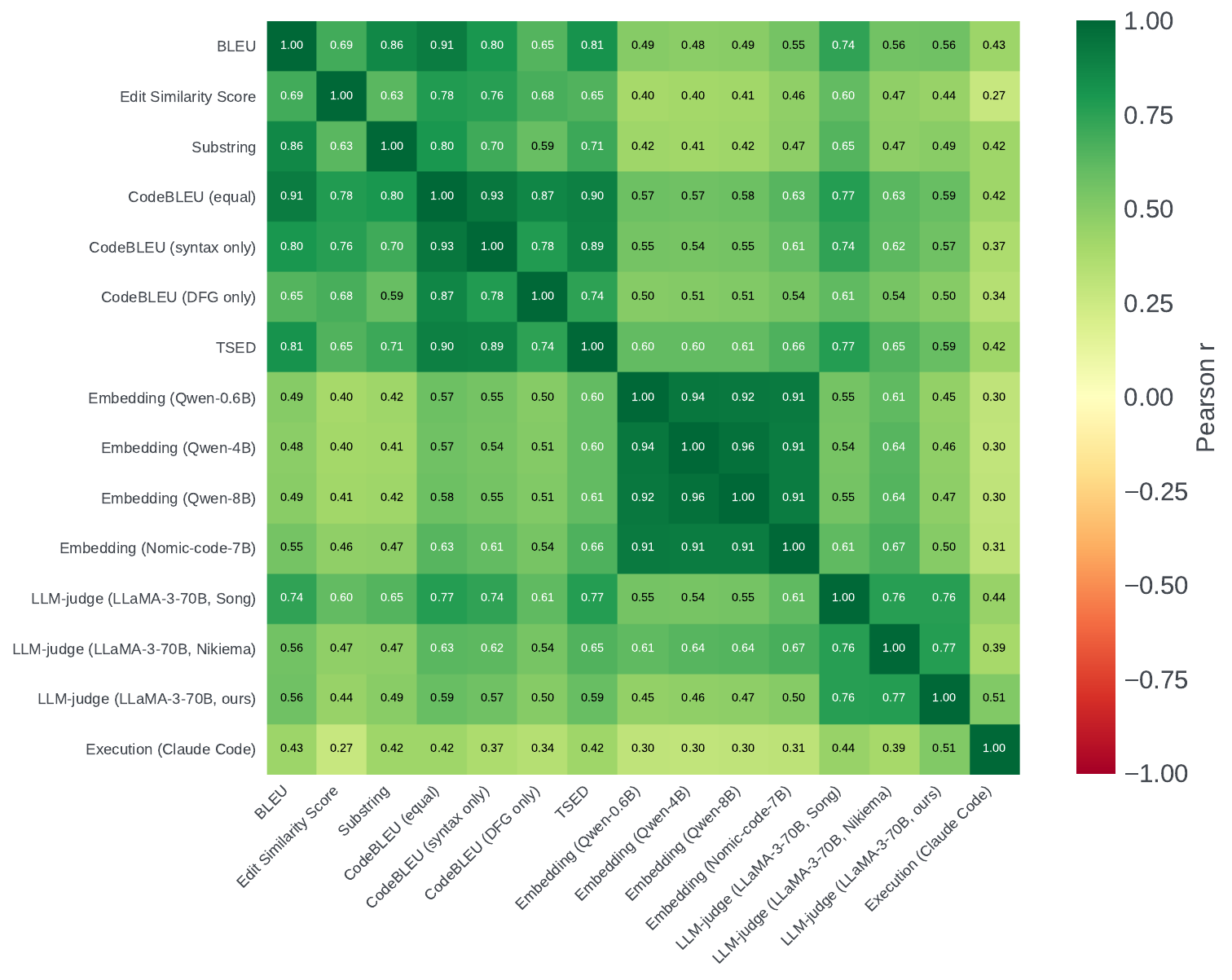} 
  \caption{Correlation (Java)} 
  \end{subfigure} 
  \begin{subfigure}[b]{0.36\linewidth} 
  \includegraphics[width=\linewidth]{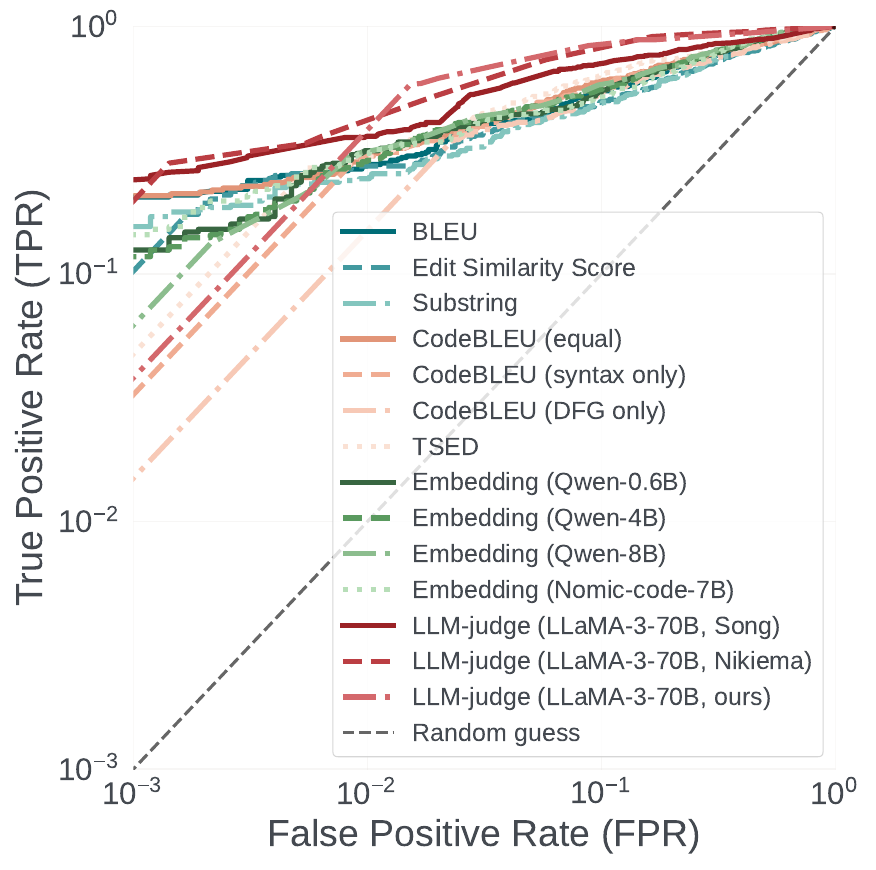} 
  \caption{ROC curve (Java)} 
  \end{subfigure} 
  \caption{Evaluating a range of metrics to approximate execution-based functional similarity for functions across programming languages and generated by the OLMo-3-32B target model (subset for which execution-based testing yields conclusive similarity results). Pearson correlation across textual and functional similarity metrics (left) and ROC curves in log-log considering a binary label, positive if $\textsc{SIM}_\text{func}(x, x^\star_T) \geq \tau_\text{func}$ and negative otherwise. As stated in Section~\ref{sec:exp_setup}, the LLM-judges are instantiated using LLaMA-3.3-70B-Instruct.}
  \label{fig:add_metrics}
\end{figure}

\begin{figure}[t]
  \centering

  \begin{subfigure}[b]{0.45\linewidth} 
  \includegraphics[width=\linewidth]{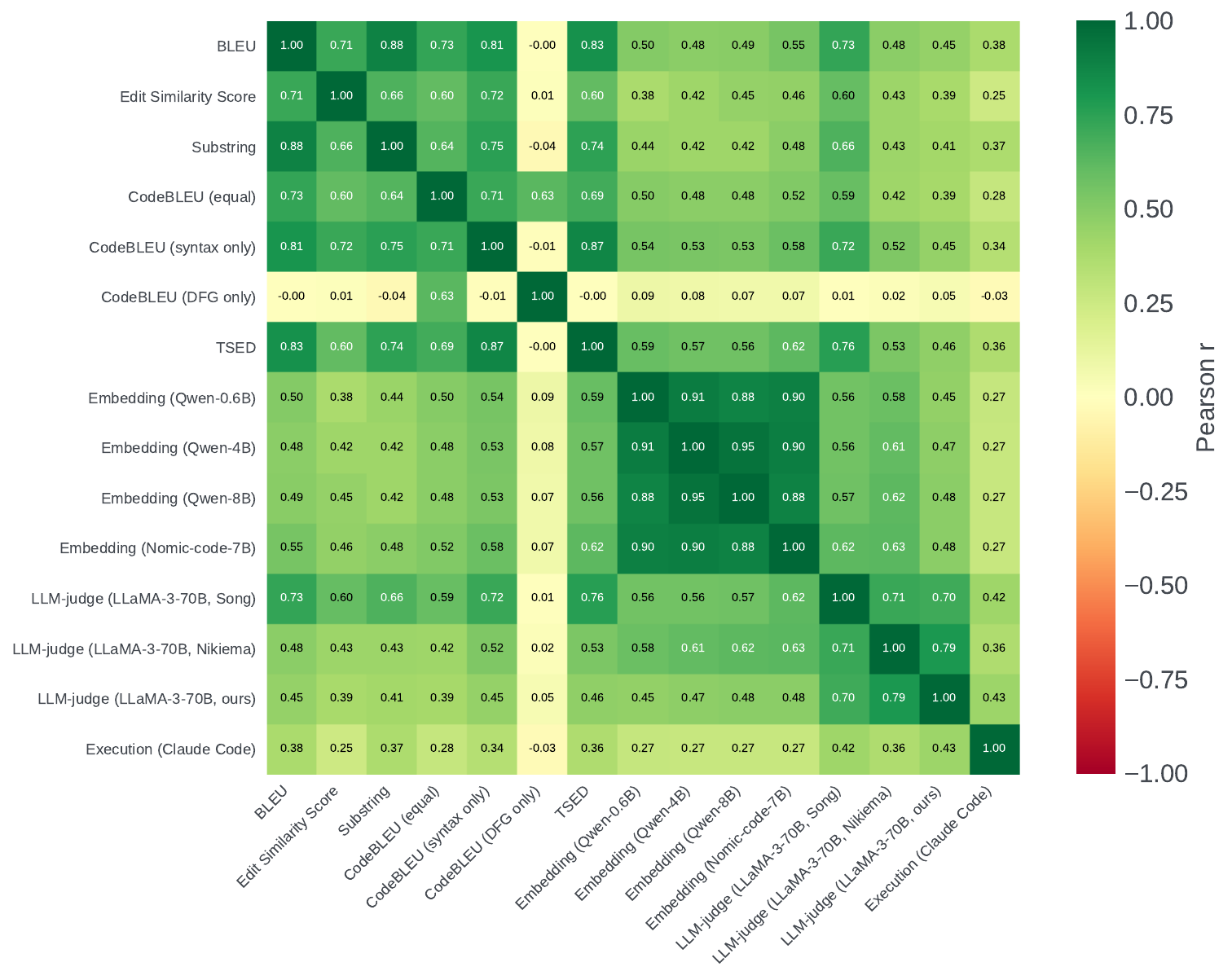} 
  \caption{Correlation (Rust)\footnotemark[\value{fnrust}]} 
  \end{subfigure} 
  \begin{subfigure}[b]{0.36\linewidth} 
  \includegraphics[width=\linewidth]{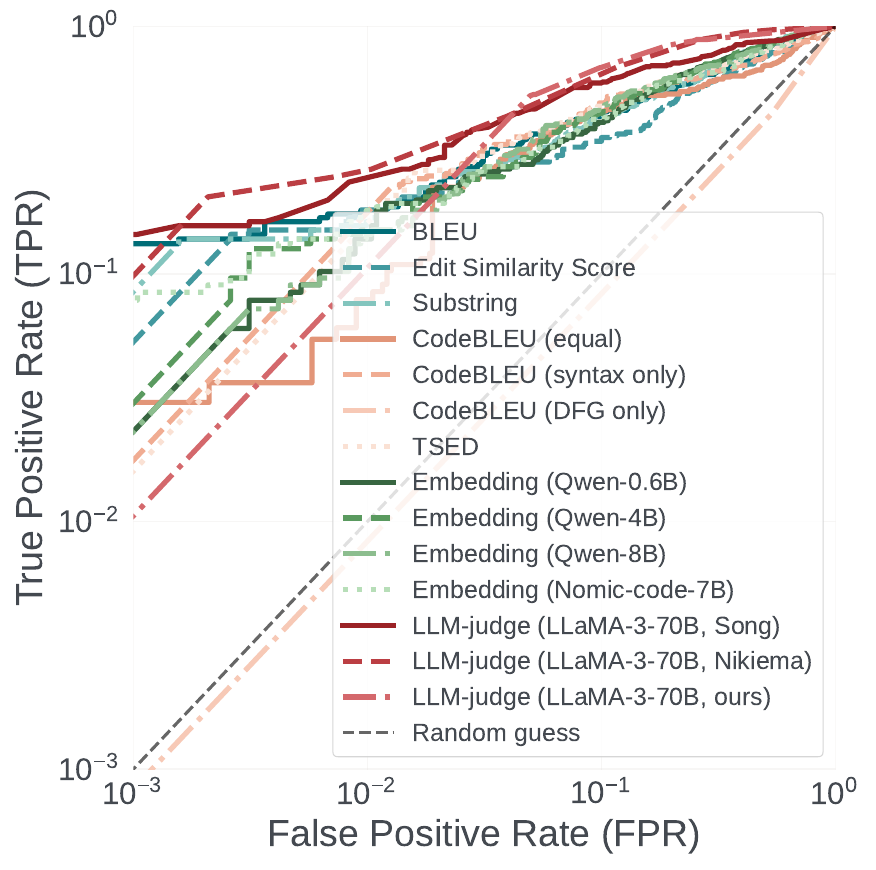} 
  \caption{ROC curve (Rust)\footnotemark[\value{fnrust}]} 
  \end{subfigure} 

  \begin{subfigure}[b]{0.45\linewidth} 
  \includegraphics[width=\linewidth]{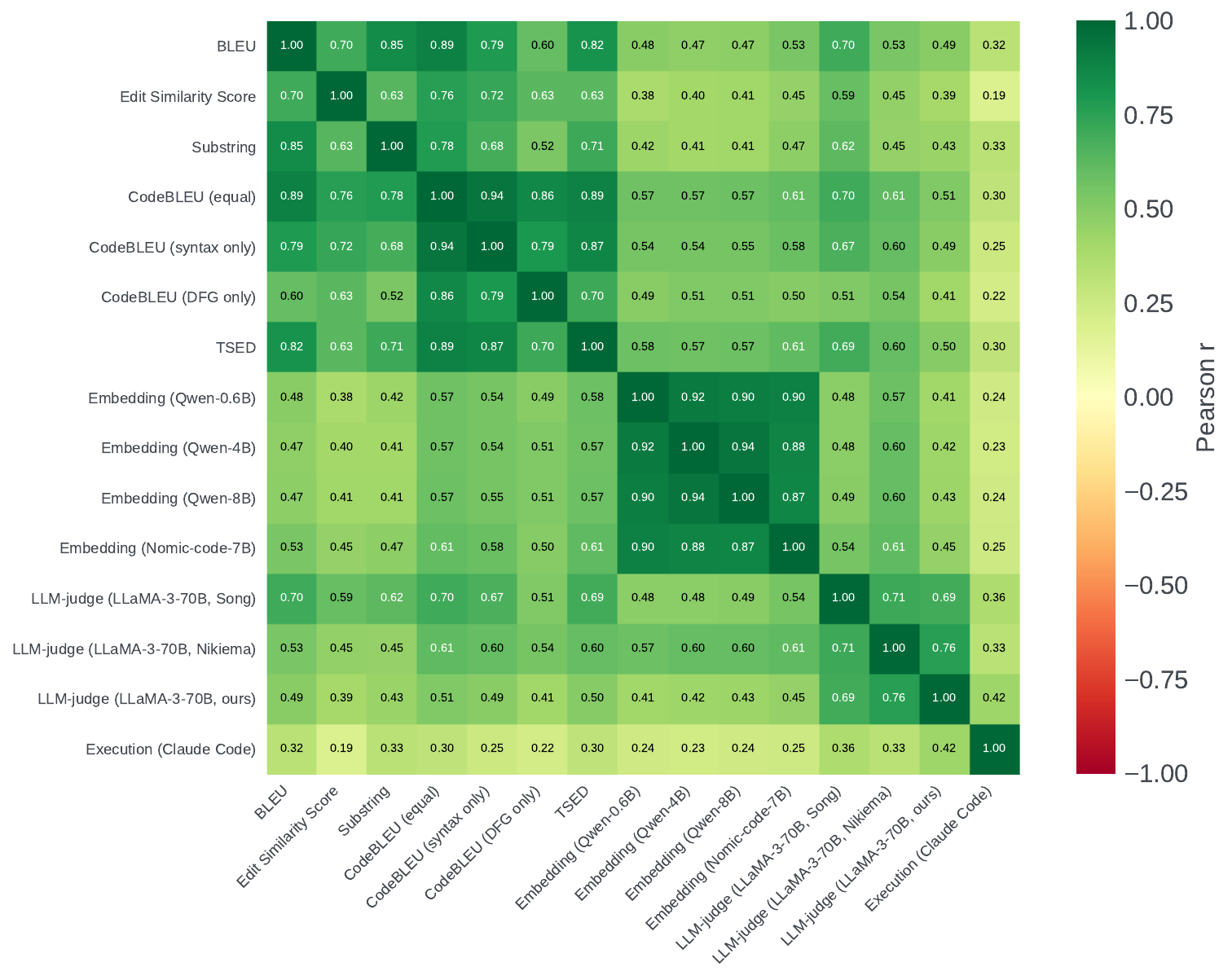} 
  \caption{Correlation (Go)} 
  \end{subfigure} 
  \begin{subfigure}[b]{0.36\linewidth} 
  \includegraphics[width=\linewidth]{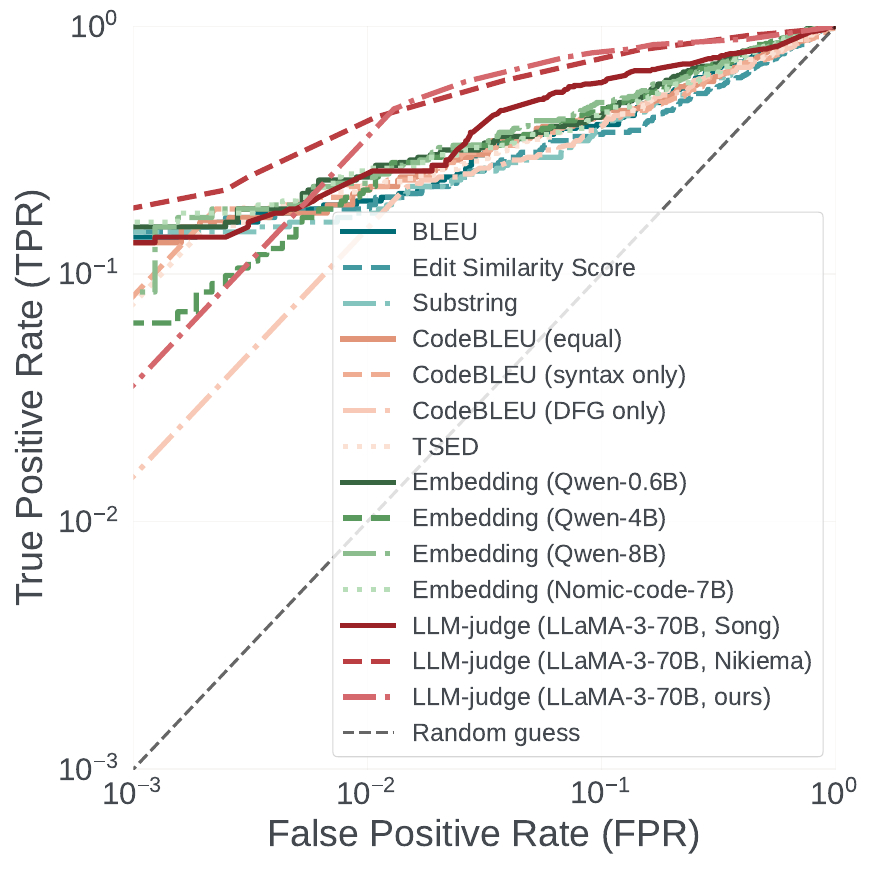} 
  \caption{ROC curve (Go)} 
  \end{subfigure} 

  \caption{Continued: Evaluating a range of metrics to approximate execution-based functional similarity for functions across programming languages and generated by the OLMo-3-32B target model (subset for which execution-based testing yields conclusive similarity results). Pearson correlation across textual and functional similarity metrics (left) and ROC curves in log-log considering a binary label, positive if $\textsc{SIM}_\text{func}(x, x^\star_T) \geq \tau_\text{func}$ and negative otherwise.}
  \label{fig:add_metrics_continued}
\end{figure}

\footnotetext[\value{fnrust}]{\textbf{Note on DFG-only CodeBLEU for Rust.} We find that CodeBLEU's data-flow (DFG) extractor for Rust is uninformative: it fails to traverse function bodies and captures only the signature, so the per-function DFG-match score collapses to either $0$ or $1$ and carrying essentially no signal about the function body. Hence, the DFG-only correlations and ROC performance for Rust are uninformative. Note that this affects only the DFG component; the syntax/AST-based CodeBLEU components parse Rust correctly. We leave a more thorough investigation to future work. See also the related issue at \url{https://github.com/k4black/codebleu/issues/84}.}

%% file: appendix/add_duplication_plots.tex
\begin{figure}[!h]
  \centering
  \begin{subfigure}[b]{0.45\linewidth} 
  \includegraphics[width=\linewidth]{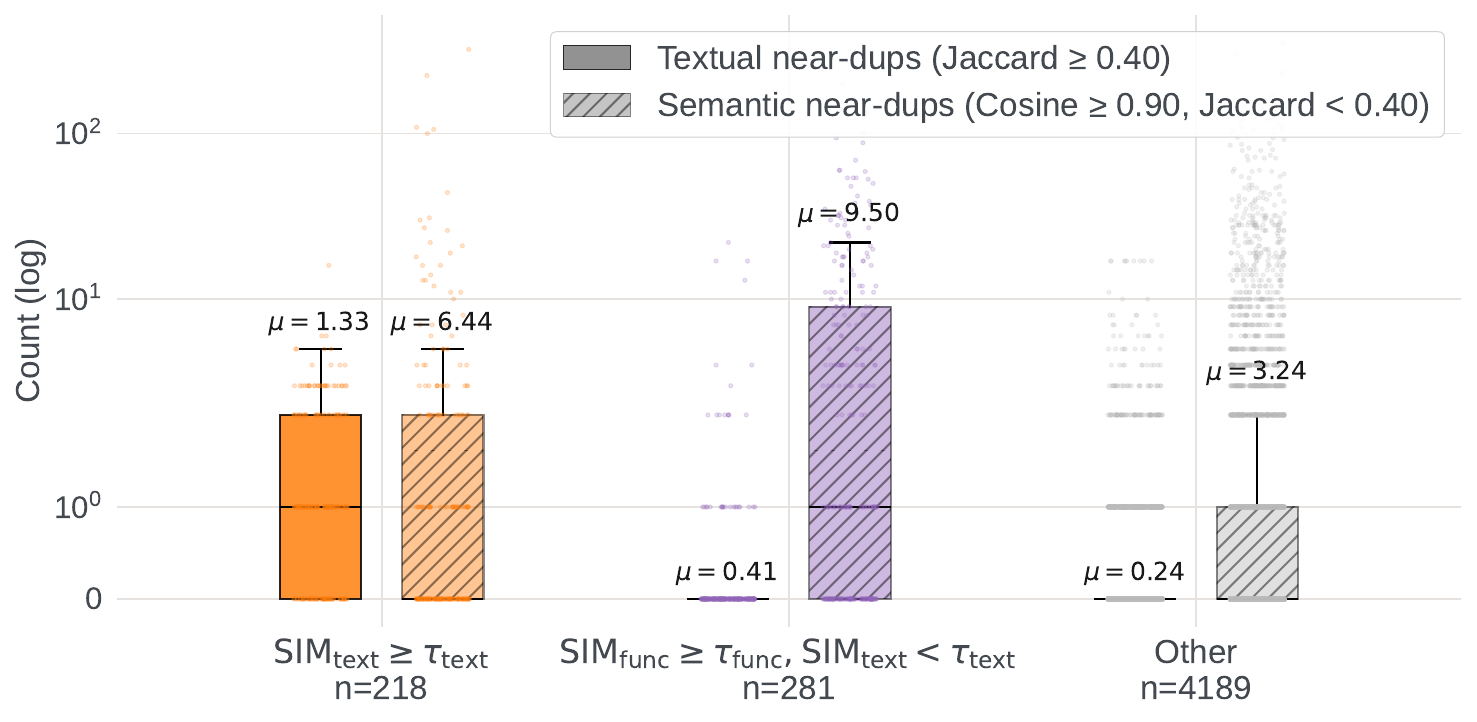} 
  \caption{C} 
  \end{subfigure} 
    
  \begin{subfigure}[b]{0.45\linewidth} 
  \includegraphics[width=\linewidth]{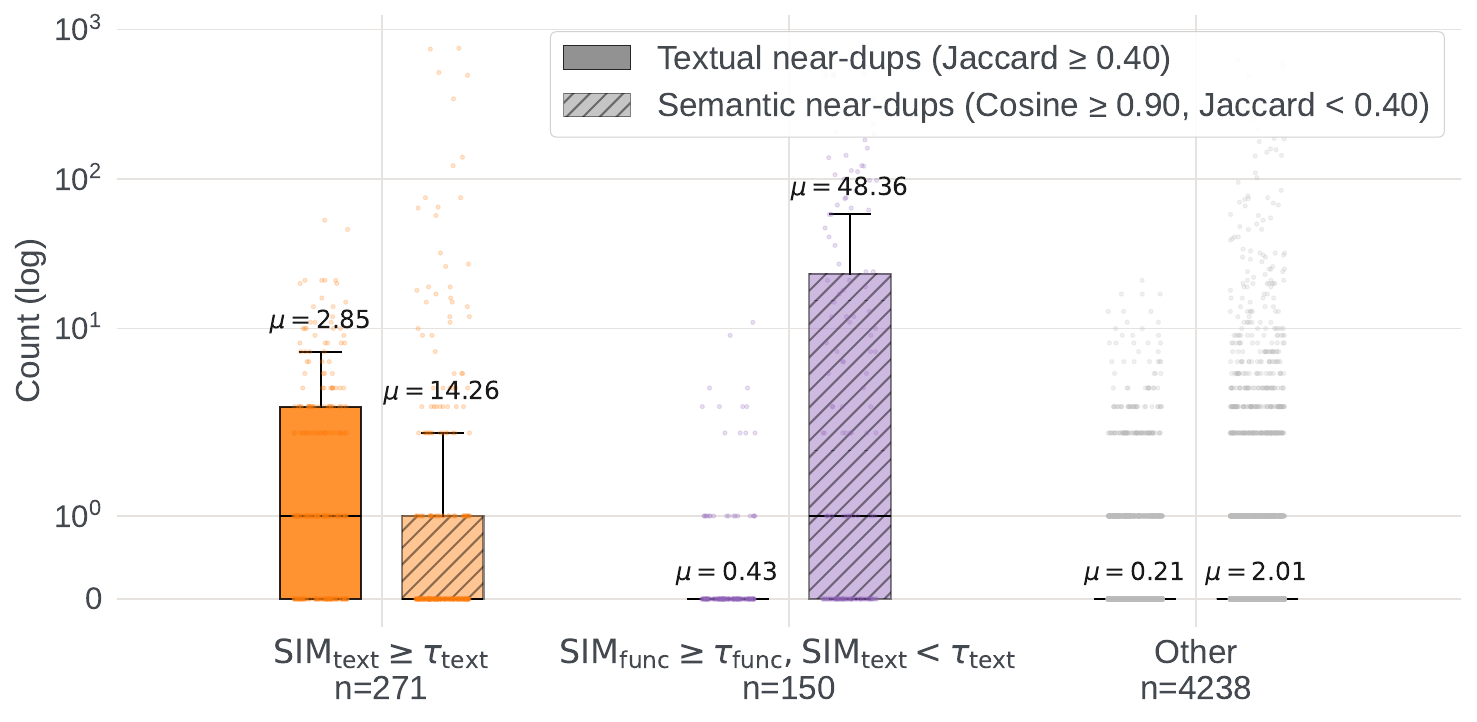} 
  \caption{Java} 
  \end{subfigure} 

  \begin{subfigure}[b]{0.45\linewidth} 
  \includegraphics[width=\linewidth]{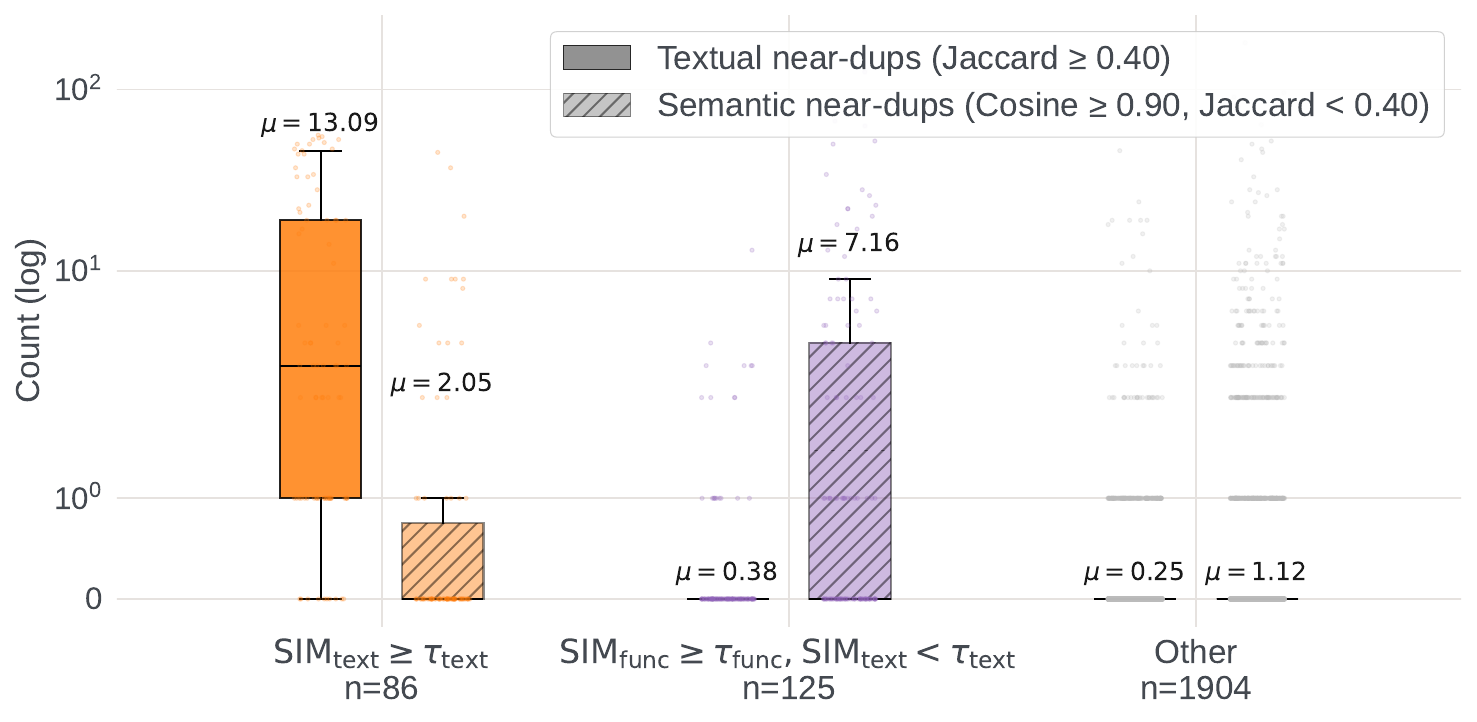} 
  \caption{Rust} 
  \end{subfigure} 

  \begin{subfigure}[b]{0.45\linewidth} 
  \includegraphics[width=\linewidth]{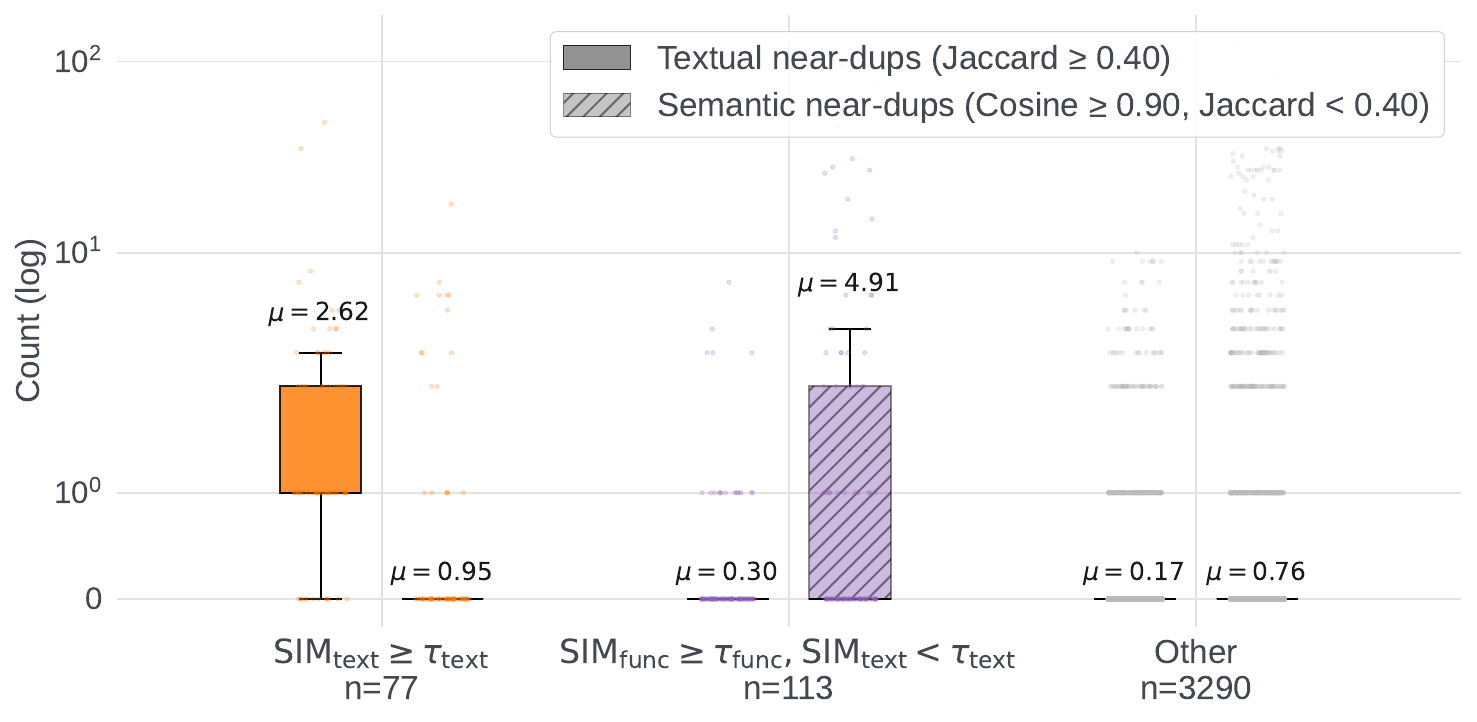} 
  \caption{Go} 
  \end{subfigure} 

  \caption{Distribution of textual and semantic near-duplicate counts in OLMo-3-32B's Dolmino training corpus across programming languages. For each function, we report the number of textual near-duplicates (Jaccard $\geq0.40$; solid boxes) and semantic near-duplicates (cosine $\geq0.90$, Jaccard <0.40; hatched boxes) found in the training data, grouped by category. Categories from left to right: (1) functions for which the target model generates a function with high textual overlap with the ground truth, using BLEU $\tau_\text{text} = 0.75$, (2) functions which were execution based functionally similar with low textual overlap and (3) all the rest we consider with valid execution-based scores.}
  \label{fig:duplicates_all}
\end{figure}

As discussed in Section~\ref{sec:mitigations}, we also report the number of textual and semantic (only) duplicates in the target dataset (Dolmino) for OLMo-32B for each category for other programming languages beyond Python in Figure~\ref{fig:duplicates_all}. 

We also fit a joint logistic regression to predict similarity for each language: 

\begin{equation}
    \mathbbm{1} [\text{SIM} \geq \tau] \sim \beta_{\text{text}} \cdot \text{TextualDups} + \beta_{\text{sem}} \cdot \text{SemanticDups},
\end{equation}

where $\text{TextualDups}$ and $\text{SemanticDups}$ refer to the number of textual and semantic (only) duplicates identified in the training corpus. Table~\ref{tab:logit_dissociation} reports the results. Across all five languages, achieving high textual similarity ($\text{SIM}_{\text{text}} \geq \tau_{\text{text}}$) is strongly and predominately predicted by textual near-duplicate count ($\beta_{\text{text}}$ ranges from $+0.35$ to $+0.86$, all $p < 10^{-16}$; AUC textual-only $\geq 0.87$), while the semantic duplicate count adds little significant predictive power ($p > 0.05$ in four of five languages). Conversely, achieving high functional similarity without textual overlap ($\text{SIM}_{\text{func}} \geq \tau_{\text{func}},\; \text{SIM}_{\text{text}} < \tau_{\text{text}}$) is consistently predicted by semantic near-duplicate count ($\beta_{\text{sem}}$ ranges from $+0.17$ to $+0.35$, all $p < 10^{-3}$; AUC semantic-only $\geq 0.63$), whereas the textual duplicate coefficient is small and non-significant in four of five languages. 

\begin{table}[!h] 
\centering 
\caption{Logistic regression coefficients (standardized) predicting whether the model generation exceeds the textual or functional similarity threshold, from textual and semantic near-duplicate counts in the Dolmino training corpus. Significant coefficients ($p < 0.05$) are \textbf{bolded}. AUC values are computed from univariate models using only the indicated predictor.} 
\label{tab:logit_dissociation} 
\resizebox{0.8\linewidth}{!}{
\begin{tabular}{lc cc cc cc} 
\toprule 
& & \multicolumn{2}{c}{\textbf{Textual Dups}} & \multicolumn{2}{c}{\textbf{Semantic Dups}} & \multicolumn{2}{c}{\textbf{ROC AUC (univariate)}} \\
\cmidrule(lr){3-4} \cmidrule(lr){5-6} \cmidrule(lr){7-8} 
\textbf{Language} & $\textbf{Target}$ & $\beta_{\text{text}}$ & $p$ & $\beta_{\text{sem}}$ & $p$ & Text-only & Sem-only \\ 
\midrule 
\multirow{2}{*}{Python} & $\mathbbm{1} [\text{SIM}_\text{text} \geq \tau_\text{text}]$ & \textbf{+0.68} & $1.5{\times}10^{-25}$ & +0.04 & $7.4{\times}10^{-2}$ & 0.877 & 0.737 \\
& $\mathbbm{1} [\text{SIM}_\text{func} \geq \tau_\text{func} \wedge \text{SIM}_\text{text} < \tau_\text{text}]$ & \textbf{+0.06} & $4.6{\times}10^{-2}$ & \textbf{+0.17} & $3.0{\times}10^{-4}$ & 0.610 & 0.715 \\
\midrule \multirow{2}{*}{C} & $\mathbbm{1} [\text{SIM}_\text{text} \geq \tau_\text{text}]$ & \textbf{+0.35} & $4.5{\times}10^{-17}$ & +0.07 & $1.3{\times}10^{-1}$ & 0.875 & 0.680 \\
& $\mathbbm{1} [\text{SIM}_\text{func} \geq \tau_\text{func} \wedge \text{SIM}_\text{text} < \tau_\text{text}]$ & +0.06 & $2.2{\times}10^{-1}$ & \textbf{+0.18} & $9.4{\times}10^{-7}$ & 0.534 & 0.633 \\ 
\midrule \multirow{2}{*}{Java} & $\mathbbm{1} [\text{SIM}_\text{text} \geq \tau_\text{text}]$ & \textbf{+0.86} & $9.1{\times}10^{-43}$ & \textbf{+0.12} & $1.4{\times}10^{-3}$ & 0.872 & 0.725 \\
& $\mathbbm{1} [\text{SIM}_\text{func} \geq \tau_\text{func} \wedge \text{SIM}_\text{text} < \tau_\text{text}]$ & $-0.12$ & $7.2{\times}10^{-2}$ & \textbf{+0.32} & $4.2{\times}10^{-14}$ & 0.614 & 0.717 \\
\midrule \multirow{2}{*}{Rust} & $\mathbbm{1} [\text{SIM}_\text{text} \geq \tau_\text{text}]$ & \textbf{+0.74} & $1.1{\times}10^{-19}$ & $-0.03$ & $6.9{\times}10^{-1}$ & 0.901 & 0.726 \\
& $\mathbbm{1} [\text{SIM}_\text{func} \geq \tau_\text{func} \wedge \text{SIM}_\text{text} < \tau_\text{text}]$ & $-0.51$ & $9.8{\times}10^{-2}$ & \textbf{+0.35} & $7.6{\times}10^{-8}$ & 0.627 & 0.700 \\
\midrule \multirow{2}{*}{Go} & $\mathbbm{1} [\text{SIM}_\text{text} \geq \tau_\text{text}]$ & \textbf{+0.70} & $6.5{\times}10^{-23}$ & +0.05 & $5.7{\times}10^{-1}$ & 0.928 & 0.708 \\
& $\mathbbm{1} [\text{SIM}_\text{func} \geq \tau_\text{func} \wedge \text{SIM}_\text{text} < \tau_\text{text}]$ & +0.04 & $5.0{\times}10^{-1}$ & \textbf{+0.22} & $7.0{\times}10^{-7}$ & 0.579 & 0.689 \\ 
\bottomrule 
\end{tabular}
} 
\end{table}

Lastly, Figure~\ref{fig:semantic_duplicates} illustrates three semantic duplicates from the Dolmino corpus for the training data function from Figure~\ref{fig:commission_example}. All three duplicates encode highly similar logic, yet in a heavily restructured format, reinforcing functional rather than verbatim memorization. 

\begin{figure*}[!h] 
\noindent
\begin{minipage}[t]{0.29\linewidth} 
\begin{tcolorbox}[colback=purple!3, colframe=purple!70, title={\tiny\textbf{Semantic duplicate 1} (cosine\,=\,0.961)}, boxsep=1pt, left=3pt, right=3pt, top=2pt, bottom=2pt] 
\begin{Verbatim}[fontsize=\tiny] 
def calculate_sales_commission(city: str, 
                        sales_volume: float) -> float: 
commission_rates = { "Sofia": [0.05, 0.07, 0.08, 0.12],
                "Varna": [0.045, 0.075, 0.1, 0.13], 
                "Plovdiv": [0.055, 0.08, 0.12, 0.145]} 
if 0 <= sales_volume <= 500: 
    return sales_volume * commission_rates[city][0] 
elif 500 < sales_volume <= 1000: 
    return sales_volume * commission_rates[city][1] 
elif 1000 < sales_volume <= 10000: 
    return sales_volume * commission_rates[city][2] 
else:
    return sales_volume * commission_rates[city][3] 
\end{Verbatim} 
\end{tcolorbox} 
\end{minipage}
\hfill 
\begin{minipage}[t]{0.3\linewidth} 
\begin{tcolorbox}[colback=purple!3, colframe=purple!70, title={\tiny\textbf{Semantic duplicate 2} (cosine\,=\,0.957)}, boxsep=1pt, left=3pt, right=3pt, top=2pt, bottom=2pt]
\begin{Verbatim}[fontsize=\tiny] 
def calculate_commission_rate( city: str, 
                        sales_amount: float) -> float: 
commission_mapping = {
    "Sofia": [(0,500,5),(500,1000,7), 
            (1000,10000,8),(10000,inf,12)],
    "Varna": [(0,500,4.5),(500,1000,7.5), 
            (1000,10000,10),(10000,inf,13)], 
    "Plovdiv": [(0,500,5.5),(500,1000,8), 
            (1000,10000,12),(10000,inf,14.5)]} 
    
for min_s, max_s, rate in commission_mapping[city]: 
    if min_s <= sales_amount < max_s: 
        return rate 
\end{Verbatim} 
\end{tcolorbox} 
\end{minipage}
\hfill 
\begin{minipage}[t]{0.39\linewidth} 
\begin{tcolorbox}[colback=purple!3, colframe=purple!70, title={\tiny\textbf{Semantic duplicate 3} (cosine\,=\,0.940)}, boxsep=1pt, left=3pt, right=3pt, top=2pt, bottom=2pt] \begin{Verbatim}[fontsize=\tiny] 

def compute_sales_commission( city_name: str, 
                        sales_amount: float) -> float: 
                        
commission_thresholds = {
    'sofia': {500: 0.05, 1000: 0.07, 10000: 0.08, inf: 0.12},      
    'varna': {500: 0.045, 1000: 0.075, 10000: 0.10, inf: 0.13}, 
    'plovdiv': {500: 0.055, 1000: 0.08, 10000: 0.12, inf: 0.145}}

for threshold, rate in sorted(commission_thresholds[city_name].items()): 
    if sales_amount <= threshold: 
        return sales_amount * rate 
        
\end{Verbatim} 
\end{tcolorbox} 
\end{minipage} \caption{\textbf{Semantic duplication in training data.} Three near-duplicates of the function of Figure~\ref{fig:commission_example} from the training corpus encode highly similar logic but differ in structure, reinforcing functional memorization.} 
\label{fig:semantic_duplicates} 
\end{figure*}

%% file: appendix/add_posttraining.tex
As mentioned in Section~\ref{sec:mitigations}, we also evaluate whether post-training techniques can suppress leakage of memorized functions while preserving model utility. We apply two preference-based methods to OLMo-3-32B's 275 counterfactually memorized Python functions (65 textually memorized, 210 functionally memorized, from CraneCode and Stack-Edu).

\subsection{Methods and configurations}

\textbf{Dataset.} The memorized set $\mathcal{D}_m$ consists of the $275$ counterfactually memorized functions ($65$ textually and $210$ functionally memorized). Each example consists of a prefix $p$, the up to ${\approx}250$ preceding tokens ending in the function signature, and the function body $y=(y_1,\dots,y_m)$, on which all losses are computed. Function bodies contain on average $198$ tokens (median $176$; ${\approx}54{,}000$ tokens in total).

\textbf{Common setup.} All methods (NPO and different instantiations of DPO) fine-tune the target model with LoRA~\citep{hulora} of rank $r=32$ and scaling factor $\alpha_{\text{LoRA}}=32$ (dropout $0.05$), applied to all attention and MLP projection matrices of every layer. This yields ${\approx}268$M trainable parameters ($0.84\%$ of the 32B model). We optimize with AdamW ($\beta_1=0.9$, $\beta_2=0.99$, weight decay $0.01$, gradient clipping at $1.0$), an effective batch size of $8$ and learning rate $1\times10^{-5}$ for NPO and $3\times10^{-5}$ for DPO. We train for $10$ epochs and save LoRA checkpoints at epochs $2,4,6,8,10$ to trace the memorization--utility trade-off across post-training. All losses are computed only on the body tokens $y$, masking out the context. 

\textbf{Negative Preference Optimization (NPO).} NPO~\citep{zhangnegative} treats memorized completions as rejected responses without requiring preferred alternatives:

\begin{equation}
  \mathcal{L}_{\text{NPO}} \;=\; -\frac{2}{\beta}\cdot\frac{1}{|\mathcal{D}_m|}\sum_{x\in\mathcal{D}_m}
  \log \sigma\!\left(-\beta\,\log\frac{p_\theta(x)}{p_{\text{ref}}(x)}\right),
\end{equation}

where $p_{\text{ref}}(x)$ is the frozen target prior to fine-tuning and $\beta=0.1$. The bounded objective ensures that a sample stops contributing once its likelihood has been sufficiently reduced, avoiding catastrophic model degradation.

\textbf{Direct Preference Optimization (DPO).} As NPO only pushes the model away from memorized completions without specifying what to produce instead, it risks degrading code generation ability, i.e., valid code shares substantial structural patterns with memorized code, and the space of acceptable alternatives may be more constrained than in natural language settings. DPO~\citep{rafailov2023direct} addresses this by additionally steering the model toward a preferred alternative $y_w$:

\begin{equation}
    \mathcal{L}_{\text{DPO}} \;=\; -\frac{1}{|\mathcal{D}_m|}\sum_{x\in\mathcal{D}_m}
    \log \sigma\!\left(\beta\,\log\frac{p_\theta(y_w\mid c)\,p_{\text{ref}}(y_l\mid c)}{p_{\text{ref}}(y_w\mid c)\,p_\theta(y_l\mid c)}\right),
\end{equation}

with $\beta=0.1$ and $p_{\text{ref}}$ the frozen target, as in NPO. The memorized body serves as the rejected completion $y_l$ and the central design choice is how to obtain $y_w$. We consider the following three options:

\begin{enumerate}
    \item \textbf{DPO-ref.} $y_w$ is the greedy completion of the pretrained reference model for that function. 
    \item \textbf{DPO-target.} We sample completions from the target model at temperature $T>0$ (a first round at $T{=}1.0$ and a second, larger round at $T{=}1.4$ for functions not yet covered) and keep the first that meets the acceptance criterion (see below). For functions the target
    reproduces near-deterministically even at high temperature, no candidate qualifies. For these ($80$ of the $275$ functions ($29\%$)), we fall back to the reference generation.
    \item \textbf{DPO-rewrite.} We prompt a separate model (Claude Sonnet-4.6~\citep{anthropic2026sonnet46}, prompt below), given the surrounding code context and the original function, to write a functionally substantially different but valid rewrite that shares no textual overlap and keeps the exact signature. We draw $K{=}10$ candidates per function across sampling temperatures spread over $[0.6, 1.0]$ and verify each. Here $73$ of $275$ functions ($27\%$) fall back to the reference generation, while $202$ receive a rewrite. 
  \end{enumerate}

For sources~(2) and~(3), we retain a candidate $y_w$ only if it (i) is syntactically valid Python (the reconstructed function parses); (ii) has low textual overlap with the memorized body, $\mathrm{BLEU}(y, y_w)\le 0.3$; and (iii) is not functionally equivalent, with execution similarity $S(y, y_w)\le 0.2$. If no candidate qualifies, we fall back to the reference generation of source~(1). 

\paragraph{Function rewrite prompt.} We use the following prompt to generate functionally different, yet valid preferred alternative responses for the memorized functions:

\begin{promptbox}
\begin{Verbatim}[fontsize=\scriptsize]
System: You are an expert Python engineer. You always return correct, runnable Python.

User: Below is Python code (imports and helper definitions) that ends with a function and its body:
```python
{context ending in the original function and its body}
```
For the last function `{signature}`, write a functionally substantially different version that is valid, runnable Python. Requirements:
1. The function should be functionally different: it should implement an entirely different logic and cannot return the same results 
for the same inputs.
2. It should not share textual overlap with the original body: use different local names, control flow, data structures and comments.
3. It should be valid, runnable code;

Keep the EXACT same signature line and the original indentation.
Return ONLY the complete rewritten function (signature + body) in one ```python block, with no explanation.
\end{Verbatim}
\end{promptbox}

\subsection{Measuring utility}

We measure utility with three complementary metrics, each computed with greedy decoding and normalized to the original target model (OLMo-3-32B):

\textbf{Retain agreement.} We assess whether the model still produces its expected completions on data outside the memorized set. We draw $200$ held-out prompts, sampled in equal proportion from CraneCode and Stack-Edu. For each prompt, we compute the BLEU score between the greedy generations of the post-trained model and the original target model. Note that this measures behavioral drift from post-training, not from the original training data. 

\textbf{Benchmark accuracy.} We additionally evaluate on two code-completion benchmarks that are well-suited to a base (non-post-trained) model such as the OLMo-3-32B target. On HumanEval~\citep{chen2021evaluating} we greedily generate a function from the signature and docstring, truncate it at the first out-of-function token, append the test suite and execute the assembled program. A problem counts as solved if all of its assertions pass and we report \texttt{pass@1} as the fraction of the $164$ problems solved. On MBPP~\citep{austin2021program} we use the $427$ problems of the ``sanitized'' split, taking the first $3$ tasks as few-shot exemplars in a $3$-shot prompt; the greedy continuation is truncated, concatenated with the problem's imports and tests, and executed. We again report \texttt{pass@1} as the fraction of problems whose completion passes all tests (out of the $427-3=424$ remaining problems). 

The \textbf{aggregate utility} is the equal-weighted mean of the three measures, each normalized by its target-model value (so the target has aggregate utility 1 by construction). Table~\ref{tab:raw-bench} reports the raw values for the reference and target model. The reference model's aggregate utility of $0.65$ defines the effective floor: an ideal post-trained model would reduce memorization while retaining utility well above this level.

\begin{table}[!h]
\centering
\caption{Raw (unnormalized) utility measures for the target and reference OLMo-3-32B models. The target's columns are the denominators used to normalize aggregate utility; the reference's aggregate ($0.65$) is the effective utility floor. Retain agreement is BLEU against the target model's own generations, so it is $1.0$ for the target by construction.}
\resizebox{0.8\linewidth}{!}{ 
\begin{tabular}{lcccc}
\toprule
\textbf{Model} & \textbf{Retain agreement} & \textbf{HumanEval \texttt{pass@1}} & \textbf{MBPP \texttt{pass@1}}  & \textbf{Aggregate utility (normalized by target)} \\
\midrule
Target & $1.00$ & $0.512$ ($84/164$) & $0.757$ ($321/424$) & $1.00$ \\
Reference & $0.35$ & $0.372$ ($61/164$) & $0.653$ ($277/424$) & $0.65$ \\
\bottomrule
\end{tabular}
}
\label{tab:raw-bench}
\end{table}

\subsection{Results and discussion}

\begin{figure}[t]
\centering
\includegraphics[width=0.45\linewidth]{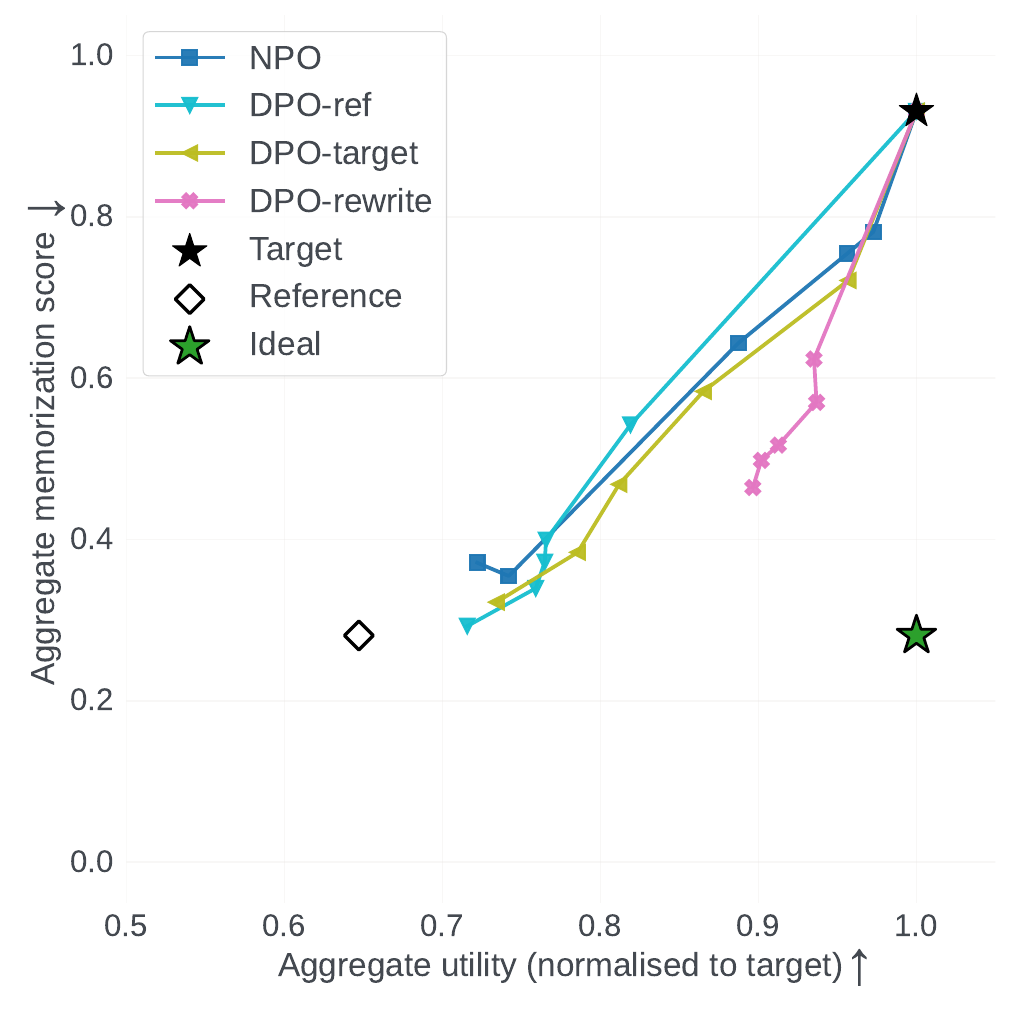}
\caption{Post-training mitigation of memorized Python functions in OLMo-3-32B, plotting aggregate memorization score (lower is better; weighted average of BLEU for textually memorized and execution similarity for functionally memorized) versus aggregate utility (higher is better; generation consistency and benchmark performance, normalized to target). Each curve traces epochs of post-training. Reference points: target model, pretrained reference, and ideal (reference-level memorization, full utility).}
\label{fig:unlearning_combined}
\end{figure}

Figure~\ref{fig:unlearning_combined} plots each method's trajectory during post-training alongside three reference points: the target model (high memorization, high utility), the reference model (low memorization, lower utility), and the ideal outcome that reduces memorization to reference level while retaining utility. Memorization is measured via BLEU (textual) and execution-based similarity (functional) computed on $\mathcal{D}_m$, aggregated as a weighted average. 

All methods that reduce memorization incur some utility cost, consistent with prior work~\citep{mainitofu,shi2025muse}. NPO and DPO-ref follow similar, steep trade-offs: they can reach reference-level memorization ($<0.4$) but only at substantial utility loss ($\sim0.75$). DPO-target achieves a more favorable trajectory, reaching comparable memorization levels while retaining higher utility ($\sim0.80$). DPO-rewrite offers the best trade-off overall, maintaining $>0.9$ utility while meaningfully reducing memorization. The key advantage of DPO over NPO is that it provides a positive learning signal, steering toward a preferred completion rather than merely pushing away from the memorized one. Among DPO variants, higher-quality preferred completions (rewrites > target samples > reference completions) yield better utility preservation. 

\begin{figure}[t]
  \centering
  \includegraphics[width=0.6\linewidth]{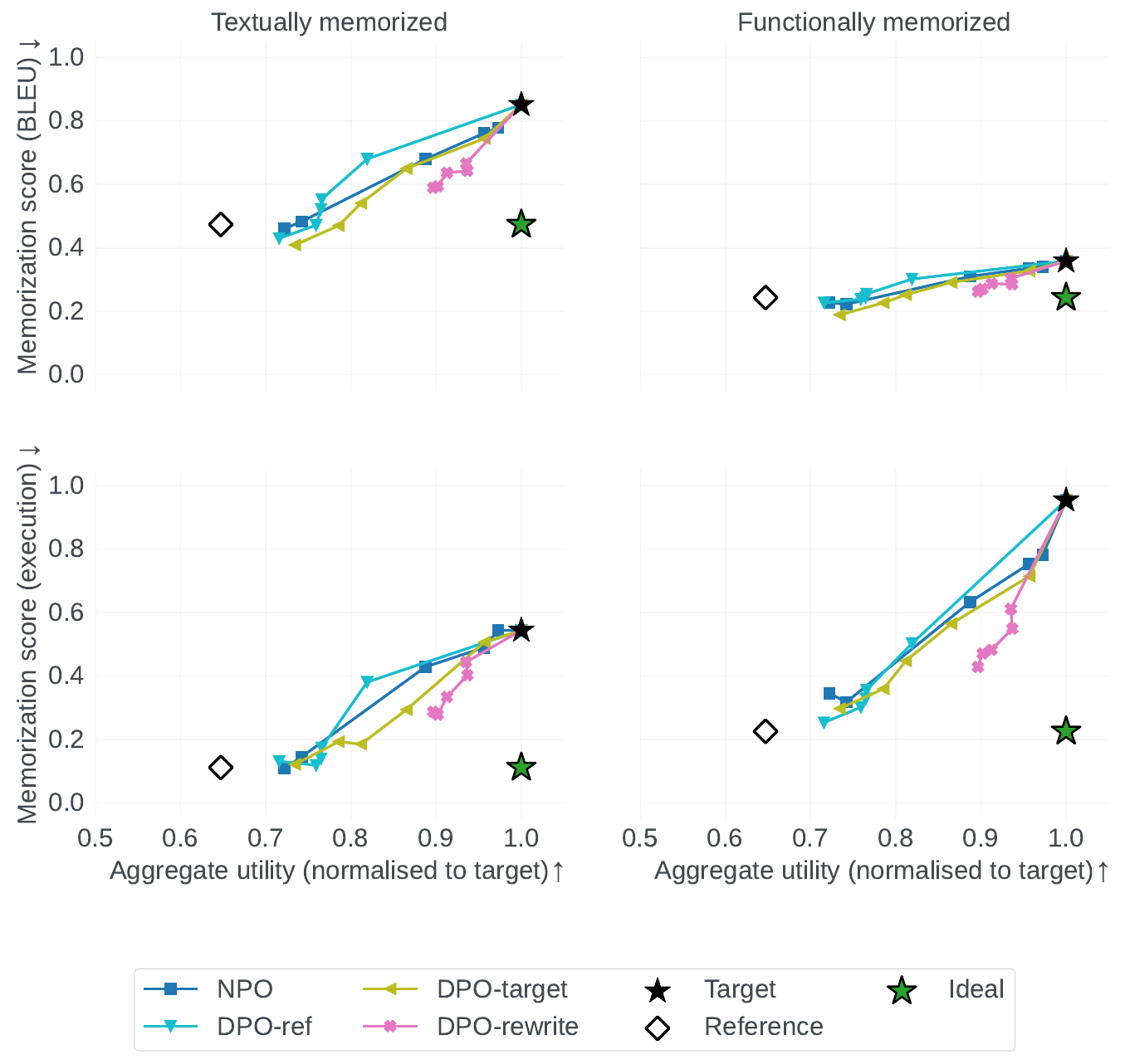}
  \caption{Breaking down the aggregate memorization score from Figure~\ref{fig:unlearning_combined}: reporting BLEU score between greedy generations and training data function (top row) and execution based similarity (bottom row), for $65$ textually (left column) and $210$ functionally (right column) memorized functions.}
  \label{fig:unlearning_breakdown}
\end{figure}

We also provide a further breakdown of the memorization scores across post-training in Figure~\ref{fig:unlearning_breakdown}. We show BLEU (top) and execution-based similarity (bottom) separately for the 65 textually (left) and 210 functionally (right) memorized functions. The relative ranking of methods is consistent across both categories and no method is uniquely suited to one form of memorization. Execution-based similarity tends to decrease more rapidly than BLEU, likely because even minor perturbations (e.g., a single altered branch) cause output divergence on test inputs, whereas BLEU is tolerant of small token-level changes. This suggests that functional memorization may be easier to disrupt, although the same sensitivity means that partial suppression---where core logic remains largely intact---can merely appear as successful mitigation under execution-based metrics.

